\documentclass[conference]{IEEEtran}
\IEEEoverridecommandlockouts
\usepackage[show]{chato-notes}
\usepackage{cite}
\usepackage{url}
\usepackage{hyperref}
\usepackage{amsmath,amssymb,amsfonts}
\usepackage[symbols,toc,automake,acronym]{glossaries-extra}
\usepackage{algorithm, algpseudocode}
\usepackage{graphicx}
\usepackage{textcomp}
\usepackage{tikz}
\usepackage{pgfplots}
\usepgfplotslibrary{groupplots}  % Required for grouping plots
\usepackage{subcaption}  % Required for subfigures
\usepackage{xcolor}
\usepackage{bm}
\usepackage{csvsimple}
\usepackage{filecontents} 
\usepgfplotslibrary{units}  % Optional for units
\pgfplotsset{compat=1.18}
\usetikzlibrary{external}
\usepgfplotslibrary{external}
\def\BibTeX{{\rm B\kern-.05em{\sc i\kern-.025em b}\kern-.08em
    T\kern-.1667em\lower.7ex\hbox{E}\kern-.125emX}}

\newcommand\hypothesis{h} 
\newcommand{\algomap}{\mathcal{A}}
 
\newcommand\estlocalparamsiter[2]{\widehat{\weights}^{(#1)}_{#2}}

\newcommand\estlocalhypositer[2]{\widehat{\hypothesis}^{(#1)}_{#2}}

\newcommand\localmodel[1]{\hypospace^{(#1)}} 
\newcommand\localhypothesis[1]{\hypothesis^{(#1)}} 
\newcommand\learntlocalhypothesis[1]{\widehat{\hypothesis}^{(#1)}}

\newcommand{\publicds}[1]{\mathcal{P}^{(#1)}}
\newcommand\defeq{:=}

\newcommand{\vx}[0]{{\bf x}}
\newcommand{\vv}[0]{{\bf v}}
\newcommand{\vu}[0]{{\bf u}}

\newcommand{\vm}[0]{{\bf m}}
\newcommand{\vq}[0]{{\bf q}}
\newcommand{\mX}[0]{{\bf X}}
\newcommand{\mC}[0]{{\bf C}}
\newcommand{\mA}[0]{{\bf A}}
\newcommand{\mL}[0]{{\bf L}}

\newcommand{\vw}[0]{{\bf w}}

\newcommand{\mQ}{\mathbf{Q}}
\newcommand{\mU}{\mathbf{U}}
\newcommand{\mV}{\mathbf{V}}

\newcommand{\vy}[0]{{\bf y}}
\newcommand{\va}[0]{{\bf a}}

\newcommand{\vz}[0]{{\bf z}}

\newcommand{\vg}[0]{{\bf g}}

\newcommand{\prob}[1]{p({#1})} 
 
\newcommand{\meanvec}[1]{{\bm \mu}^{(#1)}} 
\newcommand{\covmtx}[1]{\mathbf{C}^{(#1)}}

\def \expect {\mathbb{E} }

\newcommand{\biasterm}{B}

\newcommand{\neighbourhood}[1]{\mathcal{N}^{(#1)}}
\newcommand{\nrfolds}{k}

\newcommand{\sensattr}{s} 
\newcommand{\kernel}{K} 
\newcommand{\kernelmap}[2]{K\big(#1,#2\big)}

\newcommand{\normgeneric}[2]{\left\Vert  {#1} \right\Vert_{#2}}

\newcommand{\bmx}[0]{\begin{bmatrix}}
\newcommand{\emx}[0]{\end{bmatrix}}

\newcommand{\featuredim}{d}
\newcommand{\nrfeatures}{\featuredim}

\newcommand{\featurelen}{\featuredim}

\newcommand{\samplesize}{m}
\newcommand{\sampleidx}{r} 
 
\newcommand{\datapoint}{\vz} 
 
\newcommand{\clusteridx}{c} 

\newcommand{\nrcluster}{k} 
 
\newcommand{\featureidx}{j} 
\newcommand{\clustermean}{{\bm \mu}} 
\newcommand{\clustercov}{{\bm \Sigma}}

\newcommand\truelabel{y}

\newcommand\labelvec{\vy}
\newcommand\featurevec{\vx}

\newcommand\feature{x}
\newcommand\predictedlabel{\hat{\truelabel}}
\newcommand\dataset{\mathcal{D}}
\newcommand\trainset{\dataset^{(\rm train)}}

\newcommand\effdim[1]{d_{\rm eff} \left( #1 \right)}

\newcommand{\learnthypothesis}{\hat{\hypothesis}}

\newcommand{\hypospace}{\mathcal{H}}
\newcommand\netmodel[1]{\mathcal{H}^{(#1)}}
\newcommand{\emperror}{\widehat{L}}
\newcommand\emprisk[2]{\widehat{L}\big(#1|#2\big)}
\newcommand\risk[1]{\bar{L} \big( #1 \big) } %\expect \big \{ \loss{(\featurevec,\truelabel)}{#1} \big\}}
\newcommand{\featurespace}{\mathcal{X}}
\newcommand{\labelspace}{\mathcal{Y}}

\newcommand{\eigval}[1]{\lambda_{#1}}
\newcommand{\eigvalgen}{\lambda}
\newcommand{\regparam}{\alpha}
\newcommand{\lrate}{\eta}

\newcommand{\variation}[4]{\delta^{(#1,#2)}\big(#3,#4 \big)}
\newcommand{\gtvloss}[3]{L^{(\delta)} \left({#1},{#2},{#3} \right)}

\DeclareMathOperator*{\argmin}{argmin}
\newcommand{\itercntr}{k}

\newcommand{\iteridx}{k}

\newcommand{\loss}{L}
\newcommand{\lossfun}{L}
\newcommand{\lossfunc}[2]{L\left(#1,#2 \right)}
 
\newcommand{\cluster}{\mathcal{C}}

\newcommand{\featuremtx}{\mX}
\newcommand{\weight}{w}
\newcommand{\weights}{\vw}
\newcommand{\regularizer}[1]{\mathcal{R}\big\{ #1 \big\}}
\newcommand{\decreg}[1]{\mathcal{R}_{#1}}

\newcommand{\featuremapvec}{{\bf \Phi}}

\newcommand{\batch}{\mathcal{B}}

\newcommand{\proximityop}[3]{{\rm\bf prox}_{#1,#3}(#2)}

\newcommand{\locallossfunc}[2]{L_{#1}\left(#2 \right)}

\newcommand{\localdataset}[1]{\mathcal{D}^{(#1)}}

\newcommand{\edges}{\mathcal{E}}

\newcommand{\edgeweight}{A}

\newcommand{\edgeidx}{e}

\newcommand{\graph}{\mathcal{G}}
\newcommand{\nodes}{\mathcal{V}}
\newcommand{\LapMat}[1]{\mL^{(#1)}}
\newcommand{\LapMatEntry}[3]{L^{(#1)}_{#2,#3}}

\newcommand{\nodedegree}[1]{d^{(#1)}}

\newcommand{\nodeidx}{i}
\newcommand{\nrnodes}{n}

\newcommand{\edge}[2]{\{#1,#2\}}

\newcommand{\mvnormal}[2]{\mathcal{N}\left(#1,#2\right)}
 
\newcommand{\gtv}[1]{{\rm GTV}\left\{#1\right\}}

\newcommand{\localsamplesize}[1]{m_{#1}}

\newcommand{\dimlocalmodel}{d}
\newcommand{\pair}[2]{\left( #1,#2 \right)}
\newcommand{\localparams}[1]{\mathbf{w}^{(#1)}}

\newcommand{\estlocalparams}[1]{\widehat{\mathbf{w}}^{(#1)}}

\newcommand{\mutualinformation}[2]{I \left( #1;#2\right)}

\newcommand{\linmodel}[1]{\hypospace^{(#1)}}

\newglossaryentry{minimum}
{
	name=minimum,
	description={Given a set of real numbers, the minimum\index{minimum} is the smallest of those numbers.},
	first={minimum},text={minimum}
}

\newglossaryentry{maximum}
{name=maximum,
 description={Given a set of real numbers, the maximum\index{maximum} is the largest of those numbers.},
 first={maximum},text={maximum}
}

\newglossaryentry{discrepancy}
{name=discrepancy,
	description={
		Consider\index{discrepancy} a \gls{fl} application with \gls{netdata} represented by a \gls{empgraph}. 
		\gls{fl} methods use a discrepancy measure to compare \gls{hypothesis} maps from 
	local models at nodes $\nodeidx,\nodeidx'$ connected by an edge in the \gls{empgraph}.},
	first={discrepancy},text={discrepancy}
}

\newglossaryentry{hfl}
{name={horizontal \gls{fl}},description=
	{Horizontal \gls{fl}\index{horizontal FL} uses \gls{localdataset}s that are 
		constituted by different \gls{datapoint}s but using the same \gls{feature}s 
		to characterize them \cite{HFLChapter2020}. For example, weather 
		forecasting uses a network of spatially distributed weather (observation) 
		stations. Each weather station measures the same quantities such as daily 
		temperature, air pressure and precipitation. However, different weather 
		stations measure the characteristics or \gls{feature}s of different spatio-temporal regions 
		(each such region being a separate \gls{datapoint}).},
	first={horizontal \gls{fl}},text={horizontal \gls{fl}}
} 

\newglossaryentry{dimred}
{name={dimensionality reduction},
	description={Dimensionality reduction\index{dimensionality reduction} methods 
		map (typically many) raw \gls{feature}s to a (relatively small) set of new features. 
		These methods can be used to visualize \gls{datapoint}s by learning two \gls{feature}s 
		that can be used as the coordinates of a depiction in a \gls{scatterplot}.}, first={dimensionality reduction},text={dimensionality reduction}
} 

\newglossaryentry{featlearn}
{name={feature learning},
	description={Feature learning\index{feature learning} refers to the task of learning a map $\featuremapvec$ that 
		reads in raw \gls{feature}s of a \gls{datapoint} and delivers new \gls{feature}s. Different 
		\gls{feature} learning methods are obtained for different quantitative measures for the 
		usefulness of the new \gls{feature}s.},
	first={feature learning},text={feature learning}
} 

\newglossaryentry{autoencoder}
{name={autoencoder},
	description={An autoencoder\index{autoencoder} is a ML method that jointly learns an encoder map 
		$\hypothesis(\cdot) \in \hypospace$ and a decoder map $\hypothesis^{*}(\cdot) \in \hypospace^{*}$. 
		It is an instance of \gls{erm} using a \gls{loss} computed from the reconstruction error 
		$\featurevec - \hypothesis^{*}\big(  \hypothesis \big( \featurevec \big) \big)$.},
	first={autoencoder},text={autoencoder}
} 

\newglossaryentry{vfl}
{name={vertical \gls{fl}},description=
	{Vertical \gls{fl}\index{vertical FL} uses \gls{localdataset}s that are constituted 
	 by the same \gls{datapoint}s but characterizing them with different \gls{feature}s \cite{VFLChapter}. 
     For example, different healthcare providers might all contain information 
     about the same population of patients. However, different healthcare providers 
     collect different measurements (blood values, electrocardiography, lung X-ray) 
     for the same patients.},
	first={vertical \gls{fl}},text={vertical \gls{fl}}
} 

\newglossaryentry{interpretability}
{name={interpretability},description=
		{A ML method is interpretable\index{interpretability} for a specific user if 
			they can well anticipate the \gls{prediction}s delivered by the method. 
			The notion of interpretability can be made precise using quantitative 
			measures of the uncertainty about the \gls{prediction}s \cite{JunXML2020}.},
		first={interpretability},text={interpretability}
}

\newglossaryentry{multitask learning}
{name={multitask learning},description=
	{Multitask learning\index{multitask learning} aims at leveraging relations between 
	 different \gls{learningtask}s. Consider two \gls{learningtask}s obtained from the 
	 same \gls{dataset} of webcam snapshots. The first task is to predict the presence 
	 of a human, while the second is the predict the presence of a car. It might be useful 
	 to use the same \gls{deepnet} structure for both tasks and only allow the weights of 
	 the final output layer to be different.},
	first={multitask learning},text={multitask learning}
}

\newglossaryentry{learningtask}
{name={learning task},description=
	{Consider\index{learning task} a \gls{dataset} $\dataset$ constituted by several \gls{datapoint}s, each of them 
	 characterized by \gls{feature}s $\featurevec$. For example, the \gls{dataset} $\dataset$ 
	 might be constituted by the images of a particular database. Sometimes it might be useful 
	 to represent a \gls{dataset} $\dataset$, along with the choice of \gls{feature}s, by a \gls{probdist} $p(\featurevec)$. 
	 A learning task associated with $\dataset$ consists of a specific 
	 choice for the \gls{label} of a \gls{datapoint} and the corresponding \gls{labelspace}. 
	 Given a choice for the \gls{lossfunc} and \gls{model}, a learning task gives rise to an 
	 instance of \gls{erm}. Thus, we could define a learning task also via an instance of \gls{erm}, i.e., 
	 via an objective function.  Note that, for the same \gls{dataset}, we obtain different learning tasks by using 
	 different choices for the \gls{feature}s and \gls{label} of a \gls{datapoint}. These learning 
	 tasks are related, as they are based on the same \gls{dataset}, and solving them jointly 
	 (via multitask learning methods) is typically preferable over solving them separately \cite{Caruana:1997wk,JungGaphLassoSPL,CSGraphSelJournal}.},
	first={learning task},text={learning task}
}

\newglossaryentry{explainability}
{name={explainability},description=
		{We\index{explainability} define the (subjective) explainability of a ML method 
			as the level of simulatability \cite{Colin:2022aa} of the \gls{prediction}s 
			delivered by a ML system to a human user. Quantitative measures for the 
			(subjective) explainability of a trained \gls{model} can be constructed by 
			comparing its \gls{prediction}s with the \gls{prediction}s provided by a user 
			on a test-set \cite{Zhang:2024aa,Colin:2022aa}. Alternatively, we can use 
			\gls{probmodel}s for \gls{data} and measure explainability of a trained ML model 
			via the conditional (differential) entropy of its \gls{prediction}s, given the user \gls{prediction}s \cite{JunXML2020,Chen2018}. 
		},
		first={explainability},text={explainability}
	}

\newglossaryentry{linmodel}{name={linear model},
	description={Consider\index{linear model}  \gls{datapoint}s, each characterized by a numeric \gls{feature} 
		vector $\featurevec \in \mathbb{R}^{\featuredim}$. A linear \gls{model} is 
		a \gls{hypospace} which consists of all linear maps, 
	\begin{equation} 
		\label{equ_def_lin_model_hypspace}
		\linmodel{\nrfeatures} \defeq \left\{ \hypothesis(\featurevec)= \weights^{T} \featurevec: \weights \in \mathbb{R}^{\nrfeatures} \right\}. 
	\end{equation} 
	Note that \eqref{equ_def_lin_model_hypspace} defines an entire family of \gls{hypospace}s, which is 
	parameterized by the number $\nrfeatures$ of \gls{feature}s that are linearly combined to form the 
	\gls{prediction} $\hypothesis(\featurevec)$. The design choice of $\nrfeatures$ is guided by \gls{compasp} 
	(smaller $\nrfeatures$ means less computation), \gls{statasp} (increasing $\nrfeatures$ might 
	reduce \gls{prediction} error) and \gls{interpretability}. A linear \gls{model} using few carefully chosen 
	\gls{feature}s tends to be considered more interpretable \cite{Ribeiro2016,rudin2019stop}.}, 
   first={linear model},text={linear model}}

\newglossaryentry{gradstep}{name={gradient step},description={Given a \gls{differentiable} 
		real-valued function $f(\cdot): \mathbb{R}^{\nrfeatures} \rightarrow \mathbb{R}$ 
		 and a vector $\weights \in \mathbb{R}^{\nrfeatures}$, the \gls{gradient} step\index{gradient step} 
		 updates $\weights$ by adding the scaled negative \gls{gradient} $\nabla f(\weights)$ to obtain 
		 the new vector 
		 \begin{equation}
		 \label{equ_def_gd_basic} 
		\widehat{\weights}  \defeq \weights - \lrate \nabla f(\weights).
		\end{equation} 
		Mathematically, the gradient step is a (typically non-linear) operator $\mathcal{T}^{(f,\lrate)}$ 
		that is parametrized by the function $f$ and the \gls{stepsize} $\lrate$. 
		\begin{figure}[htbp]
			\begin{center}
				\begin{tikzpicture}[scale=0.8]
					\draw[loosely dotted] (-4,0) grid (4,4);
					\draw[blue, ultra thick, domain=-4.1:4.1] plot (\x,  {(1/4)*\x*\x});
					\draw[red, thick, domain=2:4.7] plot (\x,  {2*\x - 4});
					\draw[<-] (4,4) -- node[right] {$\nabla f(\weights^{(\itercntr)})$} (4,2);
					\draw[->] (4,4) -- node[above] {$-\lrate \nabla f(\weights^{(\itercntr)})$} (2,4);
					\draw[<-] (4,2) -- node[below] {$1$} (3,2) ;
					\node[left] at (-4.1, 4.1) {$f(\cdot)$}; 
					\draw[shift={(0,0)}] (0pt,2pt) -- (0pt,-2pt) node[below] {$\overline{\weights}$};
					\draw[shift={(4,0)}] (0pt,2pt) -- (0pt,-2pt) node[below] {$\weights$};
					\draw[shift={(2,0)}] (0pt,2pt) -- (0pt,-2pt) node[below] {$\mathcal{T}^{(f,\lrate)}(\weights)$};
				\end{tikzpicture}
			\end{center}
			\caption{The basic gradient step \eqref{equ_def_gd_basic} maps a given vector $\weights$ 
			to the updated vector $\weights'$. It defines an operator 
			$\mathcal{T}^{(f,\lrate)}(\cdot): \mathbb{R}^{\nrfeatures} \rightarrow \mathbb{R}^{\nrfeatures}:
			 \weights \mapsto \widehat{\weights}$.}
			\label{fig_basic_GD_step_single}
		\end{figure}
		Note that the gradient step \eqref{equ_def_gd_basic} optimizes locally 
	         (confined to a neighbourhood defined by the \gls{stepsize} $\lrate$) a linear approximation 
		to the function $f(\cdot)$. A natural generalization of \eqref{equ_def_gd_basic} is to locally 
		optimize the function itself (instead of its linear approximation), 
		\begin{align} 
		\label{equ_approx_gd_step}
		\widehat{\weights} = \argmin_{\weights' \in \mathbb{R}^{\dimlocalmodel}} f(\weights')\!+\!(1/\lrate)\normgeneric{\weights-\weights'}{2}^2. 
		\end{align}
		We intentionally use the same symbol $\lrate$ for the parameter in \eqref{equ_approx_gd_step} 
		as we used for the step-size in \eqref{equ_def_gd_basic}. The larger we choose $\lrate$ in 
		\eqref{equ_approx_gd_step}, the more progress the update will make towards reducing the 
		function value $f(\widehat{\weights})$. Note that, much like the gradient step \eqref{equ_def_gd_basic}, 
		also the update \eqref{equ_approx_gd_step} defines a (typcially non-linear) operator 
		that is paramterized by the function $f(\cdot)$ and the parameter $\lrate$. For convex $f(\cdot)$, this operator is 
		known as the \gls{proxop} of $f(\cdot)$ \cite{ProximalMethods}. 
		\begin{figure}
			\begin{center}
				\begin{tikzpicture}[scale=0.8]
					\draw[blue, ultra thick, domain=-4.1:4.1] plot (\x, {(1/4)*\x*\x}) node[above right] {$f(\weights')$};		
					\draw[red, thick, domain=1:3] plot (\x, {2*(\x - 2)*(\x - 2)}) node[below right] {$(1/\lrate)\normgeneric{\weights-\weights'}{2}^{2}$};
					\draw[shift={(2,0)}] (0pt,2pt) -- (0pt,-2pt) node[below] {$\weights$};
				\end{tikzpicture}
			\end{center}
			\caption{A generalized gradient step updates a vector $\weights$ by minimizing a penalized version 
			of the function $f(\cdot)$. The penalty term is the squared Euclidean distance from the vector $\weights$.}
			\label{fig_quadratic_comparison}
		\end{figure}
		},first={gradient step},text={gradient step}}

\newglossaryentry{proxop}{name={proximal operator},description={Given\index{proximal operator} a \gls{convex} function 
		and a vector $\weights'$, we define its proximal operator as \cite{ProximalMethods,Bauschke:2017} 
		$$\proximityop{\locallossfunc{\nodeidx}{\cdot}}{\weights'}{2 \regparam}\defeq \argmin_{\weights \in \mathbb{R}^{\dimlocalmodel}} \bigg[ f(\weights)\!+\!(\rho/2) \normgeneric{\weights- \weights'}{2}^{2}\bigg] \mbox{ with } \rho > 0. $$ 
		\Gls{convex} functions for which the proximal operator can be computed efficiently 
		are sometimes referred to as \emph{proximable} or \emph{simple} \cite{Condat2013}.},first={proximal operator},text={proximal operator}}

\newglossaryentry{proximable}{name={proximable},description={A\index{proximable} 
		\gls{convex} function for which the \gls{proxop} can be computed efficiently are 
		sometimes referred to as \emph{proximable} or \emph{simple} \cite{Condat2013}.},first={proximable},text={proximable}}

\newglossaryentry{connected}{name ={connected graph}, description={A\index{connected graph} 
		undirected graph $\graph=\pair{\nodes}{\edges}$ is connected\index{connected graph} if 
		it does not contain a (non-empty) subset $\nodes' \subset \nodes$ with no edges leaving 
		$\nodes'$.}, first={connected},text={connected}}

\newglossaryentry{mvndist}{name ={multivariate normal distribution}, description={The\index{multivariate normal distribution} 
		multivariate normal distribution $\mvnormal{\vm}{\mC}$ is an 
		important family of \gls{probdist}s for a continuous \gls{rv} $\featurevec \in \mathbb{R}^{\nrfeatures}$ \cite{BertsekasProb,GrayProbBook,Lapidoth09}. 
		This family is parameterized by the mean $\vm$ and \gls{covmtx} $\mC$ of $\featurevec$. 
		If the \gls{covmtx} is invertible, the \gls{probdist} of $\featurevec$ is 
		$$p(\featurevec) \propto \exp\bigg(-(1/2) \big( \featurevec - \vm \big)^{T} \mC^{-1} \big( \featurevec - \vm \big) \bigg).$$}, first={multivariate normal distribution},text={multivariate normal distribution}}

\newglossaryentry{statasp}{name ={statistical aspects}, description={By statistical aspects\index{statistical aspects} 
		of a ML method, we refer to (properties of) the \gls{probdist} of its output 
		under a \gls{probmodel} for the data fed into the method.},first={statistical aspects},text={statistical aspects}}

\newglossaryentry{compasp}{name ={computational aspects}, description={By computational 
		aspects\index{computational aspects} of a ML method, we mainly refer to the computational 
		resources required for its implementation. For example, if a ML method uses iterative 
		optimization techniques to solve \gls{erm}, then its computational aspects include (i) how 
		many arithmetic operations are needed to implement a single iteration (\gls{gradstep}) 
		and (ii) how many iterations are needed to obtain useful \gls{modelparams}. One important 
		example of an iterative optimization technique is \gls{gd}.}, first={computational aspects},text={computational aspects}}

\newglossaryentry{zerooneloss}{name={$0/1$ \gls{loss}},
	description={The $0/1$ \gls{loss}\index{$0/1$ loss} $\lossfunc{\pair{\featurevec}{\truelabel}}{\hypothesis}$ 
		measures the quality of a \gls{classifier} $\hypothesis(\featurevec)$ that delivers a \gls{prediction} $\predictedlabel$ (e.g., 
	via thresholding \eqref{equ_def_threshold_bin_classifier}) for the \gls{label} $\truelabel$ of a \gls{datapoint} with \gls{feature}s 
	$\featurevec$. 
	It is equal to $0$ if the \gls{prediction} is correct, i.e., 
	$\lossfunc{\pair{\featurevec}{\truelabel}}{\hypothesis}=0$ when $\predictedlabel=\truelabel$. It is 
	equal to $1$ if the \gls{prediction} is wrong, $\lossfunc{\pair{\featurevec}{\truelabel}}{\hypothesis}=1$ 
	when $\predictedlabel\neq\truelabel$.},
	sort=zerooneloss, 
    first={$0/1$ \gls{loss}},text={$0/1$ \gls{loss}}}

\newglossaryentry{probability}{name={probability},
	description={We\index{probability} assign a probability value, typically chosen in the 
		interval $[0,1]$, to each event that might occur in a random experiment \cite{KallenbergBook,BertsekasProb,BillingsleyProbMeasure,HalmosMeasure}.},first={probability},text={probability}}
	
\newglossaryentry{underfitting}{name={underfitting},description={Consider\index{underfitting} a ML method that uses 
		\gls{erm} to learn a \gls{hypothesis} with minimum \gls{emprisk} on a given \gls{trainset}. 
		Such a method is \emph{underfitting} the \gls{trainset} if it is not able to learn a \gls{hypothesis} 
		with sufficiently small \gls{emprisk} on the \gls{trainset}. If a method is underfitting it will typically 
	also not be able to learn a \gls{hypothesis} with a small \gls{risk}.},first={underfitting},text={underfitting}}

\newglossaryentry{overfitting}{name={overfitting},description={Consider\index{overfitting} a ML method that uses 
		\gls{erm} to learn a \gls{hypothesis} with minimum \gls{emprisk} on a given \gls{trainset}. 
		Such a method is \emph{overfitting} the \gls{trainset} if it learns \gls{hypothesis} with small 
		\gls{emprisk} on the \gls{trainset} but significantly larger \gls{loss} outside the \gls{trainset}.},first={overfitting},text={overfitting}}

\newglossaryentry{gdpr}{name={General Data Protection Regulation},description={
			The\index{GDPR} General Data Protection Regulation (GDPR) was enacted by the European Union (EU), 
			effective from May 25, 2018 \cite{GDPR2016}. It safeguards the privacy and data rights of individuals in the EU. 
			The GDPR has significant implications for how data is collected, stored, and used in ML 
			applications. Key provisions include:
			\begin{itemize}
				\item \Gls{dataminprinc}: ML systems should only use necessary amount of personal 
				data for their purpose.
				\item Transparency and \Gls{explainability}: ML systems should enable their users to 
				understand how they make decisions that impact them.
				\item Data Subject Rights: Including the rights to access, rectify, and delete 
				personal data, as well as to object to automated decision-making and profiling.
				\item Accountability: Organizations must ensure robust data security and demonstrate 
				compliance through documentation and regular audits.
			\end{itemize}
			}, 
	first={general data protection regulation (GDPR) },text={GDPR}}
	
\newglossaryentry{gaussrv}{name={Gaussian random variable},description={
		A \index{Gaussian random variable} standard Gaussian \gls{rv} is a 
		real-valued random variable $x$ with \gls{pdf} \cite{papoulis,BertsekasProb,GrayProbBook}
		\begin{equation}
			\nonumber
			p(x) = \frac{1}{\sqrt{2\pi}} \exp^{-x^2/2}. 
		\end{equation}
		Given a standard Gaussian \gls{rv} $x$, we can construct a general Gaussian \gls{rv} $x'$ with 
		mean $\mu$ and variance $\sigma^2$ via $x' \defeq \sigma (x+\mu)$. The \gls{probdist} of a 
		Gaussian \gls{rv} is referred to as normal distribution, denoted $\mathcal{N}(\mu,\sigma)$.  \\ 
		A Gaussian random vector $\featurevec \in \mathbb{R}^{\featuredim}$ with 
		\gls{covmtx} $\mathbf{C}$ and mean ${\bm \mu}$ can be constructed via 
		$\featurevec \defeq \mathbf{A} \big( \vz + {\bm \mu} \big)$. Here, $\mA$ 
		is any matrix that satisfies $\mA\mA^{T} = \mC$ and $\vz \defeq \big( z_{1},\ldots,z_{\featuredim} \big)^{T}$
		is a vector whose entries are \gls{iid} standard Gaussian \gls{rv}s $z_{1},\ldots,z_{\featuredim}$. Gaussian 
		random processes generalize Gaussian random vectors by applying linear transformations to infinite sequences 
		of standard Gaussian \gls{rv}s \cite{Rasmussen2006Gaussian}.\\
		Gaussian \gls{rv}s are widely used \gls{probmodel}s for the statistical analysis of 
		machine learning methods. Their significance arises partly from the central limit theorem 
		which states that the sum of many independent \gls{rv}s (not necessarily Gaussian themselves) 
		tends to a Gaussian \gls{rv} \cite{ross2013first}. 
},first={Gaussian RV},text={Gaussian RV}}

\newglossaryentry{trustworthiness}{name={trustworthiness},description=
	{Beside the \gls{compasp} and \gls{statasp}, a third main design aspect for ML methods 
		is their trustworthiness\index{trustworthy AI} \cite{pfau2024engineeringtrustworthyaideveloper}. 
		The European Union has put forward seven key requirements (KRs) for trustworthy 
		AI (that typically build on ML methods)
	\cite{ALTAIEU}: {\bf KR1 - Human Agency and Oversight}, {\bf KR2 - Technical Robustness and Safety}, 
	{\bf KR3 - Privacy and Data Governance}, {\bf KR4 - Transparency}, {\bf KR5 - Diversity Non-Discrimination and Fairness}, 
	{\bf KR6 Societal and Environmental Well-Being}, {\bf KR7 - Accountability}. 
	},first={trustworthiness},text={trustworthiness}}

\newglossaryentry{sqerrloss}{name={squared error loss},description={The squared 
		error\index{squared error loss} \gls{loss} measures the prediction error of a 
		\gls{hypothesis} $\hypothesis$ when predicting a numeric \gls{label} $\truelabel \in \mathbb{R}$ 
		from the \gls{feature}s $\featurevec$ of a \gls{datapoint}. It is 
	defined as 
\begin{equation} 
	\nonumber
	\lossfunc{(\featurevec,\truelabel)}{\hypothesis} \defeq \big(\truelabel - \underbrace{\hypothesis(\featurevec)}_{=\predictedlabel} \big)^{2}. 
\end{equation} 
},first={squared error loss},text={squared error loss}}

\newglossaryentry{projgd}{name={projected gradient descent (projected GD)},description={Projected\index{projected gradient descent (projected GD)} \gls{gd} 
		extends basic \gls{gd} for unconstrained optimization to handle constraints on the 
		optimization variable (\gls{modelparams}). A single iteration of projected \gls{gd} consists 
		of first taking a \gls{gradstep} and then projecting the result back into a 
		constrain set.},first={projected \gls{gd}},text={projected \gls{gd}}}

\newglossaryentry{diffpriv}
{name=differential privacy,
  description={
  	Consider some ML method $\algomap$ that reads in a \gls{dataset} (e.g., the \gls{trainset} 
  	used for \gls{erm}) and delivers some output $\algomap(\dataset)$. The output 
  	could be either the learnt \gls{modelparams} or the \gls{prediction}s for specific \gls{datapoint}s. 
  	Differential privacy is a precise measure of privacy leakage incurred by revealing the 
  	output. Roughly speaking, a ML method is differentially private if the \gls{probdist} 
  	of the output $\algomap(\dataset)$ does not change too much if the \gls{sensattr} 
  	of one \gls{datapoint} in the \gls{trainset} is changed. Note that differential privacy 
  	builds on a \gls{probmodel} for a ML method, i.e., we interpret its output $\algomap(\dataset)$ 
  	as the \gls{realization} of a \gls{rv}. The randomness in the output can be ensured 
  	by intentionally adding the \gls{realization} of an auxiliary \gls{rv} (\emph{noise}) to 
  	the output of the ML method.}, 
	first = {differential privacy (DP)}, text={DP} 
}

\newglossaryentry{privprot}
{name=privacy protection,
     description={Consider\index{privacy protection} some ML method $\algomap$ that reads in a $\dataset$ 
     	and delivers some output $\algomap(\dataset)$. The output could be the learnt \gls{modelparams} 
     	$\widehat{\weights}$ or the \gls{prediction} $\learnthypothesis(\featurevec)$ obtained for a specific \gls{datapoint} 
		with \gls{feature}s $\featurevec$. Many important ML applications involve \gls{datapoint}s 
		representing humans. Each \gls{datapoint} is characterized by \gls{feature}s $\featurevec$, 
		potentially a \gls{label} $\truelabel$ and a \gls{sensattr} $\sensattr$ (e.g., a recent medical diagnosis). 
		Roughly speaking, \gls{privprot} means that it should be impossible to infer, from the output $\algomap(\dataset)$, 
		any of the \gls{sensattr}s of \gls{datapoint}s in $\dataset$. Mathematically, \gls{privprot} requires non-invertibility 
		of the map $\algomap(\dataset)$. In general, just making $\algomap(\dataset)$ non-invertible 
		is typically insufficient for \gls{privprot}. We need to make $\algomap(\dataset)$ sufficiently non-invertible. 
	}, 
	first = {privacy protection}, text={privacy protection} 
}

\newglossaryentry{privleakage}
{
	name=privacy leakage,
	description={Consider\index{privacy leakage} a (ML or \gls{fl}) system that processes a \gls{localdataset} $\localdataset{\nodeidx}$ 
		and shares data, such as the predictions obtained for new \gls{datapoint}s, with 
		other parties. Privacy leakage arises if the shared data carries information about a 
		private (sensitive) \gls{feature} of a \gls{datapoint} (which might be a human) of $\localdataset{\nodeidx}$.  
		The amount of privacy leakage can be measured via \gls{mutualinformation} using a 
		\gls{probmodel} for the local \gls{dataset}. Another quantitative measure for privacy 
		leakage is \gls{diffpriv}. 
	}, 
	first = {privacy leakage}, text={privacy leakage} 
}

\newglossaryentry{probmodel}
{
	name=probabilistic model,
	description={A probabilistic model\index{probabilistic model} interprets \gls{datapoint}s 
		as \gls{realization}s of \gls{rv}s with a joint \gls{probdist}. This joint \gls{probdist} typically 
		involves parameters which have to be manually chosen or learnt via statistical inference 
		methods such as \gls{ml} \cite{LC}. }, 
	first = {probabilistic model}, text={probabilistic model} 
}

\newglossaryentry{mean}
{
	name=mean,
	description={The\index{mean} expectation $\expect \{ \featurevec \}$ of a numeric \gls{rv} $\featurevec$.}, 
		first = {mean}, text={mean} 
}

\newglossaryentry{variance}
{
	name={variance},
	description={The\index{variance} variance of a real-valued \gls{rv} $\feature$ is defined as the expectation 
		$\expect\big\{ \big( x - \expect\{x \} \big)^{2} \big\}$ of the squared difference between $\feature$ 
		and its expectation $\expect\{x \}$. We extend this definition to vector-valued \gls{rv}s $\featurevec$ 
		as $\expect\big\{ \big\| \featurevec - \expect\{\featurevec \} \big\|_{2}^{2} \big\}$.} ,first={variance},text={variance} 
}

\newglossaryentry{nn}
{
	name={nearest neighbour},
	description={Nearest neighbour\index{nearest neighbour} methods learn a \gls{hypothesis} 
		$\hypothesis: \featurespace \rightarrow \labelspace$ whose function value $\hypothesis(\featurevec)$ 
		is solely determined by the nearest neighbours within a given \gls{dataset}. Different 
		methods use different metrics for determining the nearest neighbours. If \gls{datapoint}s 
		are characterized by numeric \gls{feature} vectors, we can use their Euclidean distances as 
		the metric.},
	first={nearest neighbour (NN)},text={NN} 
}

\newglossaryentry{neighbourhood}
{
	name={neighbourhood},
	description={The\index{neighbourhood} neighbourhood of a node $\nodeidx \in \nodes$ is 
	the subset of nodes constituted by the \gls{neighbours} of $\nodeidx$.},
	first={neighbourhood},text={neighbourhood} 
}

\newglossaryentry{neighbours}
{
	name={neighbours},
	description={The\index{neighbours} neighbours of a node $\nodeidx \in \nodes$ within a \gls{empgraph} are 
	those nodes $\nodeidx' \in \nodes \setminus \{ \nodeidx\}$ that are connected (via an edge) to node $\nodeidx$.},
	first={neighbours},text={neighbours} 
}

\newglossaryentry{bias}
{
	name={bias},
	description={Consider\index{bias} a ML method using a parameterized \gls{hypospace} $\hypospace$. 
		It learns the \gls{modelparams} $\weights \in \mathbb{R}^{\dimlocalmodel}$ using the \gls{dataset} $\dataset=\big\{ \pair{\featurevec^{(\sampleidx)}}{\truelabel^{(\sampleidx)}} \big\}_{\sampleidx=1}^{\samplesize}$. 
		To analyze the properties of the ML method, we typically interpret the \gls{datapoint}s as \gls{realization}s 
		of \gls{iid} \gls{rv}s, $$ \truelabel^{(\sampleidx)} = \hypothesis^{(\overline{\weights})}\big( \featurevec^{(\sampleidx)} \big) + \bm{\varepsilon}^{(\sampleidx)}, \sampleidx=1,\ldots,\samplesize.$$ 
		We can then interpret the ML method as an estimator $\widehat{\weights}$, 
		computed from $\dataset$ (e.g., by solving \gls{erm}). The (squared) bias incurred by the estimate $\widehat{\weights}$ 
		is then defined as $\biasterm^{2} \defeq \big\| \expect \{ \widehat{\weights}  \}- \overline{\weights}\big\|_{2}^{2}$. },
first={bias},text={bias} 
}

\newglossaryentry{classification}
{
	name={classification},
	description={Classification\index{classification} is the task of determining a 
		discrete-valued label $\truelabel$ of a \gls{datapoint} based solely on its 
		features $\featurevec$. The label $\truelabel$ belongs to a finite set, such 
		as $\truelabel \in \{ -1,1\}$, or $\truelabel \in \{1,\ldots,19\}$ and represents a 
		category to which the corresponding \gls{datapoint} belongs to. Some classification 
		problems involve a countably infinite \gls{labelspace}.},first={classification},text={classification} 
}

\newglossaryentry{privfunnel}
{
	name={privacy funnel},
	description={The privacy funnel is a method for learning privacy-friendly \gls{feature}s 
		of \gls{datapoint}s \cite{PrivacyFunnel}.},
	first={privacy funnel},text={privacy funnel} 
}

\newglossaryentry{condnr}
{
	name={condition number},
	description={The condition number\index{condition number} $\kappa(\mathbf{Q}) \geq 1$ of a 
		positive definite 
		matrix $\mathbf{Q} \in \mathbb{R}^{\featurelen \times \featurelen}$ is the ratio 
		$\eigvalgen_{\rm max} /\eigvalgen_{\rm min}  $ between the 
		largest $\eigvalgen_{\rm max}$ and the smallest $\eigvalgen_{\rm min}$ \gls{eigenvalue} of 
		$\mathbf{Q}$. The condition number is useful for the analysis of ML methods. 
		The computational complexity of \gls{gdmethods} for \gls{linreg} crucially depends on the 
		condition number of the matrix $\mQ = \mX \mX^{T}$, with the \gls{featuremtx} $\mX$ 
		of the \gls{trainset}. Thus, from a computational perspective, we prefer \gls{feature}s of 
		\gls{datapoint}s such that $\mQ$ has a condition number close to $1$.},first={condition number},text={condition number} 
}

\newglossaryentry{classifier}
{
	name={classifier},
	description={A classifier\index{classifier} is a \gls{hypothesis} (map) $\hypothesis(\featurevec)$ 
		used to predict a \gls{label} taking values from a finite \gls{labelspace}. We might use the 
		function value $\hypothesis(\featurevec)$ itself as a \gls{prediction} $\predictedlabel$ for 
		the \gls{label}. However, it is customary to use a map $\hypothesis(\cdot)$ that delivers 
		a numeric quantity. The \gls{prediction} is then obtained by a simple thresholding step. 
		For example, in a binary \gls{classification} problem with \label{labelspace} $\labelspace \in  \{ -1,1\}$, 
		we might use real-valued \gls{hypothesis} map $\hypothesis(\featurevec) \in \mathbb{R}$ 
		as classifier. A \gls{prediction} $\predictedlabel$ can then be obtained via thresholding,  
		 \begin{equation} 
		 	\label{equ_def_threshold_bin_classifier}
		 	\predictedlabel =1   \mbox{ for } \hypothesis(\featurevec) \geq 0, \mbox{ and } 	\predictedlabel =-1  \mbox{ otherwise.}
	 		\end{equation}
 		We can characterize a classifier by its \gls{decisionregion}s $\decreg{a}$, for 
 		every possible \gls{label} value $a \in \labelspace$. },first={classifier},text={classifier} 
}

\newglossaryentry{emprisk}
{name={empirical risk},
  description={The empirical risk\index{empirical risk} $\emprisk{\hypothesis}{\dataset}$ 
  	of a \gls{hypothesis} on a \gls{dataset} $\dataset$ is the average \gls{loss} incurred 
  	by $\hypothesis$ when applied to the \gls{datapoint}s in $\dataset$.},
  first={empirical risk},text={empirical risk} 
}

\newglossaryentry{nodedegree}
{name={node degree},
	description={The degree\index{node degree} $\nodedegree{\nodeidx}$ of a node $\nodeidx \in \nodes$ 
		in an undirected \gls{graph} is the number of its \gls{neighbours}, $\nodedegree{\nodeidx} \defeq \big|\neighbourhood{\nodeidx}\big|$.},first={node degree},text={node degree} 
}

\newglossaryentry{graph}
{name={graph},
	description={A graph\index{graph} $\graph = \pair{\nodes}{\edges}$ is a pair that consists of 
		a node set $\nodes$ and an edge set $\edges$. In its most general form, a graph is 
		specified by a map that assigns to each edge $\edgeidx \in \edges$ a pair of nodes \cite{RockNetworks}. 
		One important family of graphs are simple undirected graphs. A simple undirected graph 
		is obtained by identifying each edge $\edgeidx \in \edges$ with two different nodes $\{\nodeidx,\nodeidx'\}$. 
		Weighted graphs also specify numeric weights $\edgeweight_{\edgeidx}$ for each 
		edge $\edgeidx \in \edges$.},first={graph},text={graph} 
}

\newglossaryentry{empgraph}
{name={federated learning (FL) network},
	description={A federated network\index{federated learning (FL) network} is an undirected weighted \gls{graph} whose 
		nodes represent data generators that aim to train a local (or personalized) \gls{model}. 
		Each node in a federated network represents some device, capable to collect a \gls{localdataset} 
		and, in turn, train a local \gls{model}. 
	    \Gls{fl} methods learn a local \gls{hypothesis} $\localhypothesis{\nodeidx}$, for 
	    each node $\nodeidx \in \nodes$, such that it incurs small \gls{loss} on the \gls{localdataset}s.},first={federated learning (FL) network},text={FL network} 
}

\newglossaryentry{norm}
{name={norm},
	description={A norm\index{norm} is a function that maps each element (vector) 
		of a linear vector space to a non-negative real number. This function must be 
		homogeneous, definite and satisfy the triangle inequality \cite{HornMatAnalysis}. },
	first={norm},text={norm} 
}

\newglossaryentry{explanation}
{name={explanation},
	description={One approach to make ML methods transparent, is to provide an 
		explanation\index{explanation} along with the \gls{prediction} delivered by an 
		ML method. Explanations can take on many different forms. An explanation 
		could be some natural text or some quantitative measure for the importance 
		of individual \gls{feature}s of a \gls{datapoint} \cite{Molnar2019}. We can also 
		use visual forms of explanations such as intensity plots for image classification \cite{GradCamPaper}.},
	first={explanation},text={explanation} 
}

\newglossaryentry{risk}
{name={risk},
	description={Consider\index{risk} a \gls{hypothesis} $\hypothesis$ used to predict the \gls{label} 
		$\truelabel$ of a \gls{datapoint} based on its \gls{feature}s $\featurevec$. We measure 
		the quality of a particular \gls{prediction} using a \gls{lossfunc} $\lossfunc{(\featurevec,\truelabel)}{\hypothesis}$. 
		If we interpret \gls{datapoint}s as the \gls{realization}s of \gls{iid} \gls{rv}s, 
		also the $\lossfunc{(\featurevec,\truelabel)}{\hypothesis}$ becomes the \gls{realization} 
		of a \gls{rv}. The \gls{iidasspt} allows to define the risk of a \gls{hypothesis} 
		as the expected \gls{loss} $\expect \big\{\lossfunc{(\featurevec,\truelabel)}{\hypothesis} \big\}$. 
		Note that the risk of $\hypothesis$ depends on both, the specific choice for the \gls{lossfunc} and the 
		\gls{probdist} of the \gls{datapoint}s.},
	first={risk},text={risk} 
}

\newglossaryentry{actfun}
{name={activation function},
	description={Each\index{activation function} artificial neuron within an \gls{ann} is 
		assigned an activation function $g(\cdot)$ that maps a weighted combination of 
		the neuron inputs $\feature_{1},\ldots,\feature_{\nrfeatures}$ to a single output 
		value $a = g\big(\weight_{1} \feature_{1}+\ldots+\weight_{\nrfeatures} \feature_{\nrfeatures} \big)$. 
		Note that each neuron is parameterized by the weights $\weight_{1},\ldots,\weight_{\nrfeatures}$.},
first={activation function},text={activation function} 
}

\newglossaryentry{transparency}
{name={transparency},
	description={Transparency\index{transparency} is a key requirement for 
		trustworthy AI \cite{HLEGTrustworhtyAI}. In the context of ML methods, 
		such as \gls{erm}-based methods, transparency is mainly used synonymously for \gls{explainability} \cite{gallese2023ai,JunXML2020}. 
		However, in the wide context of AI systems, transparency also includes providing information 
		about limitations and reliability of the AI system. As a point in case, \gls{logreg} provides a 
		quantitative measure for the reliability of a \gls{classification} in the form of the value $|\hypothesis(\featurevec)|$. 
		Transparency also includes the user interface, by requiring to clearly indicate when a user is 
		interaction with an AI system. Another component of transparency is the documentation 
		of the system’s purpose, design choices and intended use cases \cite{Shahriari2017,DatasheetData2021,10.1145/3287560.3287596}. },
	first={transparency},text={transparency} 
}

\newglossaryentry{sensattr}
{name={sensitive attribute},
	description={ML\index{sensitive attribute} revolves around learning a \gls{hypothesis} map that allows 
		to predict the \gls{label} of a \gls{datapoint} from its \gls{feature}s. In some 
		applications we must ensure that the output delivered by an ML system does 
		not allow to infer sensitive attributes of a \gls{datapoint}. Which parts of a \gls{datapoint} 
		is considered as a sensitive attribute is a design choice that varies across 
		different application domains.},
	first={sensitive attribute},text={sensitive attribute} 
}

\newglossaryentry{sbm}
{name={stochastic block model},
	description={The\index{stochastic block model} stochastic block model (SBM) is a 
		probabilistic generative model for an undirected graph $\graph = \big( \nodes, \edges \big)$ 
		with a given set of nodes $\nodes$ \cite{AbbeSBM2018}. In its most basic variant, 
		the SBM generates a graph by first randomly assigning each node $\nodeidx \in \nodes$ to 
		a cluster index $\clusteridx_{\nodeidx} \in \{1,\ldots,\nrcluster\}$. A pair of different nodes in the 
		graph is connected by an edge with probability $p_{\nodeidx,\nodeidx'}$ that depends 
		solely on the labels $\clusteridx_{\nodeidx}, \clusteridx_{\nodeidx'}$. 
		The presence of edges between different pairs of 
		nodes is statistically independent. },
	first={stochastic block model (SBM)},text={SBM} 
}

\newglossaryentry{deepnet}
{name={deep net},
	description={A\index{deep net} deep net is a \gls{ann} with a (relatively) large number of 
	hidden layers. Deep learning is an umbrella term for ML methods that use a deep net as 
	their model \cite{Goodfellow-et-al-2016}.},
	first={deep net},text={deep net} 
}

\newcommand{\gaussiancenter}{3}

\newglossaryentry{baseline}
{name={baseline},
    description={Consider\index{baseline} some ML method that delivers a learnt \gls{hypothesis} (or trained \gls{model}) 
    $\learnthypothesis \in \hypospace$. We evaluate the quality of a trained \gls{model} 
    by computing the average \gls{loss} on a \gls{testset}. But how do we know that the resulting 
    \gls{testset} performance is good (enough)? How can we determine if the trained \gls{model} performs 
    close to optimal and there is little point in investing more resources (for \gls{data} collection 
	or computation) to improve it? To this end, it is useful to have a reference (or baseline) level 
    against which we can compare the performance of the trained \gls{model}. 
    Such a reference value might be obtained from human performance, e.g., the misclassification
    rate of dermatologists who diagnose cancer from visual inspection of skin. Another 
    source for a baseline is an existing, but for some reason unsuitable, ML method. 
    For example, the existing ML method might be computationally too expensive for 
    the intended ML application. However, we might still use its \gls{testset} error 
    as a baseline. Another, somewhat more principled, approach to constructing a 
    baseline is via a \gls{probmodel}. For a wide range of \gls{probmodel}s $p(\featurevec,\truelabel)$ 
    we can precisely determine the minimum achievable \gls{risk} among \emph{any} \gls{hypothesis} (not even 
    required to belong to the \gls{hypospace} $\hypospace$) \cite{LC}. 
    This minimum achievable \gls{risk} (referred to as the \gls{bayesrisk}) is the \gls{risk} 
    of the \gls{bayesestimator} for the \gls{label} $\truelabel$ of a \gls{datapoint}, given
    its \gls{feature}s $\featurevec$. Note that, for a given choice of \gls{lossfunc}, the 
    \gls{bayesestimator} (if it exists) is completely determined by the \gls{probdist} $p(\featurevec,\truelabel)$ \cite[Chapter 4]{LC}. 
    However, there are two challenges to computing the \gls{bayesestimator} and \gls{bayesrisk}:
    i) the \gls{probdist} $p(\featurevec,\truelabel)$ is unknown and needs to be estimated 
    but (ii) even if we know $p(\featurevec,\truelabel)$ it might be computationally too expensive  
	to compute the \gls{bayesrisk} exactly. 
	A widely used \gls{probmodel} is the \gls{mvndist} $\pair{\featurevec}{\truelabel} \sim \mathcal{N}({\bm \mu},{\bm \Sigma})$ 
	for \gls{datapoint}s characterised by numeric \gls{feature}s and \gls{label}s.
    Here, for the \gls{sqerrloss}, the \gls{bayesestimator} is given by the posterior 
	mean $\mu_{\truelabel|\featurevec}$ of the \gls{label} $\truelabel$ given the 
	\gls{feature}s $\featurevec$ \cite{LC,GrayProbBook}. The corresponding \gls{bayesrisk} 
	is given by the posterior \gls{variance} 
	$\sigma^{2}_{\truelabel|\featurevec}$ (see Figure \ref{fig_post_baseline}).
	\begin{figure}[h]
		\begin{center}
		\begin{tikzpicture}
			\draw[->] (-1,0) -- (7,0) node[right] {$\truelabel$}; % x-axis
			\draw[thick,domain=-1:7,smooth,variable=\x] 
			  plot ({\x}, {2*exp(-0.5*((\x-\gaussiancenter)^2))});
			\draw[dashed] (\gaussiancenter,0) -- (\gaussiancenter,2.5);
			\node[anchor=south] at ([yshift=-5pt] \gaussiancenter,2.5) {\small $\mu_{\truelabel|\featurevec}$};
			\draw[<->,thick] (\gaussiancenter-1,1) -- (\gaussiancenter+1,1.0);
			\node[anchor=west] at ([yshift=2pt] \gaussiancenter,1.2) {\small $\sigma_{\truelabel|\featurevec}$};
  			\foreach \x in {0.5} {
				\node[red] at (\x, 0) {\bf \large $\times$};
 			 }
  			\node[anchor=north] at (0.5,-0.2) {\small $\learnthypothesis(\featurevec)$};
		  \end{tikzpicture}
		\end{center}
		\caption{If \gls{feature}s and \gls{label} of a \gls{datapoint} are drawn from a \gls{mvndist}, we 
		can achievive minimum \gls{risk} (under \gls{sqerrloss}) by using the \gls{bayesestimator} $\mu_{\truelabel|\featurevec}$ 
		to predict the \gls{label} $\truelabel$ of a \gls{datapoint} with \gls{feature}s $\featurevec$. The corresponding 
		minimum \gls{risk} is given by the posterior \gls{variance} $\sigma^{2}_{\truelabel|\featurevec}$. We can use 
		this quantity as a baseline for the average \gls{loss} of a trained \gls{model} $\learnthypothesis$. \label{fig_post_baseline}}
	\end{figure} },
    first={baseline},text={baseline}
}

\newglossaryentry{spectrogram}
{name={spectrogram},
	description={
		A\index{spectrogram} spectrogram represents the time-frequency distribution of the energy of a time signal $x(t)$.  
		Intuitively, it quantifies the amount of signal energy present within a specific time segment 
		$[t_{1},t_{2}] \subseteq \mathbb{R}$ and frequency interval $[f_{1},f_{2}]\subseteq \mathbb{R}$. 
		Formally, the spectrogram of a signal is defined as the squared magnitude of its 
		short-time Fourier transform (STFT) \cite{cohen1995time}.
        Figure \ref{fig:spectrogram} depicts a time signal along with its spectrogram. 
	\begin{figure}
		\centering
		\includegraphics[width=0.8\textwidth]{assets/spectrogram.png}
		\caption{Left: A time signal consisting of two modulated Gaussian pulses. Right: Intenstity 
		plot of the spectrogram.
		\label{fig:spectrogram}}
	\end{figure}
        The intensity plot of its spectrogram can serve as an image of a signal. A 
		simple recipe for audio signal classification is to feed this \emph{signal image} 
		into \gls{deepnet}s originally developed for image classification and object detection \cite{Li:2022aa}. 
		It is worth noting that, beyond the spectrogram, several alternative representations exist 
		for the time-frequency distribution of signal energy \cite{TimeFrequencyAnalysisBoashash,MallatBook}. 
		}, 
	first={spectrogram},text={spectrogram} 
}

\newglossaryentry{graphclustering}
{name={graph clustering},
	description={Graph clustering\index{graph clustering} aims at 
		clustering \gls{datapoint}s that are represented as the nodes 
		of a \gls{graph} $\graph$. The edges of $\graph$ represent 
		pair-wise similarities between \gls{datapoint}s. Sometimes we
		can quantity the extend of these similarities by an edge weight \cite{Luxburg2007,FlowSpecClustering2021}. }, 
	first={graph clustering},text={graph clustering} 
}

\newglossaryentry{specclustering}
{name={spectral clustering},
	description={Spectral clustering\index{spectral clustering} is a particular instance of 
		\gls{graphclustering}, i.e., it clusters \gls{datapoint}s 
		represented as the nodes $\nodeidx=1,\ldots,\nrnodes$ of a \gls{graph} $\graph$. 
		Spectral clustering uses the \gls{eigenvector}s of the \gls{LapMat} $\LapMat{\graph}$ 
		to construct \gls{featurevec}s $\featurevec^{(\nodeidx)} \in \mathbb{R}^{\nrfeatures}$ 
		for each node (\gls{datapoint}) $\nodeidx=1,\ldots,\nrnodes$. We can feed these \gls{featurevec}s 
		into \gls{euclidspace}-based \gls{clustering} methods such as \gls{kmeans} 
		or \gls{softclustering} via \gls{gmm}. Roughly speaking, the \gls{featurevec}s of nodes 
		belonging to a well-connected subset (or \gls{cluster}) of nodes in $\graph$, are located 
		nearby in the \gls{euclidspace} $\mathbb{R}^{\nrfeatures}$ (see Figure \ref{fig_lap_mtx_specclustering}). 
		\begin{figure}
			\begin{center}
				\begin{minipage}{0.4\textwidth}
			\begin{tikzpicture}
				\begin{scope}[every node/.style={circle, fill=black, inner sep=0pt, minimum size=0.3cm}]
					\node (1) at (0,0) {};
					\node (2) [below left=of 1, xshift=-0.2cm, yshift=-1cm] {};
					\node (3) [below right=of 1, xshift=0.2cm, yshift=-1cm] {};
					\node (4) [below=of 1, yshift=0.5cm] {}; % Isolated node
				\end{scope}
				\draw (1) -- (2);
				\draw (1) -- (3);
				\node[above=0.2cm] at (1) {$\nodeidx=1$};
				\node[left=0.3cm] at (2) {$2$};
				\node[right=0.3cm] at (3) {$3$};
				\node[below=0.2cm] at (4) {$4$};
			\end{tikzpicture}
				\end{minipage} 
				\hspace*{5mm}
				\begin{minipage}{0.4\textwidth}
					\begin{equation} 
						\LapMat{\graph}\!=\!
						\begin{pmatrix} 
							2 & -1 & -1 & 0 \\ 
							-1 & 1 & 0 & 0 \\  
							-1 & 0 & 1 & 0 \\ 
							0 & 0 & 0 & 0 
						\end{pmatrix}\!=\!\mathbf{V} {\bm \Lambda} \mathbf{V}^{T}  
						\nonumber
					\end{equation} 
				\end{minipage}
				\vspace*{20mm}\\
				  \begin{minipage}{0.4\textwidth}
				\begin{tikzpicture}[scale=3]
					\draw[->] (-0.2, 0) -- (1.2, 0) node[right] {$v^{(1)}_{\nodeidx}$};
					\draw[->] (0, -0.2) -- (0, 1.2) node[above] {$v^{(2)}_{\nodeidx}$};
					\filldraw[blue] (0.577, 0) circle (0.03cm) node[above right] {$\nodeidx=1,2,3$};
					\filldraw[blue] (0.577, 0) circle (0.03cm); % Second point overlaps
					\filldraw[blue] (0.577, 0) circle (0.03cm); % Third point overlaps
					\filldraw[red] (0, 1) circle (0.03cm) node[above right] {$4$};
				\end{tikzpicture}
				\end{minipage} 
    		\begin{minipage}{0.4\textwidth}
										\begin{align}
											& \mathbf{V} = \big( \vv^{(1)},\vv^{(2)},\vv^{(3)},\vv^{(4)} \big) \nonumber \\
											&	\mathbf{v}^{(1)}\!=\!\frac{1}{\sqrt{3}} \begin{pmatrix} 1 \\ 1 \\ 1 \\ 0 \end{pmatrix}, \,
												\mathbf{v}^{(2)}\!=\!\begin{pmatrix} 0 \\ 0 \\ 0 \\ 1 \end{pmatrix} \nonumber 
												\end{align}
				\end{minipage} 
				\caption{\label{fig_lap_mtx_specclustering} {\bf Top.} Left: An undirected \gls{graph} 
					$\graph$ with four nodes $\nodeidx=1,2,3,4$, each representing a \gls{datapoint}. Right: Laplacian matrix 
					$\LapMat{\graph}  \in \mathbb{R}^{4 \times 4}$ and its \gls{evd}. 
					{\bf Bottom.} Left: \Gls{scatterplot} of \gls{datapoint}s using the \gls{featurevec}s 
					$\featurevec^{(\nodeidx)} = \big( v^{(1)}_{\nodeidx},v^{(2)}_{\nodeidx} \big)^{T}$. 
					Right: Two \gls{eigenvector}s $\vv^{(1)},\vv^{(2)} \in \mathbb{R}^{\nrfeatures}$ 
					of the \gls{LapMat} $\LapMat{\graph}$ corresponding to the \gls{eigenvalue} $\lambda=0$. 
					} 
			\end{center}
		\end{figure}
	\newpage}, 
	first={spectral clustering},text={spectral clustering} 
}

\newglossaryentry{flowbasedclustering}
{name={flow-based clustering},
	description={Flow-basted clustering\index{flow-based clustering} groups the nodes 
		of an undirected graph by applying \gls{kmeans} clustering to node-wise feature 
		vectors. These \gls{feature} vectors are built from networks flows between 
		carefully selected source and destination nodes \cite{FlowSpecClustering2021}. }, 
	first={flow-based clustering},text={flow-based clustering} 
}

\newglossaryentry{esterr}
{name={estimation error},
	description={Consider\index{estimation error} \gls{datapoint}s with feature vectors $\featurevec$ and \gls{label} 
		$\truelabel$. In some applications we can model the relation between the \gls{feature}s and the \gls{label}
		of a \gls{datapoint} as $\truelabel = \bar{\hypothesis}(\featurevec) + \varepsilon$. Here, we 
		used some true underlying \gls{hypothesis} $\bar{\hypothesis}$ and a noise term $\varepsilon$ 
		which summarized any modelling or labelling errors. The estimation error incurred by a ML 
		method that learns a \gls{hypothesis} $\widehat{\hypothesis}$, e.g., using \gls{erm}, is defined as 
		$\widehat{\hypothesis}(\featurevec) - \bar{\hypothesis}(\featurevec)$, for some \gls{feature} vector. 
		For a parameterized \gls{hypospace}, consisting of \gls{hypothesis} maps that are determined by 
		a \gls{modelparams} $\weights$, we can define the estimation error as $\Delta \weights = \widehat{\weights} - \overline{\weights}$ \cite{kay,hastie01statisticallearning}.},
	first={estimation error},text={estimation error} 
}

\newglossaryentry{dob}
{name={degree of belonging},
	description={A\index{degree of belonging} number that indicates the extent by which a \gls{datapoint} 
		belongs to a \gls{cluster} \cite[Ch. 8]{MLBasics}. The degree of belonging can be 
		interpreted as a soft \gls{cluster} assignment. \Gls{softclustering} methods can 
		encode the degree of belonging by a real number in the interval $[0,1]$. 
		\Gls{hardclustering} is obtained as the extreme case when the degree of belonging 
		only takes on values $0$ or $1$.}, first={degree of belonging},text={degree of belonging} 
}

\newglossaryentry{msee}
{name={mean squared estimation error},
	description={Consider\index{mean squared estimation error} a ML method that 
		learns \gls{modelparams} $\widehat{\weights}$ based on some \gls{dataset} $\dataset$. 
		If we interpret the \gls{datapoint}s in $\dataset$ as \gls{iid} \gls{realization}s of a \gls{rv} $\datapoint$, 
		we define the \gls{esterr} $\Delta \weights \defeq \widehat{\weight} - \overline{\weights}$. 
		Here, $\overline{\weights}$ denotes the true \gls{modelparams} of the \gls{probdist} 
		of $\datapoint$.The mean squared estimation error is 
		defined as the \gls{expectation} $\expect \big\{ \big\| \Delta \weights \big\|^{2} \big\}$ of the 
		squared Euclidean norm of the \gls{esterr} \cite{LC,kay}.},
	first={mean squared estimation error (MSEE)},text={MSEE} 
}

\newglossaryentry{gtvmin}
{name={GTV minimization},
	description={GTV minimization\index{total variation minimization} is an instance of \gls{rerm} 
		using the \gls{gtv} of local \gls{modelparams} as a \gls{regularizer} \cite{ClusteredFLTVMinTSP}.},
	first={GTV minimization (GTVMin)},text={GTVMin} 
}

\newglossaryentry{regression}
{name={regression},
	description={Regression\index{regression} problems revolve around the problem of 
		predicting a numeric \gls{label} solely from the \gls{feature}s of a \gls{datapoint} \cite[Ch. 2]{MLBasics}.},
	first={regression},text={regression} 
}

\newglossaryentry{acc}
{name={accuracy},
	description={Consider\index{accuracy} \gls{datapoint}s characterized by \gls{feature}s $\featurevec \in \featurespace$ and 
		a categorical label $\truelabel$ which takes on values from a finite \gls{labelspace} $\labelspace$. The 
		accuracy of a \gls{hypothesis} $\hypothesis: \featurespace \rightarrow \labelspace$, when applied 
		to the \gls{datapoint}s in a \gls{dataset} $\dataset = \big\{ \big(\featurevec^{(1)}, \truelabel^{(1)} \big), \ldots, \big(\featurevec^{(\samplesize)},\truelabel^{(\samplesize)}\big) \big\}$ 
		is then defined as $1 - (1/\samplesize)\sum_{\sampleidx=1}^{\samplesize} \lossfunc{\big(\featurevec^{(\sampleidx)},\truelabel^{(\sampleidx)}\big)}{\hypothesis}$ using the \gls{zerooneloss}.},
	first={accuracy},text={accuracy} 
}

\newglossaryentry{expert}
{name={expert},
	description={ML\index{expert} aims to learn a \gls{hypothesis} $\hypothesis$ that accurately predicts the \gls{label} 
		of a \gls{datapoint} based on its \gls{feature}s. We measure the prediction error using 
		some \gls{lossfunc}. Ideally we want to find a \gls{hypothesis} that incurs minimum \gls{loss} 
		on any \gls{datapoint}. We can make this informal goal precise via the \gls{iidasspt} 
		and using the \gls{bayesrisk} as the \gls{baseline} for the (average) \gls{loss} of a \gls{hypothesis}. 
		An alternative approach to obtain a \gls{baseline} is to use the \gls{hypothesis} $\hypothesis'$ learnt 
		by an existing ML method. We refer to this \gls{hypothesis} $\hypothesis'$ as an expert \cite{PredictionLearningGames}. Regret minimization methods learn a \gls{hypothesis}
		that incurs a \gls{loss} comparable to the best expert \cite{PredictionLearningGames,HazanOCO}.},
	first={expert},text={expert} 
}

\newglossaryentry{nfl}
{name={networked federated learning},
	description={Networked\index{networked federated learning} federated learning refers 
		to methods that learn personalized models in a distributed fashion from \gls{localdataset}s 
		that are related by an intrinsic network structure.},
 first={networked federated learning (NFL)},text={NFL} 
}

\newglossaryentry{regret}
{name={regret},
	description={The regret\index{regret} of a \gls{hypothesis} $\hypothesis$ relative to 
		another \gls{hypothesis} $\hypothesis'$, which serves as a \gls{baseline}, 
		is the difference between the \gls{loss} incurred by $\hypothesis$ and the \gls{loss} 
		incurred by $\hypothesis'$ \cite{PredictionLearningGames}. 
		The \gls{baseline} \gls{hypothesis} $\hypothesis'$ is also referred to as an \gls{expert}.},
	first={regret},text={regret} 
}

\newglossaryentry{strcvx}
{name={strongly convex},
	description={A\index{strongly convex} continuously \gls{differentiable} real-valued 
		function $f(\featurevec)$ is strongly convex with coefficient $\sigma$ if $f(\vy) \geq f(\vx) + \nabla f(\vx)^{T} (\vy - \vx) + (\sigma/2) \normgeneric{\vy - \vx}{2}^{2}$ \cite{nesterov04},\cite[Sec. B.1.1.]{CvxAlgBertsekas}.},
	first={strongly convex},text={strongly convex} 
}

\newglossaryentry{differentiable}
{name={differentiable},
	description={A\index{differentiable} function real-valued function $f: \mathbb{R}^{\featuredim} \rightarrow \mathbb{R}$ 
		is differentiable if it can, at any point, be approximated locally by a linear function. The local linear approximation 
		at the point $\mathbf{x}$ is determined by the \gls{gradient} $\nabla f ( \mathbf{x})$ \cite{RudinBookPrinciplesMatheAnalysis}.},
	first={differentiable},text={differentiable} 
}

\newglossaryentry{gradient}
{name={gradient},
	description={For\index{gradient} a real-valued function $f: \mathbb{R}^{\featuredim} \rightarrow \mathbb{R}: \weights \mapsto f(\weights)$, 
	a vector $\vg$ such that $\lim_{\weights \rightarrow \weights'} \frac{f(\weights) - \big(f(\weights')+ \vg^{T} (\weights- \weights') \big) }{\| \weights-\weights'\|}=0$ 
	is referred to as the gradient of $f$ at $\weights'$. If such a vector exists it is 
	denoted $\nabla f(\weights')$ or $\nabla f(\weights)\big|_{\weights'}$ \cite{RudinBookPrinciplesMatheAnalysis}.},
	first={gradient},text={gradient} 
}

\newglossaryentry{subgradient}
{name={subgradient},
description={For\index{subgradient} a real-valued function $f: \mathbb{R}^{\featuredim} \rightarrow \mathbb{R}: \weights \mapsto f(\weights)$, 
		a vector $\va$ such that $f(\weights) \geq  f(\weights') +\big(\weights-\weights' \big)^{T} \va$ is 
		referred to as a subgradient of $f$ at $\weights'$ \cite{BertCvxAnalOpt,BertsekasNonLinProgr}.},
	first={subgradient},text={subgradient} 
}

\newglossaryentry{fedavg}
{name={federated averaging (FedAvg)},
	description={Federated\index{federated averaging (FedAvg)} averaging is an iterative \gls{fl} 
		algorithm that alternates between local model trainings and averaging the resulting local \gls{modelparams}. 
		Different variants of this algorithm are obtained by different techniques for the model training. 
		The authors of \cite{pmlr-v54-mcmahan17a} consider federated averaging methods where the 
		local model training is implemented by running several \gls{gd} steps.}, 
		first = {federated averaging (FedAvg)}, text={FedAvg} 
}

\newglossaryentry{relu}
{name={rectified linear unit (ReLU)},
	description={The\index{rectified linear unit (ReLU)} rectified linear unit (ReLU) is 
		a popular choice for the \gls{actfun} of a neuron within an \gls{ann}. It is defined 
		as $g(z) = \max\{0,z\}$ with $z$ being the weighted input of the artificial neuron.}, first = {rectified linear unit (ReLU)}, text={ReLU} 
}

\newglossaryentry{hypothesis}
{name={hypothesis},
	description={A\index{hypothesis} map (or function) $\hypothesis: \featurespace \rightarrow \labelspace$ from the 
		\gls{featurespace} $\featurespace$ to the \gls{labelspace} $\labelspace$. 
		Given a \gls{datapoint} with \gls{feature}s $\featurevec$, we use a hypothesis map $\hypothesis$
		to estimate (or approximate) the \gls{label} $\truelabel$ using the \gls{prediction}  
		$\hat{\truelabel} = \hypothesis(\featurevec)$. ML is all about learning (or finding) a 
		hypothesis map $\hypothesis$ such that $\truelabel \approx \hypothesis(\featurevec)$ 
		for any \gls{datapoint} (having \gls{feature}s $\featurevec$ and \gls{label} $\truelabel$).},
	first={hypothesis},text={hypothesis}  
}

\newglossaryentry{vcdim}
{name={Vapnik–Chervonenkis (VC) dimension},
	description={The\index{VC dimension} VC dimension of an infinite \gls{hypospace} is a widely-used measure 
		for its size. We refer to \cite{ShalevMLBook} for a precise definition of VC dimension 
		as well as a discussion of its basic properties and use in ML.},
	first={Vapnik–Chervonenkis (VC) dimension},text={VC dimension}  
}

\newglossaryentry{effdim}
{name={effective dimension},
	description={The\index{effective dimension} effective dimension $\effdim{\hypospace}$ of 
		an infinite \gls{hypospace} $\hypospace$ is a measure of its size. Loosely speaking, the 
		effective dimension is equal to the effective number of independent tunable parameters 
		of the model. These parameters might be the coefficients used in a linear map or the 
		weights and bias terms of an \gls{ann}.},
	first={effective dimension},text={effective dimension}  
}

\newglossaryentry{labelspace}
{name={label space},
	description={Consider\index{label space} a ML application that involves \gls{datapoint}s characterized by features 
		and labels. The \gls{label} space is constituted by all potential values that the \gls{label} 
		of a \gls{datapoint} can take on. Regression methods, aiming at predicting numeric \gls{label}s, often
		 use the \gls{label} space $\labelspace = \mathbb{R}$. Binary \gls{classification} methods use a label space 
 		that consists of two different elements, e.g., $\labelspace =\{-1,1\}$, $\labelspace=\{0,1\}$ 
		or $\labelspace = \{ \mbox{``cat image''}, \mbox{``no cat image''} \}$  }, first={label space},text={label space}  
}

\newglossaryentry{prediction}
{name={prediction},
	description={A\index{prediction} prediction is an estimate or approximation for some 
		quantity of interest. ML revolves around learning or finding a \gls{hypothesis} map $\hypothesis$ 
		that reads in the \gls{feature}s $\featurevec$ of a \gls{datapoint} and delivers a \gls{prediction} 
		$\widehat{\truelabel} \defeq \hypothesis(\featurevec)$ for its \gls{label} $\truelabel$. },
	first={prediction},text={prediction}  
}

\newglossaryentry{histogram}
{name={histogram},
	description={Consider\index{histogram} a \gls{dataset} $\dataset$ that consists of $\samplesize$ \gls{datapoint}s 
		$\datapoint^{(1)},\ldots,\datapoint^{(\samplesize)}$, each of them belonging to some 
		cell $[-U,U] \times \ldots \times [-U,U] \subseteq \mathbb{R}^{\featuredim}$ with side 
		length $U$. We partition this cell evenly into smaller elementary cells with side 
		length $\Delta$. The histogram of $\dataset$ assigns each elementary cell to 
		the corresponding fraction of \gls{datapoint}s in $\dataset$ that fall into this 
		elementary cell. 
	},
	first={histogram},text={histogram}  
}

\newglossaryentry{bootstrap}
{name={bootstrap},
	description={For\index{bootstrap}  the analysis of ML methods it is often useful to interpret 
		a given set of \gls{datapoint}s $\dataset = \big\{ \datapoint^{(1)},\ldots,\datapoint^{(\samplesize)}\big\}$ 
		as \gls{realization}s of \gls{iid} \gls{rv}s with a common \gls{probdist} $p(\datapoint)$. In general, we 
		do not know $p(\datapoint)$ exactly, but we need to estimate it. The bootstrap uses the 
		histogram of $\dataset$ as an estimator for the underlying \gls{probdist} $p(\datapoint)$. 
	},
	first={bootstrap},text={bootstrap}  
}

\newglossaryentry{featurespace}
{name={feature space},
	description={
		The\index{feature space} \gls{feature} space of a given ML application or method is 
		constituted by all potential values that the \gls{feature} vector of a \gls{datapoint} can 
		take on. A widely used choice for the feature space is the \gls{euclidspace} $\mathbb{R}^{\featuredim}$ 
		with dimension $\featurelen$ being the number of individual \gls{feature}s of a \gls{datapoint}.},
	first={feature space},text={feature space}  
}

\newglossaryentry{missingdata}
{name={missing data},
	description={Consider\index{missing data} a \gls{dataset} constituted by \gls{datapoint}s collected via 
		some physical device. Due to imperfections and failures, some of the \gls{feature} 
		or \gls{label} values of \gls{datapoint}s might be corrupted or simply \emph{missing}. 
		Data imputation aims at estimating these missing values \cite{Abayomi2008DiagnosticsFM}. 
		We can interpret data imputation as a ML problem where the \gls{label} of a \gls{datapoint} is 
		the value of the corrupted \gls{feature}. },
	first={missing data},text={missing data}  
}

\newglossaryentry{psd}
{name={positive semi-definite},
	description={A\index{positive semi-definite} (real-valued) symmetric matrix $\mQ = \mQ^{T} \in \mathbb{R}^{\featuredim \times \featuredim}$ 
	                      is referred to as positive semi-definite if $\featurevec^{T} \mQ \featurevec \geq 0$ for every vector $\featurevec \in \mathbb{R}^{\featuredim}$. 
	                      The property of being psd can be extended from matrices to (real-valued) symmetric \gls{kernel} maps 
	                     $\kernel: \featurespace \times \featurespace \rightarrow \mathbb{R}$ (with $\kernel(\featurevec,\featurevec') = \kernel(\featurevec',\featurevec)$ 
	                      as follows: For any finite 
	                      set of \gls{feature} vectors $\featurevec^{(1)},\dots,\featurevec^{(\samplesize)}$, 
	                      the resulting matrix $\mQ \in \mathbb{R}^{\samplesize \times \samplesize}$ with 
	                      entries $Q_{\sampleidx,\sampleidx'} = \kernelmap{\featurevec^{(\sampleidx)}}{\featurevec^{(\sampleidx')}}$ is psd \cite{LearningKernelsBook}.},
	first={positive semi-definite (psd)},text={psd}  
}

\newglossaryentry{feature}
{name={feature},
	description={A\index{feature} feature of a \gls{datapoint} is one of its properties that can be 
		measured or computed easily without the need for human supervision. For example, if a \gls{datapoint} 
		is a digital image (e.g,, stored in as a jpeg file), then we could use the red-green-blue intensities 
		of its pixels as features. Domain-specific synonyms for the term feature are \emph{covariate}, \emph{explanatory variable}, 
		\emph{independent variable}, \emph{input (variable)}, \emph{predictor (variable)} or \emph{regressor} \cite{Gujarati2021,Dodge2003,Everitt2022}. 
		}, first={feature},text={feature}  
}

\newglossaryentry{featurevec}
{name={feature vector},
	description={A\index{feature vector} vector $\vx = \big(x_{1},\ldots,x_{\nrfeatures}\big)^{T}$ 
	whose entries are individual \gls{feature}s $x_{1},\ldots,x_{\nrfeatures}$. Many ML methods 
	use feature vectors that belong to some finite-dimensional \gls{euclidspace} $\mathbb{R}^{\nrfeatures}$. 
	However, for some ML methods it is more convenient to work with feature vectors that 
	belong to an infinite-dimensional vector space (see, e.g., \gls{kernelmethod}). 
		}, first={feature vector},text={feature vector}  
}

\newglossaryentry{label}
{name={label},
	description={A\index{label} higher-level fact or quantity of interest associated with a \gls{datapoint}. 
		For example, if the \gls{datapoint} is an image, the label could indicate whether the 
		image contains a cat or not. Synonyms for label, commonly used in specific domains, 
		include \emph{response variable}, \emph{output variable}, and \emph{target} \cite{Gujarati2021,Dodge2003,Everitt2022}.
 },
	first={label},text={label}  
}

\newglossaryentry{data}
{name={data},
	 description={See\index{data} \gls{dataset}.},
	text={data}
}

\newglossaryentry{dataset}
{name={dataset},
	description={With\index{dataset} a slight abuse of language we use the term \emph{dataset} 
		or \emph{set of \gls{datapoint}s} 
		to refer to an indexed list of \gls{datapoint}s $\datapoint^{(1)},\datapoint^{(2)},\ldots$. 
		Thus, there is a first \gls{datapoint} $\datapoint^{(1)}$, a second \gls{datapoint} $\datapoint^{(2)}$ and so on. 
		Strictly speaking, a dataset is a list and not a set \cite{HalmosSet}. We need to keep 
		track of the order of \gls{datapoint}s in order to cope with several \gls{datapoint}s 
		having the same \gls{feature}s and \gls{label}s. Database theory studies formal languages 
		for defining, structuring, and reasoning about datasets \cite{silberschatz2019database}.},first={dataset},text={dataset}  
}

\newglossaryentry{predictor}
{name={predictor},
	description={A\index{predictor} predictor is a real-valued \gls{hypothesis} map. 
		Given a \gls{datapoint} with \gls{feature}s $\featurevec$, the value 
		$\hypothesis(\featurevec) \in \mathbb{R}$ is used as a \gls{prediction} for the true 
		numeric label $\truelabel \in \mathbb{R}$ of the \gls{datapoint}. },first={predictor},text={predictor}  
}

\newglossaryentry{labeled datapoint}
{name={labeled datapoint},
 description={A\index{labeled data} \gls{datapoint} whose label is known or has been determined 
 	by some means which might involve human experts.},
 first={labeled datapoint},text={labeled datapoint}  
}

\newglossaryentry{rv}
{name={random variable (RV)},
 description={A\index{random variable (RV)} random\index{probability space} 
 		variable is a mapping from a probability space $\mathcal{P}$ to a value space \cite{BillingsleyProbMeasure}. 
 	The probability space, whose elements are elementary events, is equipped with a probability 
 	 measure that assigns a probability to subsets of $\mathcal{P}$. A binary random variable maps elementary events 
 	to a set containing two different values, e.g., $\{-1,1\}$ or $\{ \mbox{cat}, \mbox{no cat} \}$. 
 	A real-valued random variable maps elementary events to real numbers $\mathbb{R}$. 
 	A vector-valued random variable maps elementary events to the \gls{euclidspace} $\mathbb{R}^{\featuredim}$. 
 	Probability theory uses the concept of measurable spaces to rigorously define and study the properties of (large) 
 	collections of random variables \cite{GrayProbBook,BillingsleyProbMeasure}.}, first={RV},text={RV}  }

\newglossaryentry{realization}
{name={realization},
	description={Consider\index{realization} a \gls{rv} $x$ which maps each element (outcome, or elementary event) $\omega \in \mathcal{P}$ of a 
		probability space $\mathcal{P}$ to an element $a$ of a measurable space $\mathcal{N}$ \cite{BillingsleyProbMeasure,RudinBookPrinciplesMatheAnalysis,HalmosMeasure}. 
		A realization of $x$ is any element $a' \in \mathcal{N}$ such that there is 
		an element $\omega' \in \mathcal{P}$ with $x(\omega') = a'$.  }, first={realization},text={realization}  }

\newglossaryentry{trainset}
{name={training set},
description={A\index{training set} \gls{dataset} $\dataset$, constituted by some \gls{datapoint}s used in \gls{erm} 
	to learn a \gls{hypothesis} $\learnthypothesis$. The average \gls{loss} of $\learnthypothesis$ on the 
	training set is referred to as the \gls{trainerr}. The comparison of the \gls{trainerr} with the 
	\gls{valerr} of $\learnthypothesis$ allows to diagnose the ML method and informs how to improve 
	them (e.g., using a different \gls{hypospace} or collecting more \gls{datapoint}s) \cite[Sec. 6.6.]{MLBasics}.},first={training set},text={training set}  
}

\newglossaryentry{netmodel}
{name={networked model},
  description={A\index{networked model} networked model over a \gls{empgraph} $\graph = \pair{\nodes}{\edges}$ assigns 
   a \gls{localmodel} (\gls{hypospace}) to each node $\nodeidx \in \nodes$ of the \gls{empgraph} $\graph$.}, 
   first={networked model},text={networked model}  
}

\newglossaryentry{batch}
{
	name={batch},
	description={In\index{batch} the context of \gls{stochGD}, a batch refers to a randomly 
	chosen subset of the overall \gls{trainset}. We use the \gls{datapoint}s in this subset 
	to estimte the \gls{gradient} of \gls{trainerr} and, in turn, to update the \gls{modelparams}.}, 
	first={batch},text={batch}  
}

\newglossaryentry{netdata}
{
	name={networked data},
	description={Networked\index{networked data} data consists of \gls{localdataset}s 
	that are related by some notion of pair-wise similarity. We can represent networked 
	data using a \gls{graph} whose nodes carry \gls{localdataset}s and edges encode 
	pairwise similarities. One example for \gls{netdata} arises in \gls{fl} applications 
	where \gls{localdataset}s are generated by spatially distributed devices.}, 
	first={networked data},text={networked data}  
}

\newglossaryentry{trainerr}
{
	name={training error},
	description={The\index{training error} average \gls{loss} of a \gls{hypothesis} when 
		predicting the \gls{label}s of \gls{datapoint}s in a \gls{trainset}. We sometimes refer 
		by training error also the minimum average \gls{loss} incurred on the \gls{trainset} 
		by the optimal \gls{hypothesis} from a \gls{hypospace}.},first={training error},text={training error}  
}

\newglossaryentry{datapoint}
{name={data point},
description={A\index{data point} \gls{datapoint} is any object that conveys information \cite{coverthomas}. Data points might be 
		students, radio signals, trees, forests, images, \gls{rv}s, real numbers or proteins. We characterize data points 
		using two types of properties. One type of property is referred to as a \gls{feature}. \Gls{feature}s are properties of a 
		\gls{datapoint} that can be measured or computed in an automated fashion. 
		A different kind of property is referred to as \gls{label}s. The \gls{label} of 
		a \gls{datapoint} represents some higher-level fact (or quantity of interest). In 
		contrast to \gls{feature}s, determining the \gls{label} of a \gls{datapoint} typically 
		requires human experts (domain experts). Roughly speaking, ML aims to predict 
		the \gls{label} of a \gls{datapoint} based solely on its \gls{feature}s. 
		}, first={data point},text={data point}  
}

\newglossaryentry{valerr}
{name={validation error},
 description={Consider\index{validation error} a \gls{hypothesis} $\learnthypothesis$ which is 
 	obtained by some ML method, e.g., using \gls{erm} on a \gls{trainset}. The average \gls{loss} 
 	of $\learnthypothesis$ on a \gls{valset}, which is different from the \gls{trainset}, is referred 
 	to as the validation error.},first={validation error},text={validation error}  
}

\newglossaryentry{validation} 
{name={validation},
	description={Consider\index{validation} a \gls{hypothesis} $\learnthypothesis$ that has been 
		learnt via some ML method, e.g., by solving \gls{erm} on a \gls{trainset} $\dataset$. 
		Validation refers to the practice of evaluating the \gls{loss} incurred by 
		\gls{hypothesis} $\learnthypothesis$ on a \gls{valset} that consists of 
		\gls{datapoint}s that are not contained in the \gls{trainset} $\dataset$. },first={validation},text={validation}  
}

\newglossaryentry{quadfunc}
{name={quadratic function},
	description={A\index{quadratic function} quadratic function $f(\weights)$, reading in a 
		vector $\weights \in \mathbb{R}^{\nrfeatures}$ as its argument, is such that $$f(\weights) =  \weights^{T} \mathbf{Q} \mathbf{w} + \mathbf{q}^{T} \weights+a,$$ with some matrix $\mQ \in \mathbb{R}^{\nrfeatures \times \nrfeatures}$, 
		vector $\vq \in \mathbb{R}^{\nrfeatures}$ and scalar $a \in \mathbb{R}$.  },first={quadratic function},text={quadratic function}  
}

\newglossaryentry{valset}
{name={validation set},
  description={A\index{validation set} set of \gls{datapoint}s used to estimate 
  	the \gls{risk} of a \gls{hypothesis} $\learnthypothesis$ that has been learnt by some 
  	ML method (e.g., solving \gls{erm}). The average \gls{loss} of $\learnthypothesis$ 
  	on the validation set is referred to as the validation error and can be used to diagnose a 
  	ML method (see \cite[Sec. 6.6.]{MLBasics}). The comparison between \gls{trainerr} 
  	and \gls{valerr} can inform directions for improvements of the ML method (such as 
  	using a different \gls{hypospace}).},first={validation set},text={validation set}  
}

\newglossaryentry{testset}
{name={test set},
	description={A\index{test set} set of \gls{datapoint}s that have neither 
		been used to train a \gls{model}, e.g., via \gls{erm}, nor in a \gls{valset} 
		to choose between different \gls{model}s.},first={test set},text={test set}  
}

\newglossaryentry{modelsel}
{name={model selection},
	description={In\index{model selection} ML, model selection refers to the 
		process of choosing between different candidate \gls{model}s. In its most 
		basic form, \gls{model} selection amounts to (i) training each candidate \gls{model}, 
		(ii) computing the \gls{valerr} for each trained \gls{model}, (iii) choosing the \gls{model} 
		with smallest \gls{valerr} \cite[Ch. 6]{MLBasics}. },first={model selection},text={model selection}  
}

\newglossaryentry{linclass}{name={linear classifier}, description={
	    Consider\index{linear classifier} \gls{datapoint}s characterized by numeric \gls{feature}s $\featurevec \in \mathbb{R}^{\nrfeatures}$ 
	    and a \gls{label} $\truelabel \in \labelspace$ from some finite \gls{labelspace} $\labelspace$. 
		A linear \gls{classifier} characterized by having \gls{decisionregion}s separated 
		by hyperplanes in the \gls{euclidspace} $\mathbb{R}^{\featuredim}$ \cite[Ch. 2]{MLBasics}.},first={linear classifier},text={linear classifier} }

\newglossaryentry{erm}{name={empirical risk minimization}, description={Empirical risk 
		minimization\index{empirical risk minimization} is the optimization problem of finding 
		a \gls{hypothesis} (out of a \gls{model}) with minimum average \gls{loss} (or \gls{emprisk}) on a given \gls{dataset} 
		$\dataset$ (the \gls{trainset}). Many ML methods are obtained from 
		\gls{emprisk} via specific design choices for the \gls{dataset}, \gls{model} and \gls{loss} \cite[Ch. 3]{MLBasics}.},
	first={empirical risk minimization (ERM)},text={ERM} }

\newglossaryentry{multilabelclass}{name={multi-label classification}, description={Multi-label 
		classification\index{multi-label classification} problems and methods use \gls{datapoint}s 
		that are characterized by several \gls{label}s. As an example, consider a \gls{datapoint} 
		representing a picture with one binary \gls{label} indicating the presence of a human 
		in this picture and another \gls{label} indicating the presence of a car.},
	    first={multi-label classification},text={multi-label classification} }

\newglossaryentry{ssl}{name={semi-supervised learning}, description={Semi-supervised\index{semi-supervised learning} 
		learning methods use unlabeled \gls{datapoint}s to support the learning of a \gls{hypothesis} 
		from labeled \gls{datapoint}s \cite{SemiSupervisedBook}. This approach is particularly useful 
		for ML applications that offer a large amount of unlabeled \gls{datapoint}s, but only a limited 
		number of  labeled \gls{datapoint}s.}, 
		first={semi-supervised learning (SSL)},text={SSL} }

\newglossaryentry{objfunc}{name={objective function}, description={An\index{objective function} 
		objective function is a map that assigns each value of an optimization variable, such 
		as the \gls{modelparams} $\weights$ of a \gls{hypothesis} $\hypothesis^{(\weights)}$, to 
		an objective value $f(\weights)$. The objective value $f(\weights)$ could be the 
		\gls{risk} or the \gls{emprisk} of a \gls{hypothesis} $\hypothesis^{(\weights)}$.},first={objective function},text={objective function} }
	
\newglossaryentry{regularizer}{name={regularizer}, description={A regularizer\index{regularizer} 
		assigns each \gls{hypothesis} $\hypothesis$ from a \gls{hypospace} $\hypospace$ a quantitative 
		measure $\regularizer{\hypothesis}$ for how much its prediction error on a \gls{trainset} might 
		differ from its prediction errors on \gls{datapoint}s outside the \gls{trainset}. \Gls{ridgeregression} 
		uses the regularizer $\regularizer{\hypothesis} \defeq \normgeneric{\weights}{2}^{2}$ for linear \gls{hypothesis} maps $\hypothesis^{(\weights)}(\featurevec) \defeq \weights^{T} \featurevec$ \cite[Ch. 3]{MLBasics}. 
		The \gls{lasso} uses the regularizer $\regularizer{\hypothesis} \defeq \normgeneric{\weights}{1}$ 
		for linear \gls{hypothesis} maps $\hypothesis^{(\weights)}(\featurevec) \defeq \weights^{T} \featurevec$ \cite[Ch. 3]{MLBasics}. },first={regularizer},text={regularizer} }

\newglossaryentry{regularization}{name={regularization}, description={
		A\index{regularization} key challenge of modern ML applications is that they often 
		use large \gls{model}s, having an \gls{effdim} in the order of billions. 
		Using basic \gls{erm}-based methods to train a high-dimensional \gls{model} 
		is prone to \gls{overfitting}: the learn \gls{hypothesis} performs well on the \gls{trainset} 
		but poorly outside the \gls{trainset}. Regularization refers to modifications of a given instance 
		of \gls{erm} in order to avoid \gls{overfitting}, i.e., to ensure the learnt \gls{hypothesis} performs 
		not much worse outside the \gls{trainset}. There are three routes for implementing 
		regularization: 
		\begin{itemize} 
			\item {\bf \Gls{model} pruning.} We prune the original \gls{model} $\hypospace$ to obtain a 
			smaller \gls{model} $\hypospace'$. For a parametrized \gls{model}, the pruning can be 
			implemented by including constraints on the \gls{modelparams} (such as $w_{1} \in [0.4,0.6]$ for 
			the weight of \gls{feature} $x_{1}$ in \gls{linreg}).
			\item {\bf \Gls{loss} penalization.} We modify the \gls{objfunc} of \gls{erm} by adding a 
			penalty term to the \gls{trainerr}. The penalty term estimates how much larger the expected \gls{loss} (\gls{risk}) 
			is compared to the average \gls{loss} on the \gls{trainset}. 
			\item {\bf \Gls{dataaug}.} We can enlarge the \gls{trainset} $\dataset$ by adding 
			perturbed copies of the original \gls{datapoint}s in $\dataset$. One example for such 
			a perturbation is to add the \gls{realization} of a \gls{rv} to the \gls{featurevec} 
			of a \gls{datapoint}. 
		\end{itemize} 
		Figure \ref{fig_equiv_dataaug_penal} illustrates the above routs to regularization. These 
		routes are often equivalent: \gls{dataaug} using \gls{gaussrv}s 
		to perturb the \gls{featurevec}s in the \gls{trainset} of \gls{linreg} 
		has the same effect as adding the penalty 
		$\lambda \normgeneric{\weights}{2}^2$ to the \gls{trainerr} (which is nothing but \gls{ridgeregression}). 
		\begin{figure}
			\begin{center} 
				\begin{tikzpicture}[scale = 1]
					\draw[->, very thick] (0,0.5) -- (7.7,0.5) node[right] {feature $\feature$};       % X-axis
					\draw[->, very thick] (0.5,0) -- (0.5,4.2) node[above] {label $\truelabel$};   % Y-axis
					\draw[color=black, thick, dashed, domain = -1: 6.2, variable = \x]  plot ({\x},{\x*0.4 + 2.0}) ;     
					\draw[color=black, thick, dashed, domain = -1: 6.2, variable = \x]  plot ({\x},{\x*0.6 + 2.0}) ;     
					\draw[blue, thick] (5, 4.5) ellipse [x radius=0.2cm, y radius=1cm];
					\node at (5, 5.8) [text=black, font=\small] {$\{ \hypothesis: \hypothesis(x)\!=\!w_{1}x\!+\!w_{0}; w_{1} \in [0.4,0.6]\}$};
					\node at (6.7,4.5) {$\hypothesis(\feature)$};    
					\coordinate (l1)   at (1.2, 2.48);
					\coordinate (l2) at (1.4, 2.56);
					\coordinate (l3)   at (1.7,  2.68);
					\coordinate (l4)   at (2.2, 2.2*0.4+2.0);
					\coordinate (l5) at (2.4, 2.4*0.4+2.0);
					\coordinate (l6)   at (2.7,  2.7*0.4+2.0);
					\coordinate (l7)   at (3.9,  3.9*0.4+2.0);
					\coordinate (l8) at (4.2, 4.2*0.4+2.0);
					\coordinate (l9)   at (4.5,  4.5*0.4+2.0);
					\coordinate (n1)   at (1.2, 1.8);
					\coordinate (n2) at (1.4, 1.8);
					\coordinate (n3)   at (1.7,  1.8);
					\coordinate (n4)   at (2.2, 3.8);
					\coordinate (n5) at (2.4, 3.8);
					\coordinate (n6)   at (2.7,  3.8);
					\coordinate (n7)   at (3.9, 2.6);
					\coordinate (n8) at (4.2, 2.6);
					\coordinate (n9)   at (4.5,  2.6);
					\node at (n1)  [circle,draw,fill=red,minimum size=6pt,scale=0.6, name=c1] {};
					\node at (n2)  [circle,draw,fill=blue,minimum size=6pt, scale=0.6, name=c2] {};
					\node at (n3)  [circle,draw,fill=red,minimum size=6pt,scale=0.6,  name=c3] {};
					\node at (n4)  [circle,draw,fill=red,minimum size=12pt, scale=0.6, name=c4] {};  
					\node at (n5)  [circle,draw,fill=blue,minimum size=12pt,scale=0.6,  name=c5] {};
					\node at (n6)  [circle,draw,fill=red,minimum size=12pt, scale=0.6, name=c6] {};  
					\node at (n7)  [circle,draw,fill=red,minimum size=12pt,scale=0.6,  name=c7] {};
					\node at (n8)  [circle,draw,fill=blue,minimum size=12pt, scale=0.6, name=c8] {};
					\node at (n9)  [circle,draw,fill=red,minimum size=12pt, scale=0.6, name=c9] {};
					\draw [<->] ($ (n7) + (0,-0.3) $)  --  ($ (n9) + (0,-0.3) $) node [pos=0.4, below] {$\sqrt{\regparam}$}; ; 
					\draw[<->, color=red, thick] (l1) -- (c1);  
					\draw[<->, color=blue, thick] (l2) -- (c2);  
					\draw[<->, color=red, thick] (l3) -- (c3);  
					\draw[<->, color=red, thick] (l4) -- (c4);  
					\draw[<->, color=blue, thick] (l5) -- (c5);  
					\draw[<->, color=red, thick] (l6) -- (c6);  
					\draw[<->, color=red, thick] (l7) -- (c7);  
					\draw[<->, color=blue, thick] (l8) -- (c8);  
					\draw[<->, color=red, thick] (l9) -- (c9);  
					\draw[fill=blue] (6.2, 3.7)  circle (0.1cm) node [black,xshift=2.3cm] {original \gls{trainset} $\dataset$};
					\draw[fill=red] (6.2, 3.2)  circle (0.1cm) node [black,xshift=1.3cm] {augmented};
					\node at (4.6,1.2)  [minimum size=12pt, font=\fontsize{12}{0}\selectfont, text=blue] {$\frac{1}{\samplesize} \sum_{\sampleidx=1}^\samplesize \lossfunc{\pair{\featurevec^{(\sampleidx)}}{ \truelabel^{(\sampleidx)}}}{\hypothesis}$};
					\node at (7.8,1.2)  [minimum size=12pt, font=\fontsize{12}{0}\selectfont, text=red] {$+\regparam \regularizer{\hypothesis}$};
				\end{tikzpicture}
				\caption{Three approaches to regularization: \gls{dataaug}, \gls{loss} penalization and \gls{model} 
				pruning (via constraints on \gls{modelparams}). \label{fig_equiv_dataaug_penal} }
			\end{center}
		\end{figure} 
		\newpage
		},first={regularization},text={regularization} }

\newglossaryentry{rerm}{name={regularized empirical risk minimization (RERM)}, 
	description={Synonym\index{regularized empirical risk minimization (RERM)} for \gls{srm}.},
	first={regularized empirical risk minimization (RERM)},text={RERM} }

\newglossaryentry{generalization}{name={generalization}, 
	description={Many\index{generalization} current ML (and AI) methods are an instance 
		of \gls{erm}: At their core, they train a \gls{model} (learn a \gls{hypothesis} 
		$\learnthypothesis \in \hypospace$) by minimizing the average \gls{loss} (or \gls{emprisk}) on some 
		\gls{datapoint}s $\vz^{(1)},\ldots,\vz^{(\samplesize)}$, which serve as a \gls{trainset} $\trainset$. 
		Generalization refers to a ML method's ability to perform well outside the \gls{trainset}. 
		Any mathematical theory of generalization needs some mathematical concept for the 
		\emph{outside the \gls{trainset}}. For example, statistical learning theory uses a 
		\gls{probmodel} such as the \gls{iidasspt} for data generation: the \gls{datapoint}s in 
		the \gls{trainset} are \gls{iid} \gls{realization}s of some underlying \gls{probdist} $p(\vz)$. 
		Using  a \gls{probmodel} allows to explore the \emph{outside of the \gls{trainset}} by 
		drawing additional \gls{iid} \gls{realization}s from $p(\vz)$. Moreover, using the \gls{iidasspt} 
		allows to define the risk of a trained \gls{model} $\learnthypothesis \in \hypospace$ as 
		the expected \gls{loss} $\risk{\learnthypothesis}$. What is more, we can use concentration 
		bounds or convergence results for sequences of \gls{iid} \gls{rv}s to bound the deviation 
		between the \gls{emprisk} $\emprisk{\learnthypothesis}{\trainset}$ of a trained model and 
		its \gls{risk} \cite{ShalevMLBook}. It is possible to study generalization also without using 
		\gls{probmodel}s. For example, we could use (deterministic) 
	    perturbations of the \gls{datapoint}s in the \gls{trainset} to study its \emph{outside}. 
	    In general, we would like the trained \gls{model} to be robust, i.e., its \gls{prediction}s 
	    should not change too much for small perturbations of a \gls{datapoint}. Consider a trained \gls{model} for detecting 
	    an object in a smartphone snapshot. The detection result should not change if we mask a 
	    small number of randomly chosen pixels in the image \cite{OnePixelAttack}. 
		  \begin{figure}
		                   	\centering
		                   	\begin{tikzpicture}[scale=0.8]
							   \draw[lightblue, fill=lightblue, opacity=0.5] (3, 2) ellipse (6cm and 2cm);
								\node[black] at (6, 3) {$p(z)$};
		                   		\fill[blue] (1, 3) circle (4pt) node[below, xshift=0pt, yshift=0pt] {$\datapoint^{(1)}$};
		                   		\fill[blue] (5, 1) circle (4pt) node[below] {$\datapoint^{(2)}$};
		                   		\fill[blue] (1.6, 3) circle (3pt);
		                   		\fill[blue] (0.4, 3) circle (3pt);
		                   		\draw[<->, thin] (1, 3) -- (1.6, 3);
		                   		\draw[<->, thin] (1, 3) -- (0.4, 3);
		                   		\fill[blue] (5.6, 1) circle (3pt);
		                   		\fill[blue] (4.4, 1) circle (3pt);
		                   		\draw[<->, thin] (5, 1) -- (5.6, 1);
		                   		\draw[<->, thin] (5, 1) -- (4.4, 1);
		                   		\draw[black, thick, domain=0:6, smooth] plot (\x, {- 1*\x + 5});
		                   		\node[black] at (3, 2.5) [right] {$\learnthypothesis$};
		                   	\end{tikzpicture}
		                   	\caption{Two \gls{datapoint}s $\datapoint^{(1)},\datapoint^{(2)}$ that are used as a \gls{trainset} 
		                   		to learn a \gls{hypothesis} $\learnthypothesis$ via \gls{erm}. We can evaluate $\learnthypothesis$ 
		                   		\emph{outside} $\trainset$ either by an \gls{iidasspt} with underlying \gls{probdist} $p(\datapoint)$ 
		                   		or by perturbing the \gls{datapoint}s.}
		                   	\label{fig:polynomial_fit}
		                   \end{figure}
		                   \newpage
		},
	first={generalization},text={generalization} }

\newglossaryentry{gtv}{name={generalized total variation}, description={Generalized\index{generalized total variation} 
		total variation measures the changes of vector-valued node attributes over a 
		weighted undirected \gls{graph}.},first={generalized total variation (GTV)},text={GTV} }
	
\newglossaryentry{srm}{name={structural risk minimization}, description={Structural\index{structural risk minimization} 
		risk minimization is the problem of finding the \gls{hypothesis} that optimally 
		balances the average \gls{loss} (or \gls{emprisk}) on a \gls{trainset} with a 
		\gls{regularization} term. The \gls{regularization} term penalizes a \gls{hypothesis}
		that is not robust against (small) perturbations of the \gls{datapoint}s in the \gls{trainset}.},first={structural risk minimization (SRM)},text={SRM} }

\newglossaryentry{datapoisoning}{name={data poisoning}, description={Data\index{data poisoning} 
		poisoning refers to the intentional manipulation (or fabrication) of \gls{datapoint}s to 
		steer the training of a ML model \cite{Liu2021,PoisonGAN}. The protection against 
		data poisoning is particularly important in distributed ML applications where \gls{dataset}s are de-centralized.},first={data poisoning},text={data poisoning} }

\newglossaryentry{backdoor}{name={backdoor}, description={A\index{backdoor} backdoor attack refers 
		to the intentional manipulation of the training process underlying a ML method. This manipulation 
		can be implemented by perturbing the \gls{trainset} (data poisoning) or the 
		optimization algorithm used by an \gls{erm}-based method. The goal of a 
		backdoor attack is to nudge the learnt \gls{hypothesis} $\learnthypothesis$ 
		towards specific \gls{prediction}s for a certain range of \gls{feature} values. This range of \gls{feature} 
		values serves as a key (or trigger) to unlock a \emph{backdoor} in the sense of 
		delivering anomalous \gls{prediction}s. The key $\featurevec$ and the corresponding 
		anomolous \gls{prediction} $\learnthypothesis(\featurevec)$ are only known to the attacker.},
	first={backdoor},text={backdoor} }

\newglossaryentry{clustasspt}{name={clustering assumption}, description={The\index{clustering assumption} 
		clustering assumption postulates that \gls{datapoint}s in a \gls{dataset} form a (small) number of 
		groups or clusters. \Gls{datapoint}s in the same \gls{cluster} are more similar with each 
		other than with those outside the cluster \cite{SemiSupervisedBook}. We obtain different 
		clustering methods by using different notions of similarity between \gls{datapoint}s.},first={clustering assumption},text={clustering assumption} }
	
\newglossaryentry{dosattack}{name={denial-of-service attack}, description={A\index{denial-of-service attack} 
		denial-of-service attack aims (e.g., via \gls{datapoisoning}) to steer the training of a \gls{model} 
		such that it performs poorly for typical \gls{datapoint}s},
	first={denial-of-service attack},text={denial-of-service attack} }

\newglossaryentry{netexpfam}{name={networked exponential families}, 
	description={A\index{networked exponential families} collection of exponential 
		families, each of them assigned to a node of a \gls{empgraph}. The \gls{modelparams} are coupled 
	   via the network structure by requiring them to have a small \gls{gtv} \cite{JungNetExp2020}. },first={networked exponential family (nExpFam)},text={nExpFam} }

\newglossaryentry{scatterplot}{name={scatterplot}, description={A\index{scatterplot} 
		visualization technique that depicts \gls{datapoint}s by markers in a two-dimensional plane. 
		\begin{figure}[htbp]
			\begin{center}
				\begin{tikzpicture}[scale=1]
					\tikzset{x=2cm,y=2cm,every path/.style={>=latex},node style/.style={circle,draw}}
					\begin{axis}[axis x line=none,
						axis y line=none,
						ylabel near ticks,
						xlabel near ticks,
						enlarge y limits=true,
						xmin=-5, xmax=30,
						ymin=-5, ymax=30,
						width=6cm, height=6cm ]
						\addplot[only marks] table [x=mintmp, y=maxtmp, col sep = semicolon] {FMIData1.csv};
						\node at (axis cs:26,2) [anchor=west] {$\feature$};
						\node at (axis cs:0,30) [anchor=west] {$\truelabel$};
						\draw[->] (axis cs:-5,0) -- (axis cs:30,0);
						\draw[->] (axis cs:0,-5) -- (axis cs:0,30);
					\end{axis}
				\end{tikzpicture}
				\vspace*{-14mm}
			\end{center}
			\caption{A scatterplot of \gls{datapoint}s that represent daily weather conditions in Finland. 
				Each \gls{datapoint} is characterized by its minimum daytime temperature $\feature$ 
				as \gls{feature} and its maximum daytime temperature $\truelabel$ as the \gls{label}. 
				The temperatures have been measured at the \gls{fmi} weather station \emph{Helsinki Kaisaniemi} 
				during 1.9.2024 - 28.10.2024.}
			\label{fig_scatterplot_temp_FMI}
			\vspace*{-3mm}
			\end{figure}
		},first={scatterplot},text={scatterplot} }

\newglossaryentry{stepsize}{name={step size}, description={
		See\index{step size} \gls{learnrate}.}, 
	first={step size},text={step size} }

\newglossaryentry{learnrate}{name={learning rate}, description={Consider\index{learning rate} 
		an iterative method for finding or learning a useful \gls{hypothesis} $\hypothesis \in \hypospace$. 
		Such an iterative method repeats similar computational (update) steps that adjust or 
		modify the current \gls{hypothesis} to obtain an improved \gls{hypothesis}. A prime example of 
		such an iterative learning method is \gls{gd} and its variants such as \gls{stochGD} or \gls{projgd}. 
		We refer by learning rate to a parameter of an iterative learning 
		method that controls the extent by which the current \gls{hypothesis} 
		can be modified during a single iteration. A prime example of such a parameter is the 
				step size used in \gls{gd} \cite[Ch. 5]{MLBasics}.},
	first={learning rate},text={learning rate} }

\newglossaryentry{featuremap}{name={feature map}, description={A\index{feature map} map 
		that transforms the original \gls{feature}s of a \gls{datapoint} into new \gls{feature}s. The 
		so-obtained new \gls{feature}s might be preferable over the original \gls{feature}s for 
		several reasons. For example, the arrangement of \gls{datapoint}s 
		 might become simpler (of \emph{more linear}) in the new \gls{featurespace}, allowing to use linear models 
		 in the new \gls{feature}s. This idea is a main driver for the development of \gls{kernel} methods \cite{LearningKernelsBook}. 
		 Moreover, the hidden layers of a \gls{deepnet} can be interpreted as a trainable feature map followed 
		 by a \gls{linmodel} in the form of the output layer. Another reason for learning a \gls{featuremap} 
		 could be that learning a small number of new \gls{feature}s helps to avoid \gls{overfitting} and 
		 ensure interpretability \cite{Ribeiro2016}. The special case of a \gls{feature} map delivering 
		 two numeric \gls{feature}s is particularly useful for data visualization. Indeed, we can depict 
		 \gls{datapoint}s in a \gls{scatterplot} by using two \gls{feature}s as the coordinates of a \gls{datapoint}.},
	first={feature map},text={feature map} }

  \newglossaryentry{lasso}{name={least absolute shrinkage and selection operator (Lasso)}, 
	description={The\index{Lasso} least absolute shrinkage and selection operator (Lasso) is an 
		instance of \gls{srm} to learn the weights $\weights$ of a linear map 
		$\hypothesis(\featurevec) = \weights^{T} \featurevec$ based on a \gls{trainset}. 
		The Lasso is obtained from \gls{linreg} by adding the scaled $\ell_{1}$-norm 
		$\regparam \normgeneric{\weights}{1}$ to the average \gls{sqerrloss} incurred on the \gls{trainset}. 
	},
	first={ least absolute shrinkage and selection operator (Lasso)},text={Lasso} }
 
 \newglossaryentry{simgraph}{name={similarity graph}, 
 	description={Some\index{similarity graph} ML applications generate \gls{datapoint}s that 
 		are related by a domain-specific notion of similarity. These similarities can be 
 		represented conveniently using a similarity \gls{graph} $\graph = \big(\nodes \defeq \{1,\ldots,\samplesize\},\edges\big)$. 
 		The node $\sampleidx \in \nodes$ represents the $\sampleidx$-th \gls{datapoint}. Two 
 		nodes are connected by an undirected edge if the corresponding \gls{datapoint}s are similar. 
 	},
 	first={similarity graph},text={similarity graph} }

 \newglossaryentry{kld}{name={Kullback-Leibler divergence}, 
 	description={
 		 The\index{KL divergence} Kullback–Leibler divergence is a quantitative 
 		 measure for how much one \gls{probdist} is different from another \gls{probdist} \cite{coverthomas}.  
 	},
 	first={Kullback-Leibler divergence},text={KL divergence} }

\newglossaryentry{LapMat}{
	name={Laplacian matrix},
	description={The\index{Laplacian matrix} structure of a \gls{graph} $\graph$, with 
		nodes $\nodeidx=1,\ldots,\nrnodes$, can be analyzed using the properties of 
		special matrices that are associated with $\graph$. One such matrix is the 
		graph Laplacian matrix $\mL^{(\graph)} \in \mathbb{R}^{\nrnodes \times \nrnodes}$ 
		which is defined for an undirected and weighted \gls{graph} \cite{Luxburg2007,Ng2001}. 
		It is defined element-wise as (see Fig.\ \ref{fig_lap_mtx})
	\begin{equation}
		\LapMatEntry{\graph}{\nodeidx}{\nodeidx'} \defeq \begin{cases} - \edgeweight_{\nodeidx,\nodeidx'} & \mbox{ for } \nodeidx\neq \nodeidx', \edge{\nodeidx}{\nodeidx'} \in \edges \\ 
			\sum_{\nodeidx'' \neq \nodeidx} \edgeweight_{\nodeidx,\nodeidx''} & \mbox{ for } \nodeidx = \nodeidx' \\ 
							0 & \mbox{ else.} \end{cases}
	 \end{equation}
  Here, $\edgeweight_{\nodeidx,\nodeidx'}$ denotes the \gls{edgeweight} of an edge $\edge{\nodeidx}{\nodeidx'} \in \edges$. 
  \begin{figure}
  	\begin{center}
    \begin{minipage}{0.45\textwidth}
	\begin{tikzpicture}
	 	 		\begin{scope}[every node/.style={circle, draw, minimum size=1cm}]
	 					 			\node (1) at (0,0) {1};
	 					 			\node (2) [below left=of 1] {2};
	 					 			\node (3) [below right=of 1] {3};
	 					 		   \draw (1) -- (2);
	 					 			\draw (1) -- (3);
	 					 		\end{scope}
	 				 	\end{tikzpicture}
	 			 	\end{minipage} 
	 			 	\hspace*{-15mm}
 		 		\begin{minipage}{0.45\textwidth}
	 			 	 \begin{equation} 
	 				 		 \LapMat{\graph} = \begin{pmatrix} 2 & -1& -1 \\ -1& 1 & 0 \\  -1 & 0 & 1 \end{pmatrix}  
	 				 		 \nonumber
	 				 		 \end{equation} 
	 			 \end{minipage}
	 	 \caption{\label{fig_lap_mtx} Left: Some undirected graph $\graph$ with three nodes $\nodeidx=1,2,3$. 
	 		 	Right: Laplacian matrix $\LapMat{\graph}  \in \mathbb{R}^{3 \times 3}$ of $\graph$.} 
	 		 	\end{center}
	 		\end{figure}
	},
	first={Laplacian matrix},
	text={Laplacian matrix}
}

\newglossaryentry{cfwmaxmin}{name ={Courant–Fischer–Weyl min-max characterization}, 
description={Consider a \gls{psd} matrix $\mQ \in \mathbb{R}^{\nrfeatures \times \nrfeatures}$ with 
	\gls{evd} (or \emph{spectral decomposition}), 
	$$ \mQ = \sum_{\featureidx=1}^{\nrfeatures} \eigval{\featureidx} \vu^{(\featureidx)} \big(  \vu^{(\featureidx)}  \big)^{T}.$$ 
	Here, we used the ordered (in increasing fashion) \gls{eigenvalue}s 
	\begin{equation}
		\nonumber
		 \eigval{1}  \leq  \ldots \leq \eigval{\nrnodes}. 
	\end{equation}. 
	The Courant–Fischer–Weyl min-max characterization \cite[Thm. 8.1.2.]{GolubVanLoanBook} amounts 
	to representing the \gls{eigenvalue}s as solutions of optimization problems.}, 
first = {Courant–Fischer–Weyl min-max characterization (CFW)}, text={CFW}}

\newglossaryentry{kernel}{name={kernel}, 
	description={Consider\index{kernel} \gls{datapoint}s characterized by a \gls{featurevec} $\featurevec \in \featurespace$ 
	with a generic \gls{featurespace} $\featurespace$. A (real-valued) kernel $\kernel: \featurespace \times \featurespace \rightarrow \mathbb{R}$ 
	assigns each pair of \gls{featurevec}s $\featurevec, \featurevec' \in \featurespace$ a real number $\kernelmap{\featurevec}{\featurevec'}$. 
	The value $\kernelmap{\featurevec}{\featurevec'}$ is often interpreted as a measure for the similarity between $\featurevec$ 
	and $\featurevec'$. \Gls{kernelmethod}s us a kernel to transform the \gls{featurevec} $\featurevec$ to a new \gls{featurevec} $\vz = \kernelmap{\featurevec}{\cdot}$. 
         This new \gls{featurevec} belongs to a linear \gls{featurespace} $\featurespace'$ which is (in general)  
          different from the original \gls{featurespace} $\featurespace$. The \gls{featurespace} $\featurespace'$ has 
          a specific mathematical structure, i.e., it is a reproducing kernel Hilbert space \cite{LampertNowKernel,LearningKernelsBook}.
          },
	first={kernel},text={kernel} }
	
\newglossaryentry{kernelmethod}{name={kernel method}, 
	description={A\index{kernel method} kernel method is a ML method that uses a 
	\gls{kernel} $\kernel$ to map the original (raw) \gls{featurevec} $\featurevec$ of a 
	\gls{datapoint} to a new (transformed) \gls{featurevec} $\vz = \kernelmap{\featurevec}{\cdot}$ \cite{LampertNowKernel,LearningKernelsBook}.
	The motivation for transforming the \gls{featurevec}s is that, using a suitable \gls{kernel}, 
	the \gls{datapoint}s have a \emph{more pleasant} geometry in the transformed \gls{featurespace}. 
	For example, in a binary \gls{classification} problem, using transformed \gls{featurevec}s $\vz$ might 
	allow to use \gls{linmodel}s, even if the \gls{datapoint}s are not linearly 
	separable in the original \gls{featurespace} (see Figure \ref{fig_linsep_kernel}). 
	\begin{figure}[htbp]
\begin{center}
 \begin{tikzpicture}[auto,scale=0.6]
        \draw [thick] (-6,2) circle (0.1cm) node[anchor=west] {\hspace*{0mm}$\featurevec^{(5)}$};
       \draw [thick] (-8,1.6) circle (0.1cm) node[anchor=west] {\hspace*{0mm}$\featurevec^{(4)}$};
        \draw [thick] (-7.4,-1.7) circle (0.1cm) node[anchor=west] {\hspace*{0mm}$\featurevec^{(3)}$};
        \draw [thick] (-6,-1.9) circle (0.1cm) node[anchor=west] {\hspace*{0mm}$\featurevec^{(2)}$};
        \draw [thick] (-6.5,0.0) rectangle ++(0.1cm,0.1cm) node[anchor=west,above] {\hspace*{0mm}$\featurevec^{(1)}$};
        \draw [thick] (4,0) circle (0.1cm) node[anchor=north] {\hspace*{0mm}$\vz^{(5)}$};
        \draw [thick] (5,0) circle (0.1cm) node[anchor=north] {\hspace*{0mm}$\vz^{(4)}$};
        \draw [thick] (6,0) circle (0.1cm) node[anchor=north] {\hspace*{0mm}$\vz^{(3)}$};
        \draw [thick] (7,0) circle (0.1cm) node[anchor=north] {\hspace*{0mm}$\vz^{(2)}$};
        \draw [thick] (2,0) rectangle ++(0.1cm,0.1cm) node[anchor=west,above] {\hspace*{0mm}$\vz^{(1)}$};
       \draw[->,bend left=30] (-3,0) to node[midway,above] {$\vz = \kernelmap{\featurevec}{\cdot}$} (1,0);
    \end{tikzpicture}
\end{center}
\caption{
Five \gls{datapoint}s characterized by \gls{featurevec}s $\featurevec^{(\sampleidx)}$ 
and \gls{label}s $\truelabel^{(\sampleidx)} \in \{ \circ, \square \}$, for $\sampleidx=1,\ldots,5$. 
With these \gls{featurevec}s, there is no way to separate the two classes 
by a straight line (representing the \gls{decisionboundary} of a \gls{linclass}). 
In contrast, the transformed \gls{featurevec}s $\vz^{(\sampleidx)} = \kernelmap{\featurevec^{(\sampleidx)}}{\cdot}$ 
allow to separate the \gls{datapoint}s using a \gls{linclass}.  \label{fig_linsep_kernel}}
\end{figure}
},first={kernel method},text={kernel method} }

\newglossaryentry{cm}{name={confusion matrix}, 
	description={Consider\index{confusion matrix} \gls{datapoint}s characterized by \gls{feature}s $\featurevec$ 
		and \gls{label} $\truelabel$ having values from the finite \gls{labelspace} $\labelspace = \{1,\ldots,\nrcluster\}$. 
		The confusion matrix is $\nrcluster \times \nrcluster$ matrix with rows representing different values $\clusteridx$ 
		of the true label of a \gls{datapoint}. The columns of a confusion matrix correspond to different values 
		$\clusteridx'$ delivered by a hypothesis $\hypothesis(\featurevec)$. The $(\clusteridx,\clusteridx')$-th entry of 
		the confusion matrix is the fraction of \gls{datapoint}s with \gls{label} $\truelabel\!=\! \clusteridx$ and the 
		\gls{prediction} $\hat{\truelabel}\!=\!\clusteridx'$ assigned by the \gls{hypothesis} $\hypothesis$.},
	first={confusion matrix},text={confusion matrix} }

\newglossaryentry{featuremtx}{name={feature matrix}, 
	description={Consider\index{feature matrix} a \gls{dataset} $\dataset$ 
		with $\samplesize$ \gls{datapoint}s with \gls{feature} vectors $\featurevec^{(1)},\ldots,\featurevec^{(\samplesize)} \in \mathbb{R}^{\nrfeatures}$. It is convenient to 
		collect the individual \gls{feature} vectors into a \gls{feature} 
		matrix $\mX \defeq \big(\featurevec^{(1)},\ldots,\featurevec^{(\samplesize)}\big)^{T}$ 
		of size $\samplesize \times \nrfeatures$.},
	first={feature matrix},text={feature matrix} }

\newglossaryentry{dbscan}{name={density-based spatial clustering of applications with noise}, 
	description={A\index{DBSCAN} clustering algorithm for \gls{datapoint}s that are characterized by numeric feature vectors. 
		Like \gls{kmeans} and \gls{softclustering} via \gls{gmm}, also DBSCAN uses the Euclidean 
		distances between \gls{feature} vectors to determine the \gls{cluster}s. However, in contrast to \gls{kmeans} 
		and \gls{gmm}, DBSCAN uses a different notion of similarity between \gls{datapoint}s. 
		DBSCAN considers two \gls{datapoint}s as similar if they are \emph{connected} 
		via a sequence (path) of close-by intermediate \gls{datapoint}s. Thus, DBSCAN might consider 
		two \gls{datapoint}s as similar (and therefore belonging to the same cluster) even if 
		their \gls{feature} vectors have a large Euclidean distance.},
	first={density-based spatial clustering of applications with noise (DBSCAN)},text={DBSCAN} }

\newglossaryentry{fl}{name={federated learning (FL)}, description={Federated\index{federated learning} 
		learning is an umbrella term for ML methods that train models in a collaborative 
		fashion using decentralized data and computation.},first={federated learning (FL)},text={FL} }
	
\newglossaryentry{cfl}{name={clustered federated learning (CFL)}, description={
		Clustered\index{clustered federated learning} \gls{fl} (CFL) assumes that \gls{localdataset}s form clusters. 
		The \gls{localdataset}s belonging to the same cluster have similar statistical properties. 
		CFL pools \gls{localdataset}s in the same cluster to obtain a \gls{trainset} 
		for training a cluster-specific \gls{model}. \Gls{gtvmin} implements this pooling implicitly 
		by forcing the local \gls{modelparams} to be approximately identical over well-connected 
		subsets of the \gls{empgraph}.},
	first={clustered \gls{fl}},text={CFL} }

\newglossaryentry{iid}{name={i.i.d.}, description={It\index{i.i.d.} can be useful to 
		interpret \gls{datapoint}s $\datapoint^{(1)},\ldots,\datapoint^{(\samplesize)}$ 
		as \gls{realization}s of independent and identically distributed \gls{rv}s with 
		a common \gls{probdist}. If these \gls{rv}s are continuous-valued, their joint \gls{pdf} is $p\big(\datapoint^{(1)},\ldots,\datapoint^{(\samplesize)} \big) = \prod_{\sampleidx=1}^{\samplesize} p \big(\datapoint^{(\sampleidx)}\big)$ with $p(\datapoint)$ being the common 
		marginal \gls{pdf} of the underlying \gls{rv}s.},
	first={independent and identically distributed (i.i.d.)},text={{i.i.d.}} }

\newglossaryentry{outlier}{name={outlier}, description={Many\index{outlier} ML methods 
		are motivated by the \gls{iidasspt} which interprets \gls{datapoint}s as realizations of 
		\gls{iid} \gls{rv}s with a common \gls{probdist}. The \gls{iidasspt} is useful for applications  
		where the statistical properties of the data generation process are stationary (or time-invariant) \cite{Brockwell91}. 
		However, in some applications the data consists of a majority of \emph{regular} \gls{datapoint}s 
		that conform with an \gls{iidasspt} and a small number of data points that have fundamentally different 
        statistical properties compared to the regular \gls{datapoint}s. We refer to a \gls{datapoint} that 
        substantially deviates from the statistical properties of most \gls{datapoint}s as an 
        outlier. Different methods for outlier detection use different measures for this deviation. 
        Stastistical learning theory studies fundamental limits on the ability to mitigate outliers reliably \cite{doi:10.1137/0222052,10.1214/20-AOS1961}.},
	          first={outlier},text={outlier} }

\newglossaryentry{decisionregion}{name={decision region}, description={Consider\index{decision region} 
		a \gls{hypothesis} map $\hypothesis$ that delivers values from a finite set $\labelspace$. 
		For each \gls{label} value (category) $a \in \labelspace$, the \gls{hypothesis} $\hypothesis$ 
		determines a subset of \gls{feature} values $\featurevec \in \featurespace$ that result 
		in the same output $\hypothesis(\featurevec)=a$. We refer to this subset as a decision 
		region of the \gls{hypothesis} $\hypothesis$.},first={decision region},text={decision region} }

\newglossaryentry{decisionboundary}{name={decision boundary}, description={Consider\index{decision boundary} a 
		\gls{hypothesis} map $\hypothesis$ that reads in a \gls{feature} vector 
		$\featurevec \in \mathbb{R}^{\featuredim}$ and delivers a value from a finite set $\labelspace$. 
		The decision boundary of $\hypothesis$ is the set of vectors $\featurevec \in \mathbb{R}^{\featuredim}$ 
		that lie between different \gls{decisionregion}s. More precisely, a 
		vector $\featurevec$ belongs to the decision boundary if and only 
		if each neighbourhood $\{ \featurevec': \| \featurevec - \featurevec' \| \leq \varepsilon \}$, 
		for any $\varepsilon >0$, contains at least two vectors with different function values.},first={decision boundary},text={decision boundary} }

\newglossaryentry{euclidspace}{name={Euclidean space}, description={The\index{Euclidean space} 
		Euclidean space $\mathbb{R}^{\featuredim}$ of dimension $\featuredim \in \mathbb{N}$ consists 
		of vectors $\featurevec= \big(\feature_{1},\ldots,\feature_{\featurelen}\big)$, with $\featuredim$ 
		real-valued entries $\feature_{1},\ldots,\feature_{\featuredim} \in \mathbb{R}$. Such an Euclidean 
		space is equipped with a geometric structure defined by the inner product 
		$\featurevec^{T} \featurevec' = \sum_{\featureidx=1}^{\featuredim} \feature_{\featureidx} \feature'_{\featureidx}$ 
		between any two vectors $\featurevec,\featurevec' \in \mathbb{R}^{\featuredim}$ \cite{RudinBookPrinciplesMatheAnalysis}.},first={Euclidean space},text={Euclidean space} }

\newglossaryentry{eerm}{name={explainable empirical risk minimization}, description={An\index{explainable empirical risk minimization} 
		instance of structural risk minimization that adds a \gls{regularization} term to the 
		average \gls{loss} in the objective function of \gls{erm}. 
		The \gls{regularization} term is chosen to favour \gls{hypothesis} maps that are intrinsically 
		explainable for a specific user. This user is characterized by their \gls{prediction}s provided 
		for the \gls{datapoint}s in a \gls{trainset} \cite{Zhang:2024aa}.},first={explainable empirical risk minimization (EERM)},text={EERM} }

\newglossaryentry{kmeans}{name={$k$-means}, description={The\index{$k$-means} $k$-means algorithm 
		is a hard \gls{clustering} method which assigns each \gls{datapoint} of a \gls{dataset} 
		to precisely one of $k$ different \gls{cluster}s. The method alternates between updating 
		the \gls{cluster} assignments (to the cluster with nearest cluster mean) and, given the updated \gls{cluster} assignments, 
		re-calculating the cluster means \cite[Ch. 8]{MLBasics}.},first={$k$-means},text={$k$-means} }

\newglossaryentry{xml}{name={explainable ML}, description={Explainable\index{explainable AI} 
		ML methods aim at complementing each \gls{prediction} with an \gls{explanation} for how the \gls{prediction} 
		has been obtained. The construction of an explicit \gls{explanation} might not be necessary 
		if the ML method uses a sufficiently simple (or interpretable) \gls{model} \cite{rudin2019stop}.},first={explainable ML},text={explainable ML} }

\newglossaryentry{fmi}{name={Finnish Meteorological Institute}, description={The\index{Finnish Meteorological Institute}
		Finnish Meteorological Institute is a government agency responsible for gathering 
		and reporting weather data in Finland.},first={Finnish Meteorological Institute (FMI)},text={FMI} }
	
\newglossaryentry{samplemean}{name={sample mean}, description={The\index{sample mean} sample mean 
			$\vm \in \mathbb{R}^{\nrfeatures}$ for a given \gls{dataset}, with \gls{featurevec}s $\featurevec^{(1)},\ldots,\featurevec^{(\samplesize)} \in \mathbb{R}^{\nrfeatures}$, 
			is defined as 
			$$\vm = (1/\samplesize) \sum_{\sampleidx=1}^{\samplesize} \featurevec^{(\sampleidx)}.$$ 
		},
		first={sample mean},text={sample mean} }
	
\newglossaryentry{samplecovmtx}{name={sample covariance matrix}, description={The\index{sample covariance matrix} 
		sample covariance matrix $\widehat{\bf \Sigma} \in \mathbb{R}^{\nrfeatures \times \nrfeatures}$ 
		for a given set of \gls{feature} vectors $\featurevec^{(1)},\ldots,\featurevec^{(\samplesize)} \in \mathbb{R}^{\nrfeatures}$ is defined as 
		$$\widehat{\bf \Sigma} = (1/\samplesize) \sum_{\sampleidx=1}^{\samplesize} (\featurevec^{(\sampleidx)}\!-\!\widehat{\vm}) (\featurevec^{(\sampleidx)}\!-\!\widehat{\vm})^{T}.$$ 
		Here, we used the \gls{samplemean} $\widehat{\vm}$. 
	},
	first={sample covariance matrix},text={sample covariance matrix} }

\newglossaryentry{covmtx}{name={covariance matrix}, description={The\index{covariance matrix} covariance matrix of 
		a \gls{rv} $\vx \in \mathbb{R}^{\featuredim}$ is defined as $\expect \bigg \{ \big( \vx - \expect \big\{ \vx \big\} \big)  \big(\vx - \expect \big\{ \vx \big\} \big)^{T} \bigg\}$.},
	first={covariance matrix},text={covariance matrix} }
	
\newglossaryentry{highdimregime}{name={high-dimensional regime}, description={The\index{high-dimensional regime} 
		high-dimensional regime of \gls{erm} is characterized by the \gls{effdim} of the \gls{model} 
		being larger than the \gls{samplesize}, i.e., the number of (labeled) \gls{datapoint}s in the \gls{trainset}. 
		For example, \gls{linreg} methods operate in the high-dimensional regime whenever the number $\featuredim$ of \gls{feature}s 
		used to characterize \gls{datapoint}s exceeds the number of \gls{datapoint}s in the \gls{trainset}. 
		Another example of ML methods that operate in the high-dimensional regime are large \gls{ann}s, having 
		far more tunable weights (and bias terms) than the number of \gls{datapoint}s in the \gls{trainset}. 
		High-dimensional statistics is a recent main thread of probability theory that studies the 
		behavior of ML methods in the high-dimensional regime \cite{Wain2019,BuhlGeerBook}.},
   first={high-dimensional regime},text={high-dimensional regime} }

\newglossaryentry{gmm}{name={Gaussian mixture model}, description={A Gaussian\index{Gaussian mixture model} 
		mixture model (GMM) is particular type of \gls{probmodel}s for a numeric vector $\featurevec$ (e.g., 
		the \gls{feature}s of a \gls{datapoint}). Within a GMM, the vector $\featurevec$ is drawn from a randomly 
		selected \gls{mvndist} $p^{(\clusteridx)} = \mvnormal{\meanvec{\clusteridx}}{\covmtx{\clusteridx}}$ with 
		$\clusteridx = I$. The index $I \in \{1,\ldots,\nrcluster\}$ is a \gls{rv} with probabilities $\prob{I=\clusteridx} = p_{\clusteridx}$.
	     Note that a GMM is parameterized by the probability $p_{\clusteridx}$, the 
		mean vector $\clustermean^{(\clusteridx)}$ and \gls{covmtx} $\clustercov^{(\clusteridx)}$ for each $\clusteridx=1,\ldots,\nrcluster$. 
		GMMs are widely used for \gls{clustering}, density estimation and as a generative model. 
	 },first={Gaussian mixture model (GMM)},text={GMM} }
 
\newglossaryentry{ml}{name={maximum likelihood}, description={
		Consider\index{maximum likelihood} \gls{datapoint}s $\dataset=\big\{ \datapoint^{(1)}, \ldots, \datapoint^{(\samplesize)} \}$ that are interpreted 
		as realizations of \gls{iid} \gls{rv}s with a common \gls{probdist} $\prob{\datapoint; \weights}$ which 
		depends on the \gls{modelparams} $\weights \in \mathcal{W} \subseteq \mathbb{R}^{n}$. 
		Maximum likelihood methods learn \gls{modelparams} $\weights$ by maximizing 
		the probability (density) $\prob{\dataset; \weights} = \prod_{\sampleidx=1}^{\samplesize} \prob{\datapoint^{(\sampleidx)}; \weights}$ 
		of observing the \gls{dataset} is maximized. Thus, the maximum likelihood estimator is a 
		solution to the optimization problem $\max_{\weights \in \mathcal{W}} \prob{\dataset; \weights}$.
	},first={maximum likelihood},text={maximum likelihood}}

\newglossaryentry{em}{name={expectation-maximization}, description={
		\index{expectation-maximization} 
		Consider a \gls{probmodel} $\prob{\datapoint; \weights}$ for the \gls{datapoint}s $\dataset$ generated in some 
		ML application. The \gls{ml} estimator for the \gls{modelparams} $\weights$ is obtained by maximizing 
		$\prob{\dataset; \weights}$. However, the resulting oprimtization problem might be computationally 
		challenging. Expectation-maximization approximates the \gls{ml} estimator by introducing a latent 
		\gls{rv} $\vz$ such that maximizing $\prob{\dataset,\vz; \weights}$ would be easier \cite{BishopBook,hastie01statisticallearning,GraphModExpFamVarInfWainJor}. Since we 
		do not observe $\vz$, we need to estimate it from the observed \gls{dataset} $\dataset$ 
		using a conditional expectation. The resulting estimate $\widehat{\vz}$ is then used to 
		compute a new estimate $\widehat{\weights}$ by solving $\max_{\weights} \prob{\dataset, \widehat{\vz}; \weights}$. 
		The crux is that the conditional expectation $\widehat{\vz}$ depends on the \gls{modelparams} $\widehat{\weights}$ 
		which we have updated based on $\widehat{\vz}$. Thus, we have to re-calculate $\widehat{\vz}$ 
		which, in turn, results in a new choice $\widehat{\weights}$ for the \gls{modelparams}. In practice, 
		we repeat the computation of the conditional expectation (the E step) and the update 
		of the \gls{modelparams} (the M step) until some \gls{stopcrit} is met. 
  },first={expectation maximization (EM)},text={EM}}

\newglossaryentry{ppca}{name={probabilistic PCA}, description={Probabilistic\index{probabilistic PCA} \gls{pca} (PPCA) 
		extends basic \gls{pca} by using a \gls{probmodel} for \gls{datapoint}s. The \gls{probmodel} of PPCA 
		reduces the task of dimensionality reduction to an estimation problem that can be solved using \gls{em} 
		methods.},first={probabilistic PCA (PPCA)},text={PPCA}}
	
\newglossaryentry{polyreg}{name={polynomial regression}, description={Polynomial\index{polynomial regression} 
		regression aims at learning a polynomial \gls{hypothesis} map to predict a numeric \gls{label} based
		 on numeric \gls{feature}s of a \gls{datapoint}. For \gls{datapoint}s characterized by a single 
		 numeric \gls{feature}, polynomial regression uses the \gls{hypospace} 
			$\hypospace^{(\rm poly)}_{\nrfeatures} \defeq \{ \hypothesis(x) = \sum_{\featureidx=0}^{\nrfeatures-1} x^{\featureidx} \weight_{\featureidx} \}.$
			The quality of a polynomial \gls{hypothesis} map is measured using the average \gls{sqerrloss} 
			incurred on a set of labeled \gls{datapoint}s (which we refer to as \gls{trainset}).},first={polynomial regression},text={polynomial regression}}

\newglossaryentry{linreg}{name={linear regression}, description={Linear\index{linear regression} 
		regression aims to learn a linear \gls{hypothesis} map to predict a numeric \gls{label} based 
		on numeric \gls{feature}s of a \gls{datapoint}. The quality of a linear \gls{hypothesis} map is 
		measured using the average \gls{sqerrloss} incurred on a set of labeled \gls{datapoint}s, 
		which we refer to as the \gls{trainset}.},first={linear regression},text={linear regression}}
        
\newglossaryentry{ridgeregression}{name={ridge regression}, description={Ridge\index{ridge regression} 
		regression learns the \gls{weights} $\weights$ of a linear \gls{hypothesis} map $\hypothesis^{(\weights)}(\featurevec)= \weights^{T} \featurevec$. 
		The quality of a particular choice for the parameter vector $\weights$ is measured by the sum 
		of two components. The first component is the average \gls{sqerrloss} incurred by $\hypothesis^{(\weights)}$ on a set of 
		labeled \gls{datapoint}s (the \gls{trainset}). The second component is the scaled squared 
		Euclidean norm $\regparam \| \weights \|^{2}_{2}$ with a \gls{regularization} parameter 
		$\regparam > 0$. It can be shown that the effect of adding to $\regparam \| \weights \|^{2}_{2}$ to 
	the average \gls{sqerrloss} is equivalent to replacing the original \gls{datapoint}s by an ensemble of 
realizations of a \gls{rv} centered around these \gls{datapoint}s.},first={ridge regression},text={ridge regression}}

\newglossaryentry{expectation}{name={expectation}, description={
		Consider\index{expectation} a numeric \gls{feature} vector $\featurevec \in \mathbb{R}^{\featuredim}$ 
		which we interpret as the \gls{realization} of a \gls{rv} with \gls{probdist} $p(\featurevec)$. 
		The expectation of $\featurevec$ is defined as the integral $\expect \{ \featurevec \} \defeq \int \featurevec p(\featurevec)$ \cite{HalmosMeasure,BillingsleyProbMeasure,RudinBookPrinciplesMatheAnalysis}. Note that 
		the expectation is only defined if this integral exists, i.e., if the \gls{rv} is integrable.},first={expectation},text={expectation}}

\newglossaryentry{logreg}{name={logistic regression}, description={Logistic\index{logistic regression} regression learns a 
		linear \gls{hypothesis} map (\gls{classifier}) $\hypothesis(\featurevec) = \weights^{T} \featurevec$ 
		to predict a binary \gls{label} $\truelabel$ based on numeric \gls{feature} vector 
		$\featurevec$ of a \gls{datapoint}. The quality of a linear \gls{hypothesis} map is 
		measured by the average \gls{logloss} on some labeled \gls{datapoint}s (the \gls{trainset}).},
		first={logistic regression},text={logistic regression}}
	
\newglossaryentry{logloss}{name={logistic loss}, description={Consider\index{logistic loss} 
		a \gls{datapoint}, characterized by the \gls{feature}s $\featurevec$ and a binary \gls{label} $\truelabel \in \{-1,1\}$. 
		We use a real-valued \gls{hypothesis} $\hypothesis$ to predict the label $\truelabel$ 
		from the features $\featurevec$. The logistic loss incurred by this \gls{prediction} is 
		defined as 
	\begin{equation} 
		\label{equ_log_loss_gls}
		\lossfunc{(\featurevec,\truelabel)}{\hypothesis} \defeq  \log ( 1 + \exp(- \truelabel \hypothesis(\featurevec))).
\end{equation}
Carefully note that the expression \eqref{equ_log_loss_gls} 
for the logistic loss applies only if for the \gls{labelspace} $\labelspace = \{ -1,1\}$ and using 
the thresholding rule \eqref{equ_def_threshold_bin_classifier}. },first={logistic loss},text={logistic loss}}
	
\newglossaryentry{hingeloss}{name={hinge loss}, description={Consider\index{hinge loss} a \gls{datapoint}, 
		characterized by a \gls{feature} vector $\featurevec \in \mathbb{R}^{\featuredim}$ and a 
		binary \gls{label} $\truelabel \in \{-1,1\}$. The hinge loss incurred by a real-valued 
		\gls{hypothesis} map $\hypothesis(\featurevec)$ is defined as 
		\begin{equation} 
			\label{equ_hinge_loss_gls}
				\lossfunc{(\featurevec,\truelabel)}{\hypothesis} \defeq \max \{ 0 , 1 - \truelabel \hypothesis(\featurevec) \}. 
			\end{equation}
			\begin{center}
\begin{tikzpicture}
    \begin{axis}[
        axis lines=middle,
        xlabel={$\truelabel\hypothesis(\featurevec)$},
        ylabel={$\lossfunc{(\featurevec,\truelabel)}{\hypothesis}$},
 	xlabel style={at={(axis description cs:1.,0.3)}, anchor=north},  % Adjusted to be relative to axis end
        ylabel style={at={(axis description cs:0.5,1.1)}, anchor=center}, % Corrected to vertical position, rotated for readability
        xmin=-3.5, xmax=3.5,
        ymin=-0.5, ymax=2.5,
        xtick={-3, -2, -1, 0, 1, 2, 3},
        ytick={0, 1, 2},
        domain=-3:3,
        samples=100,
        width=10cm, height=6cm,
        grid=both,
        major grid style={line width=.2pt, draw=gray!50},
        minor grid style={line width=.1pt, draw=gray!20},
        legend pos=south west % Positions legend at the bottom left
    ]
        \addplot[blue, thick] {max(0, 1-x)};
    \end{axis}
\end{tikzpicture}
		\end{center}
	    A regularized variant of the hinge loss is used by the \gls{svm} \cite{LampertNowKernel}. 	    
		},first={hinge loss},text={hinge loss}}

\newglossaryentry{iidasspt}{name={i.i.d.\ assumption}, description={The i.i.d.\ assumption\index{i.i.d.} interprets \gls{datapoint}s of a \gls{dataset} 
		as the realizations of \gls{iid} \gls{rv}s.},first={i.i.d.\ assumption},text={i.i.d.\ assumption} }

\newglossaryentry{hypospace}{name={hypothesis space}, description={Every\index{hypothesis space} 
		practical ML method uses a hypothesis space (or \gls{model}) $\hypospace$. The hypothesis space of a ML 
		method is a subset of all possible maps from the \gls{featurespace} to \gls{labelspace}. The design 
		choice of the hypothesis space should take into account available computational resources and 
		statistical aspects. If the computational infrastructure allows for efficient matrix operations, and there 
		is a (approximately) linear relation between \gls{feature}s and \gls{label}, a useful choice for the 
		hypothesis space might be the \gls{linmodel}.},first={hypothesis space},text={hypothesis space} }
	
\newglossaryentry{model}{name={model}, description={In the context of ML methods, 
		the term \emph{model} typically refers to the \gls{hypospace} used by a ML method \cite{ShalevMLBook,MLBasics}.},first={model},text={model} }

\newglossaryentry{modelparams}{name={model parameters}, 
	description={Model parameters\index{model parameters} are quantities that 
	are used to select a specific \gls{hypothesis} map from a \gls{model}. 
	We can think of model parameters as a unique identifier for a \gls{hypothesis} 
	map, similar to how a social security number identifies a person in Finland.},
	first={model parameters},text={model parameters} }

\newglossaryentry{ai}{name={artificial intelligence}, description={Artificial intelligence\index{artificial intelligence} 
		refers to systems that behave rational in the sense of maximizing a long-term reward. 
		The ML-based approach to AI is to train a \gls{model} that allows to predict optimal 
		actions for a given observed state of the environment. What sets AI applications apart 
		from more basic ML applications is the choice of \gls{lossfunc}. 
		AI systems rarely have access to a labeled \gls{trainset} that allows to measure the average \gls{loss} for a 
		given choice of \gls{modelparams}. Rather, AI systems typically use a \gls{lossfunc} that 
		can only be estimated from observed reward signals.},first={artificial intelligence (AI)},text={AI} }
	
\newglossaryentry{hardclustering}{name={hard clustering}, description={Hard clustering\index{hard clustering} 
		refers to the task of partitioning a given set of \gls{datapoint}s into (few) non-overlapping \gls{cluster}s. 
		The most widely used hard clustering method is \gls{kmeans}.},first={hard clustering},text={hard clustering} }
	
\newglossaryentry{softclustering}{name={soft clustering}, description={Soft clustering\index{soft clustering} 
		refers to the task of partitioning a given set of \gls{datapoint}s into (few) overlapping clusters. 
		Each \gls{datapoint} is assigned to several different \gls{cluster}s with varying \gls{dob}. Soft clustering 
		methods determine the \gls{dob} (or soft \gls{cluster} assignment) for each \gls{datapoint} and each \gls{cluster}.
		A principled approach to soft \gls{clustering} is by interpreting \gls{datapoint}s as \gls{iid} \gls{realization}s 
		of a \gls{gmm}. We then obtain a natural choice for the \gls{dob} as the conditional 
		probability of a \gls{datapoint} belonging to a specific mixture component.},first={soft clustering},text={soft clustering} }
	
\newglossaryentry{clustering}{name={clustering}, description={Clustering\index{clustering} methods decompose a given 
		set of \gls{datapoint}s into few subsets, which are referred to as \gls{cluster}s. 
		Each \gls{cluster} consists of \gls{datapoint}s that are more similar to each 
		other than to \gls{datapoint}s outside the \gls{cluster}. Different clustering methods 
		use different measures for the similarity between \gls{datapoint}s and different 
		forms of \gls{cluster} representations. The clustering method \gls{kmeans} uses the 
		average \gls{feature} vector (\emph{cluster mean}) of a \gls{cluster} as its representative. 
		A popular \gls{softclustering} method based on \gls{gmm} represents 
		a \gls{cluster} by a \gls{mvndist}.},first={clustering},text={clustering} }
	
\newglossaryentry{cluster}{name={cluster}, description={A\index{cluster} \gls{cluster} is a subset of 
		\gls{datapoint}s that are more similar to each other than to the \gls{datapoint}s outside the \gls{cluster}. 
		The quantitative measure of similarity between \gls{datapoint}s is a design choice. If \gls{datapoint}s 
		are characterized by Euclidean \gls{feature} vectors $\featurevec \in \mathbb{R}^{\nrfeatures}$, 
		we can define the similarity between two \gls{datapoint}s via the Euclidean distance between 
		their \gls{feature} vectors.},first={cluster},text={cluster} }

\newglossaryentry{huberloss}{name={Huber loss}, description={The\index{Huber loss} 
		Huber \gls{loss} unifies the \gls{sqerrloss} and the absolute error \gls{loss}.},first={Huber loss},text={Huber loss} }

\newglossaryentry{svm}{name={support vector machine}, description={The\index{support vector machine} 
		support vector machine (SVM) is a binary \gls{classification} method that learns a linear \gls{hypothesis} map. 
		Thus, like \gls{linreg} and \gls{logreg}, it is also an instance of \gls{erm} for the \gls{linmodel}. However, 
		the support vector machine uses a different \gls{lossfunc} than those methods. As illustrated in Figure \ref{fig_svm_gls}, 
		it aims to maximally separate \gls{datapoint}s from the two different classes in the \gls{feature} space 
		(\emph{maximum margin principle}). Maximizing this separation is equivalent to minimizing a regularised 
		variant of the \gls{hingeloss} \eqref{equ_hinge_loss_gls} \cite{LampertNowKernel,Cristianini_Shawe-Taylor_2000,BishopBook}
		\begin{figure}[htbp]
			\begin{center}
				\begin{tikzpicture}[auto,scale=0.8]
					\draw [thick] (1,2) circle (0.1cm)node[anchor=west] {\hspace*{0mm}$\featurevec^{(5)}$};
					\draw [thick] (0,1.6) circle (0.1cm)node[anchor=west] {\hspace*{0mm}$\featurevec^{(4)}$};
					\draw [thick] (0,3) circle (0.1cm)node[anchor=west] {\hspace*{0mm}$\featurevec^{(3)}$};
					\draw [thick] (2,1) circle (0.1cm)node[anchor=east,above] {\hspace*{0mm}$\featurevec^{(6)}$};
					\node[] (B) at (-2,0) {\emph{support vector}};
					\draw[->,dashed] (B) to (1.9,1) ; 
					\draw [|<->|,thick] (2.05,0.95)  -- (2.75,0.25)node[pos=0.5] {$\xi$} ; 
					\draw [thick] (1,-1.5) -- (4,1.5) node [right] {$\hypothesis^{(\weights)}$} ; 
					\draw [thick] (3,-1.9) rectangle ++(0.1cm,0.1cm) node[anchor=west,above]  {\hspace*{0mm}$\featurevec^{(2)}$};
					\draw [thick] (4,.-1) rectangle ++(0.1cm,0.1cm) node[anchor=west,above] {\hspace*{0mm}$\featurevec^{(1)}$};
				\end{tikzpicture}
				\caption{The \gls{svm} learns a hypothesis (or classifier) $\hypothesis^{(\weights)}$ with 
					minimum average soft-margin \gls{hingeloss}. Minimizing this \gls{loss} is equivalent 
					to maximizing the margin $\xi$ between the \gls{decisionboundary} of $\hypothesis^{(\weights)}$ 
					and each class of the \gls{trainset}.}
				\label{fig_svm_gls}
			\end{center}
		\end{figure}
		The above basic variant of SVM is only useful if the \gls{datapoint}s from different categories can be  
		(approximately) linearly separated. For a ML application where the categories are not 
		derived from a \gls{kernel}.
},first={support vector machine (SVM)},text={SVM} }

\newglossaryentry{eigenvalue}{name={eigenvalue}, description={We refer to a 
		number $\lambda \in \mathbb{R}$ as eigenvalue of a square matrix $\mathbf{A} \in \mathbb{R}^{\featuredim \times \featuredim}$ 
		if there is a non-zero vector $\vx \in \mathbb{R}^{\featuredim} \setminus \{ \mathbf{0} \}$ such that $\mathbf{A} \vx = \lambda \vx$. },first={eigenvalue},text={eigenvalue} }
	
\newglossaryentry{eigenvector}{name={eigenvector}, description={An\index{eigenvector} 
		eigenvector of a matrix $\mathbf{A} \in \mathbb{R}^{\featuredim \times \featuredim}$ 
		is a non-zero vector $\vx \in \mathbb{R}^{\featuredim} \setminus \{ \mathbf{0} \}$ 
		such that $\mathbf{A} \vx = \lambda \vx$ with some \gls{eigenvalue} $\lambda$.},first={eigenvector},text={eigenvector} }

\newglossaryentry{evd}{name={eigenvalue decomposition}, 
	description={The\index{eigenvalue decomposition} \gls{eigenvalue} 
		decomposition for a square matrix $\mA \in \mathbb{R}^{\dimlocalmodel \times \dimlocalmodel}$ 
		is a factorization of the form 
		$$\mA = \mathbf{V} {\bm \Lambda} \mathbf{V}^{-1}.$$ 
		The columns of the matrix $\mV = \big( \vv^{(1)},\ldots,\vv^{(\dimlocalmodel)} \big)$ are the 
		\gls{eigenvector}s of the matrix $\mV$. The diagonal matrix 
		${\bm \Lambda} = {\rm diag} \big\{ \eigval{1},\ldots,\eigval{\dimlocalmodel} \big\}$ 
		contains the \gls{eigenvalue}s $\eigval{\featureidx}$ corresponding to the \gls{eigenvector}s $\vv^{(\featureidx)}$. 
		Note that the above decomposition exists only if the matrix $\mA$ is diagonalizable.},first={eigenvalue decomposition (EVD)},text={EVD} }

\newglossaryentry{svd}{name={singular value decomposition}, 
  	description={The\index{singular value decomposition} singular value 
  		decomposition for a matrix $\mA \in \mathbb{R}^{\samplesize \times \dimlocalmodel}$ 
		is a factorization of the form 
		$$\mA = \mathbf{V} {\bm \Lambda} \mathbf{U}^{T},$$ 
		with orthonormal matrices $\mV \in \mathbb{R}^{\samplesize \times \samplesize}$ 
		and $\mU \in \mathbb{R}^{\dimlocalmodel \times \dimlocalmodel}$ \cite{GolubVanLoanBook}. 
		The matrix ${\bm \Lambda} \in \mathbb{R}^{\samplesize \times \dimlocalmodel}$ is 
		only non-zero along the main diagonal, whose entries $\Lambda_{\featureidx,\featureidx}$ 
		are non-negative and referred to as singular values.
	},first={singular value decomposition (SVD)},text={SVD} }

\newglossaryentry{tv}{name={total variation}, description={See \gls{gtv}.},
	first={total variation},text={total variation} }

\newglossaryentry{gdmethods}{name={gradient-based method}, 
	description={Gradient-based\index{gradient-based methods} 
		methods are iterative techniques for finding the minimum (or maximum) 
		of a \gls{differentiable} objective function of the \gls{modelparams}. These 
		methods construct a sequence of approximations to an optimal choice for 
		\gls{modelparams} that results in a minimum (or maximum) value of the \gls{objfunc}. 
		As their name indicates, gradient-based methods use the \gls{gradient}s of the \gls{objfunc} 
		evaluated during previous iterations to construct new (hopefully) improved \gls{modelparams}. 
		One important example for a gradient-based method is \gls{gd}.},
		first={gradient-based methods},text={gradient-based methods} }

\newglossaryentry{sgd}{name={subgradient descent}, description={Subgradient\index{subgradient descent} 
		descent is a generalization of \gls{gd} that does not require differentiability of the 
		function to be minimized. This generalization is obtained by replacing the concept 
		of a \gls{gradient} with that of a sub-gradient. Similar to \gls{gradient}s, also sub-gradients 
		allow to construct local approximations of an objective function. The objective function 
		might be the \gls{emprisk} $\emperror\big( \hypothesis^{(\weights)} \big| \dataset \big)$ viewed 
		as a function of the \gls{modelparams} $\weights$ that select a \gls{hypothesis} $\hypothesis^{(\weights)} \in \hypospace$.},first={subgradient descent},text={subgradient descent} }
	
\newglossaryentry{stochGD}{name={stochastic gradient descent}, description={Stochastic\index{stochastic gradient descent} 
		\gls{gd} is obtained from \gls{gd} by replacing the \gls{gradient} of the \gls{objfunc} 
		with a stochastic approximation. A main application of stochastic gradient descent 
		is to implement \gls{erm} for a parametrized \gls{model} and a \gls{trainset} $\dataset$ 
		is either very large or not readily available (e.g., when \gls{datapoint}s are stored 
		in a database distributed all over the planet). To evaluate the \gls{gradient} of the 
		\gls{emprisk} (as a function of the \gls{modelparams} $\weights$), 
		we need to compute a sum $\sum_{\sampleidx=1}^{\samplesize} \nabla_{\weights} \lossfunc{\datapoint^{(\sampleidx)}}{\weights}$  
		over all \gls{datapoint}s in the \gls{trainset}. We obtain a stochastic 
		approximation to the \gls{gradient} by replacing the sum $\sum_{\sampleidx=1}^{\samplesize} \nabla_{\weights} \lossfunc{\datapoint^{(\sampleidx)}}{\weights}$ 
		with a sum $\sum_{\sampleidx \in \batch} \nabla_{\weights} \lossfunc{\datapoint^{(\sampleidx)}}{\weights}$ 
		over a randomly chosen subset $\batch \subseteq \{1,\ldots,\samplesize\}$ (see Figure \ref{fig_sgd_approx}). 
		We often refer to these randomly chosen \gls{datapoint}s as a \emph{batch}. 
		The batch size $|\batch|$ is an important parameter of stochastic \gls{gd}. 
		Stochastic \gls{gd} with $|\batch|> 1$ is referred to as mini-batch stochastic \gls{gd} \cite{Bottou99}. 		
		\begin{figure}
			\centering
			\begin{tikzpicture}[scale=1.5, >=stealth]
				\draw[thick, blue, domain=0.5:2.5, samples=100] plot (\x, {(\x-1.5)^2 + 1});
				\node[blue,above] at (0.5, 2) {$\sum_{\sampleidx=1}^{\samplesize}$};
				\draw[thick, red, domain=1:3, samples=100] plot (\x, {(\x-2)^2 + 0.5});
				\node[red] at (3.3, 1.5) {$\sum_{\sampleidx \in \batch}$};
			\end{tikzpicture}
		\caption{Stochastic \gls{gd} for \gls{erm} approximates the \gls{gradient} 
		$\sum_{\sampleidx=1}^{\samplesize} \nabla_{\weights} \lossfunc{\datapoint^{(\sampleidx)}}{\weights}$ 
		by replacing the 
		sum over all \gls{datapoint}s in the \gls{trainset} (indexed by $\sampleidx=1,\ldots,\samplesize$) 
		with a sum over a randomly chosen subset $\batch \subseteq \{1,\ldots,\samplesize\}$.\label{fig_sgd_approx}}
		\end{figure}
},first={stochastic gradient descent (SGD)},text={SGD} }

\newglossaryentry{pca}{name={principal component analysis (PCA)}, description={Principal\index{principal component analysis} 
		component analysis determines a linear \gls{featuremap} such that the new \gls{feature}s allow to reconstruct 
		the original \gls{feature}s with minimum reconstruction error \cite{MLBasics}.},first={principal component analysis (PCA)},text={PCA} }
	
\newglossaryentry{loss}{name={loss}, description={ML\index{loss} methods use a 
		\gls{lossfunc} $\lossfunc{\datapoint}{\hypothesis}$ to measure the error incurred 
		by applying a specific \gls{hypothesis} to a specific \gls{datapoint}. With 
		slight abuse of notation, we use the term \emph{loss} for both, the \gls{lossfunc} $\loss$ 
		itself and for its value $\lossfunc{\datapoint}{\hypothesis}$ for a specific \gls{datapoint} $\datapoint$ 
		and \gls{hypothesis} $\hypothesis$.},first={loss},text={loss} }

\newglossaryentry{lossfunc}{name={loss function}, description={A\index{loss function} loss function is a map 
		$$\lossfun: \featurespace \times \labelspace \times \hypospace \rightarrow \mathbb{R}_{+}: \big( \big(\featurevec,\truelabel\big), \hypothesis\big) \mapsto  \lossfunc{(\featurevec,\truelabel)}{\hypothesis}$$ which assigns a pair 
		of a \gls{datapoint}, with features $\featurevec$ 
		and label $\truelabel$, and a \gls{hypothesis} $\hypothesis \in \hypospace$ the 
		non-negative real number $\lossfunc{(\featurevec,\truelabel)}{\hypothesis}$. The 
		loss value $\lossfunc{(\featurevec,\truelabel)}{\hypothesis}$ quantifies the discrepancy 
		between the true \gls{label} $\truelabel$ and the \gls{prediction} $\hypothesis(\featurevec)$. 
		Lower (closer to zero) values $\lossfunc{(\featurevec,\truelabel)}{\hypothesis}$ indicate a smaller 
		discrepancy between \gls{prediction} $\hypothesis(\featurevec)$ and label $\truelabel$. 
		Figure \ref{fig_loss_function_gls} depicts a \gls{lossfunc} for a given \gls{datapoint}, 
		with \gls{feature}s $\featurevec$ and label $\truelabel$, as a function of the \gls{hypothesis} $\hypothesis \in \hypospace$. 
		\begin{figure}[htbp]
			\begin{center}
				\begin{tikzpicture}[scale = 0.7]
					\begin{axis}
						[%grid, 
						axis x line=center,
						axis y line=center,
						xlabel={},
						xlabel style={below right},
						ylabel style={above right},
						xtick=\empty,
						ytick=\empty,
						xmin=-4,
						xscale = 1.4, 
						xmax=4,
						ymin=-0.5,
						ymax=2.5
						]
						\addplot [smooth, ultra thick] table [x=a, y=b, col sep=comma] {logloss.csv};    
					\end{axis}
					\node [above] at (1,5) {$\lossfunc{(\featurevec,\truelabel)}{\hypothesis}$};
					\node [above] at (10,1) {\gls{hypothesis} $\hypothesis$};
						\node [right] at (4,6) {\gls{loss}};
				\end{tikzpicture}
			\end{center}
			\vspace*{-7mm}
			\caption{Some \gls{lossfunc} $\lossfunc{(\featurevec,\truelabel)}{\hypothesis}$ for a fixed \gls{datapoint}, with 
				\gls{feature} vector $\featurevec$ and \gls{label} $\truelabel$, and varying \gls{hypothesis} $\hypothesis$. 
				ML methods try to find (learn) a \gls{hypothesis} that incurs minimum \gls{loss}.}
			\label{fig_loss_function_gls}
	\end{figure}
 },first={loss function},text={loss function} }

\newglossaryentry{decisiontree}{name={decision tree}, description={A\index{decision tree} 
		decision tree is a flow-chart-like representation of a \gls{hypothesis} map $\hypothesis$. 
		More formally, a decision tree is a directed graph containing a root node that reads 
		in the feature vector $\featurevec$ of a \gls{datapoint}. The root node then forwards 
		the \gls{datapoint} to one of its children nodes based on some elementary test on the \gls{feature}s $\featurevec$. 
		If the receiving children node is not a leaf node, i.e., it has itself children nodes, 
	  it represents another test. Based on the test result, the \gls{datapoint} is further 
	  pushed to one of its descendants. This testing and forwarding of the \gls{datapoint} is continued 
	  until the \gls{datapoint} ends up in a leaf node (having no children nodes). 
	  }
	  ,first={decision tree},text={decision tree} }

\newglossaryentry{API} 
{
	name={Application Programming Interface (API)},
	description={An\index{application programming interface} application programming 
		interface (API) is a precise specification of the services and resources 
		offered by software or hardware implementing that API.},
	first={application programming interface (API)},
	text={API}
}

\newglossaryentry{hilbertspace}{name={Hilbert space},description={A\index{Hilbert space} 
		Hilbert space is a linear vector space equipped with an inner product between 
		pairs of vectors. One important example of a Hilbert space is the \gls{euclidspace} 
		$\mathbb{R}^{\featuredim}$, for some dimension $\featuredim$, which consists of 
		Euclidean vectors $\vu = \big(u_{1},\ldots,u_{\featurelen}\big)^{T}$ along with the inner 
		product $\vu^{T} \vv$.},first={Hilbert space},text={Hilbert space}}

\newglossaryentry{sample}{name={sample},description={A\index{sample} 
		finite sequence (list) of \gls{datapoint}s $\datapoint^{(1)},\ldots,\datapoint^{(\sampleidx)}$ that 
		is obtained or interpreted as the realizations of $\samplesize$ \gls{iid} \gls{rv}s 
		with the common \gls{probdist} $p(\datapoint)$. The length $\samplesize$ of 
		the sequence is referred to as the \gls{samplesize}.},first={sample},text={sample}}
	
\newglossaryentry{samplesize}
{name=sample size,
	description={The\index{sample size} number of individual \gls{datapoint}s 
		contained in a \gls{dataset} obtained as the \gls{realization}s of \gls{iid} \gls{rv}s with 
		common \gls{probdist}.},first={sample size},text={sample size}
}

\newglossaryentry{ann}
{name=artificial neural network,
	description={An\index{artificial neural network} artificial neural network is a 
		graphical (signal-flow) representation of a map from \gls{feature}s of 
		a \gls{datapoint} at its input to a \gls{prediction} for the \gls{label} 
		as its output.},first={artificial neural network (ANN)},text={ANN}
}

\newglossaryentry{randomforest}
{name=random forest,
	description={A\index{random forest} random forest is a set (ensemble) of different \gls{decisiontree}s. 
		Each of these \gls{decisiontree}s is obtained by fitting a perturbed copy of 
		the original \gls{dataset}.},first = {random forest}, text={random forest}
}

\newglossaryentry{bagging}{name={bagging},description={Bagging\index{bagging} (or \emph{bootstrap aggregation}) 
		is a generic technique to improve (the robustness of) a given ML method. The idea is to use the \gls{bootstrap} 
		to generate perturbed copies of a given \gls{dataset} and then to learn a separate \gls{hypothesis} for 
		each copy. We then predict the \gls{label} of a \gls{datapoint} by combining or aggregating the individual 
		\gls{prediction}s of each separate \gls{hypothesis}. For \gls{hypothesis} maps delivering numeric \gls{label} 
		values, this aggregation could be implemented by computing the average of individual \gls{prediction}s.},first={bootstrap aggregation (bagging)},text={bagging}}

\newglossaryentry{gd}{name={gradient descent (GD)},description={Gradient\index{gradient descent} 
		descent is an iterative method for finding the minimum of a \gls{differentiable} function $f(\weights)$ 
		of a vector-valued argument $\weights \in \mathbb{R}^{\featurelen}$. Consider a current guess or 
		approximation $\weights^{(\itercntr)}$ for minimum. We would like to find a new (better) vector $\weights^{(\itercntr+1)}$ 
		that has smaller objective value $f(\weights^{(\itercntr+1)}) < f\big(\weights^{(\itercntr)}\big)$ than 
		the current guess $\weights^{(\itercntr)}$. We can achieve this typically by using a gradient step
		\begin{equation} 
			\label{equ_def_GD_step}
			\weights^{(\itercntr\!+\!1)} = \weights^{(\itercntr)} - \lrate \nabla f(\weights^{(\itercntr)})
		\end{equation} 
		with a sufficiently small \gls{stepsize} $\lrate>0$. Figure \ref{fig_basic_GD_step} illustrates the effect of 
		a single \gls{gd} step \eqref{equ_def_GD_step}.
		\begin{figure}[htbp]
			\begin{center}
				\begin{tikzpicture}[scale=0.8]
					\draw[loosely dotted] (-4,0) grid (4,4);
					\draw[blue, ultra thick, domain=-4.1:4.1] plot (\x,  {(1/4)*\x*\x});
					\draw[red, thick, domain=2:4.7] plot (\x,  {2*\x - 4});
					\draw[<-] (4,4) -- node[right] {$\nabla f(\weights^{(\itercntr)})$} (4,2);
					\draw[->] (4,4) -- node[above] {$-\lrate \nabla f(\weights^{(\itercntr)})$} (2,4);
					\draw[<-] (4,2) -- node[below] {$1$} (3,2) ;
					\draw[->] (-4.25,0) -- (4.25,0) node[right] {$\weights$};
					\draw[->] (0,-2pt) -- (0,4.25) node[above] {$f(\weights)$};
					\draw[shift={(0,0)}] (0pt,2pt) -- (0pt,-2pt) node[below] {$\overline{\weights}$};
					\draw[shift={(4,0)}] (0pt,2pt) -- (0pt,-2pt) node[below] {$\weights^{(\itercntr)}$};
					\draw[shift={(2,0)}] (0pt,2pt) -- (0pt,-2pt) node[below] {$\weights^{(\itercntr\!+\!1)}$};
					\foreach \y/\ytext in {1/1, 2/2, 3/3, 4/4}
					\draw[shift={(0,\y)}] (2pt,0pt) -- (-2pt,0pt) node[left] {$\ytext$};  
				\end{tikzpicture}
			\end{center}
			\caption{A single gradient step \eqref{equ_def_GD_step} towards the minimizer $\overline{\weights}$ of $f(\weights)$.}
			\label{fig_basic_GD_step}
		\end{figure}
		},first={gradient descent (GD)},text={GD}}

\newglossaryentry{abserr}{name={absolute error loss},description={
			Consider a \gls{datapoint} with \gls{feature}s $\featurevec \in \featurespace$ and 
			numeric \gls{label} $\truelabel \in \mathbb{R}$. 
			The absolute error loss \index{absolute error loss} incurred by 
			a \gls{hypothesis} $\hypothesis: \featurespace \rightarrow \mathbb{R}$ 
			is defined as $|\truelabel - \hypothesis(\featurevec)|$.},
			first={absolute error loss},text={absolute error loss}}

\newglossaryentry{device}{name={device},description={
				Any physical system that is can be used to store and process data. In the context of ML, 
				we typically mean a computer that is able to read in \gls{datapoint}s from different 
				sources and, in turn, to train a ML \gls{model} using these \gls{datapoint}s.},
				first={device},text={device}}

\newglossaryentry{llm}{name={Large Language Model},description={
	Large Language Models (LLMs) is an umbrella term for ML methods 
	that process and generate human-like text. These methods typically 
	use \gls{deepnet}s with billions (or even trillions) of parameters. 
	A widely used choice for the network architecture is referred to as 
	Transformers \cite{vaswani2017attention}. The training of LLMs is often  
	based on the task of predicting a few words that are intentionally removed 
	from a large text corpus. Thus, we can construct labelled \gls{datapoint}s 
	simply by selecting some words of a text as \gls{label}s and the remaining 
	words as \gls{feature}s of \gls{datapoint}s. This construction requires 
	very little human supervision and allows for generating sufficiently 
	large \gls{trainset}s for LLMs.},
					first={Large Language Model (LLM)},text={LLM}}

\newglossaryentry{huberreg}{name={Huber regression},description={
			Huber regression\index{Huber regression} refers \gls{erm}-based methods 
			that use the \gls{huberloss} as measure for the \gls{prediction} error. 
			Two important special cases of Huber regression are \gls{ladregression} and 
			\gls{linreg}. Tuning the threshold parameter of the \gls{huberloss} allows 
			to trade the robustness against outliers of abserr 
			against the smoothness of the \gls{sqerrloss}.},
			first={Huber regression},text={Huber regression}}

\newglossaryentry{ladregression}{name={least absolute deviation regression},description={
		Least\index{least absolute deviation regression} absolute deviation regression is 
		an instance of \gls{erm} using the absolute error loss. It is a special case of 
		\gls{huberreg}.},
		first={least absolute deviation regression},text={least absolute deviation regression}}

\newglossaryentry{bayesrisk}{name={Bayes risk},description={Consider a \gls{probmodel} with 
joint \gls{probdist} $p(\featurevec,\truelabel)$ for the \gls{feature}s $\featurevec$ 
and \gls{label} $\truelabel$ of a \gls{datapoint}. The\index{Bayes risk} Bayes \gls{risk} 
is the minimum possible \gls{risk} that can be achieved by any \gls{hypothesis} 
$\hypothesis: \featurespace \rightarrow \labelspace$. Any \gls{hypothesis} that achieves 
the Bayes risk is referred to as a \gls{bayesestimator} \cite{LC}.},first={Bayes risk},text={Bayes risk}}
	
\newglossaryentry{bayesestimator}{name={Bayes estimator},description={Consider\index{Bayes estimator} 
a \gls{probmodel} with joint \gls{probdist} $p(\featurevec,\truelabel)$ for the \gls{feature}s $\featurevec$ and \gls{label} 
$\truelabel$ of a \gls{datapoint}. For a given \gls{lossfunc} $\lossfunc{\cdot}{\cdot}$, we refer to a \gls{hypothesis} 
$\hypothesis$ as a Bayes estimator if its \gls{risk} $\expect\{ \lossfunc{\pair{\featurevec}{\truelabel}}{\hypothesis} \}$ is 
minimum \cite{LC}. Note that the property of a \gls{hypothesis} being a \gls{bayesestimator} depends on 
the underlying \gls{probdist} and the choice for the \gls{lossfunc} $\lossfunc{\cdot}{\cdot}$.},
		first={Bayes estimator},text={Bayes estimator}}

\newglossaryentry{weights}{name={weights},
	description={Consider\index{weights} a parameterized \gls{hypospace} $\hypospace$. 
		We\index{weights} use the term weights for numeric \gls{modelparams} that are 
		used to scale \gls{feature}s or their transformations in order to compute $\hypothesis^{(\weights)} \in \hypospace$. A \gls{linmodel} uses weights $\weights=\big(\weight_{1},\ldots,\weight_{\nrfeatures}\big)^{T}$ to compute 
		the linear combination $\hypothesis^{(\weights)}(\featurevec)= \weights^{T} \featurevec$. 
		Weights are also used in \gls{ann}s to form linear combinations of \gls{feature}s or the 
		outputs of neurons in hidden layers.},first={weights},text={weights}}
	
\newglossaryentry{probdist}{name={probability distribution},
	description={To\index{probability distribution} analyze ML methods it can be useful 
		to interpret \gls{datapoint}s as \gls{iid} \gls{realization}s of a \gls{rv}. The typical 
		properties of such \gls{datapoint}s are then governed by the probability distribution 
		of this \gls{rv}. The probability distribution of a binary \gls{rv} $\truelabel \in \{0,1\}$ 
		is fully specified by the probabilities $\prob{\truelabel = 0}$ and 
		$\prob{\truelabel=1}\!=\!1\!-\!\prob{\truelabel=0}$. The probability 
		distribution of a real-valued \gls{rv} $\feature \in \mathbb{R}$ might be specified 
		by a probability density function $p(\feature)$ such that $\prob{ \feature \in [a,b] } \approx  p(a) |b-a|$. 
	    In the most general case, a probability distribution is defined by a probability measure \cite{GrayProbBook,BillingsleyProbMeasure}.},first={probability distribution},text={probability distribution}}

\newglossaryentry{pdf}{name={probability density function (pdf)},
	description={The\index{probability density function} probability density function (pdf) $p(\feature)$ 
		of a real-valued \gls{rv} $\feature \in \mathbb{R}$ is a particular representation of its \gls{probdist}. 
		If the pdf exists, it can be used to compute the probability that $\feature$ takes on a value 
		from a (measurable) set $\mathcal{B} \subseteq \mathbb{R}$ via $\prob{\feature \in \mathcal{B}} = \int_{\mathcal{B}} p(\feature') d \feature'$ \cite[Ch. 3]{BertsekasProb}. The pdf of a vector-valued \gls{rv} $\featurevec \in \mathbb{R}^{\featuredim}$ (if it exists) 
        allows to compute the probability that $\featurevec$ falls into a (measurable) region $\mathcal{R}$ via 
        $\prob{\featurevec \in \mathcal{R}} = \int_{\mathcal{R}} p(\featurevec') d \feature_{1}' \ldots d \feature_{\featuredim}' $ \cite[Ch. 3]{BertsekasProb}.},
first={probability density function (pdf)},text={pdf}}

\newglossaryentry{parameters}{name={parameters},
	description={The\index{parameters} parameters of a ML \gls{model} are tunable 
		(learnable or adjustable) quantities that allow to choose between different \gls{hypothesis} maps. 
		For example, the linear model $\hypospace \defeq \{\hypothesis^{(\weights)}: \hypothesis^{(\weights)}(\feature)= \weight_{1} \feature + \weight_{2}\}$ 
		consists of all \gls{hypothesis} maps $\hypothesis^{(\weights)}(\feature)= \weight_{1} \feature + \weight_{2}$ 
		with a particular choice for the parameters $\weights = \big(\weight_{1},\weight_{2}\big)^{T} \in \mathbb{R}^{2}$. 
		Another example of parameters is the weights assigned to the connections 
		between neurons of an \gls{ann}.},first={parameters},text={parameters}}

\newglossaryentry{lln}{name={law of large numbers},
	description={The\index{law of large numbers} law of large numbers refers to the 
		convergence of the average of an increasing (large) number of \gls{iid} \gls{rv}s 
		to the \gls{mean} of their common \gls{probdist}. Different instances of the 
		law of large numbers are obtained using different notions of convergence \cite{papoulis}.},first={law of large numbers},text={law of large numbers}}
    
\newglossaryentry{stopcrit}{name={stopping criterion},
	description={Many\index{stopping criterion} ML methods use iterative algorithms that construct a 
		sequence of model parameters (such as the weights of a linear map or 
		the weights of an \gls{ann}) that (hopefully) converge to an optimal choice 
		for the model parameters. In practice, given finite computational 
		resources, we need to stop iterating after a finite number of times. 
		A stopping criterion is any well-defined condition required for stopping 
		iterating.},first={stopping criterion},text={stopping criterion}}

\newglossaryentry{kCV}{name={$k$-fold cross-validation ($\nrfolds$-fold CV)},
	description={$k$-fold cross-validation\index{k-fold cross-validation} is a 
		method for learning and validating a \gls{hypothesis} using a given \gls{dataset}. 
		This method divides the \gls{dataset} evenly into $k$ subsets or \emph{folds} 
		and then executes $k$ repetitions of \gls{model} training (e.g., via \gls{erm}) and \gls{validation}. 
		Each repetition uses a different fold as the \gls{valset} and the remaining $k-1$ folds 
		as a \gls{trainset}. The final output is the average of the \gls{valerr}s obtained 
		from the $k$ repetitions.},first={$k$-fold cross-validation ($k$-fold CV)},text={$k$-fold CV}}
	
\newglossaryentry{renyidiv}{name={R\'enyi divergence}, 
	description={The R\'enyi divergence\index{R\'enyi divergence} measures the (dis-)similarity 
		between two \gls{probdist}s \cite{RenyiInfo95}.}, 
	first = {R\'enyi divergence}, text = {R\'enyi divergence}} 

\newglossaryentry{nonsmooth}{name={non-smooth},
	description={We\index{non-smooth} refer to a function as non-smooth if it is not 
		\gls{smooth} \cite{nesterov04}.},first={non-smooth},text={non-smooth}}

\newglossaryentry{convex}{name={convex},
	description={A\index{convex} subset $\mathcal{C} \subseteq \mathbb{R}^{\featuredim}$ of the 
		\gls{euclidspace} $\mathbb{R}^{\featuredim}$ is referred to as 
		convex if it contains the line segment between any two points 
		of that set. We define a function as convex if its epigraph is a 
		convex set \cite{BoydConvexBook}.},first={convex},text={convex}}

\newglossaryentry{smooth}{name={smooth},
	description={We\index{smooth} refer to a real-valued function as smooth if it is \gls{differentiable} 
		and its \gls{gradient} is continuous \cite{nesterov04,CvxBubeck2015}. A 
		differentiable function $f(\weights)$ is referred to as $\beta$-smooth if the \gls{gradient} 
		$\nabla f(\weights)$ is Lipschitz continuous with Lipschitz constant $\beta$, i.e., 
		$$\| \nabla f(\weights) - \nabla f(\weights') \| \leq \beta \| \weights - \weights' \|.$$ },first={smooth},text={smooth}}

\newglossaryentry{dataaug}{name={data augmentation},
	description={Data augmentation\index{data augmentation} methods add synthetic \gls{datapoint}s 
		to an existing set of \gls{datapoint}s. These synthetic \gls{datapoint}s are obtained by 
		perturbations (e.g., adding noise to physical measurements) or transformations 
		(e.g., rotations of images) of the original \gls{datapoint}s. These perturbations and 
		transformations are such that the resulting synthetic \gls{datapoint}s should 
		still have the same \gls{label}. As a case in point, a rotated cat image is still 
		a cat image even if their \gls{featurevec}s (obtained by stacking pixel color intensities) 
		are very different. Data augmentation can be an efficient form of \gls{regularization}.},first={data augmentation},text={data augmentation}}

\newglossaryentry{localdataset}{name={local dataset},description={The\index{local dataset} concept of a local dataset is 
		in-between the concept of a \gls{datapoint} and a \gls{dataset}. A local dataset consists of several 
		individual \gls{datapoint}s which are characterized by \gls{feature}s and \gls{label}s. In contrast to a 
		single \gls{dataset} used in basic ML methods, a local dataset is also related to other local datasets via different notions 
		of similarities. These similarities might arise from \gls{probmodel}s or communication infrastructure and 
		are encoded in the edges of a \gls{empgraph}.},first={local dataset},text={local dataset}}
	
\newglossaryentry{localmodel}{name={local model},description={Consider\index{local model} a collections 
		of \gls{localdataset}s that are assigned to the nodes of a \gls{empgraph}. A local model $\localmodel{\nodeidx}$ 
		is a \gls{hypospace} assigned to a node $\nodeidx \in \nodes$. Different nodes might be assigned 
		different \gls{hypospace}s, i.e., in general $\localmodel{\nodeidx} \neq \localmodel{\nodeidx'}$ for different 
		nodes $\nodeidx, \nodeidx' \in \nodes$.  },first={local model},text={local model}}
	
\newglossaryentry{mutualinformation}
{name={mutual information},
 description={The\index{mutual information} mutual information $\mutualinformation{\featurevec}{\truelabel}$ 
 	between two \gls{rv}s $\featurevec$, $\truelabel$ defined on the same probability 
 	space is given by \cite{coverthomas} $$\mutualinformation{\featurevec}{\truelabel} \defeq \expect \left\{ \log \frac{p (\featurevec,\truelabel)}{p(\featurevec)p(\truelabel)} \right\}.$$ It is a measure for how well we can estimate $\truelabel$ based 
	solely from $\featurevec$. A large value of $\mutualinformation{\featurevec}{\truelabel}$ indicates that 
	$\truelabel$ can be well predicted solely from $\featurevec$. This \gls{prediction} could be obtained by a 
		\gls{hypothesis} learnt by a \gls{erm}-based ML method. 
	 }, first = {mutual information (MI)}, text={MI} 
}

\newglossaryentry{zerogradientcondition}{name={zero-gradient condition},
	description={Consider\index{zero-gradient condition} the unconstrained 
		optimization problem $\min_{\weights \in \mathbb{R}^{\dimlocalmodel}} f(\weights)$  with 
			a \gls{smooth} and \gls{convex} \gls{objfunc} $f(\weights)$. A necessary and 
			sufficient condition for a vector $\widehat{\weights} \in \mathbb{R}^{\dimlocalmodel}$ 
			to solve this problem is that the \gls{gradient} $\nabla f \big( \widehat{\weights} \big)$ is the zero-vector, 
			$$ \nabla f \big( \widehat{\weights} \big) = \mathbf{0} \Leftrightarrow  f \big( \widehat{\weights} \big) = \min_{\weights \in \mathbb{R}^{\dimlocalmodel}} f(\weights) .$$ }, 
			first={zero-gradient condition},text={zero-gradient condition}}

\newglossaryentry{edgeweight}{name={edge weight},
	description={Each\index{edge weight} edge $\edge{\nodeidx}{\nodeidx'}$ of a \gls{empgraph} is 
		assigned a non-negative \gls{edgeweight}  $\edgeweight_{\nodeidx,\nodeidx'}\geq0$. 
		A zero \gls{edgeweight} $\edgeweight_{\nodeidx,\nodeidx'}=0$ indicates the absence 
		of an edge between nodes $\nodeidx, \nodeidx' \in \nodes$.}, 
	first={edge weight},text={edge weight}}

\newglossaryentry{dataminprinc}{name={data minimization principle},
	description={European\index{data minimization principle} data protection regulation 
		includes a data minimization principle. This principle requires a data controller to 
		limit the collection of personal information to what is directly relevant and necessary 
		to accomplish a specified purpose. The data should be retained only for as long as 
		necessary to fulfill that purpose \cite[Article 5(1)(c)]{GDPR2016} \cite{EURegulation2018}.}, 
	first={data minimization principle},text={data minimization principle}}

\makeglossaries
\glsdisablehyper % to prevent ref's of glossary terms

\tikzset{
	font={\fontsize{18pt}{12}\selectfont}}

\definecolor{g1}{rgb}{0.1059, 0.4745, 0.2235} 
\definecolor{g2}{rgb}{0.67, 0.83, 0.62} 
\definecolor{n}{rgb}{0.73, 0.73, 0.73} 
\definecolor{v1}{rgb}{0.76, 0.64, 0.81} 
\definecolor{v2}{rgb}{0.45, 0.156, 0.5} 
\definecolor{o}{rgb}{0.90, 0.36, 0.0} 
\definecolor{b1}{rgb}{0, 0.447, 0.698} 
\definecolor{b2}{rgb}{0.22, 0.576, 0.76} 
\definecolor{b3}{rgb}{0.55, 0.76, 0.87} 

\pgfplotscreateplotcyclelist{mycolors}{
	{g1}, 
	{b1}, 
	{n}, 
	{o}, 
	{v2}
}

\pgfplotscreateplotcyclelist{mycolors3}{
	{g1}, 
	{b1}, 
	{n}
	}
	
\pgfplotscreateplotcyclelist{grad}{
		{b1}, 
		{b2}, 
		{b3}
	}

\tikzstyle{ncyan}=[circle, draw=cyan!70, thin, fill=white, scale=0.8, font=\fontsize{11}{0}\selectfont]
\tikzstyle{ngreen}=[circle,  draw=green!70, thin, fill=white, scale=0.8, font=\fontsize{11}{0}\selectfont]
\tikzstyle{nred}=[circle, draw=red!70, thin, fill=white, scale=0.8, font=\fontsize{11}{0}\selectfont]
\tikzstyle{ngray}=[circle, draw=gray!70, thin, fill=white, scale=0.55, font=\fontsize{14}{0}\selectfont]
\tikzstyle{nyellow}=[circle, draw=yellow!70, thin, fill=white, scale=0.55, font=\fontsize{14}{0}\selectfont]
\tikzstyle{norange}=[circle,  draw=orange!70, thin, fill=white, scale=0.55, font=\fontsize{10}{0}\selectfont]
\tikzstyle{npurple}=[circle,draw=purple!70, thin, fill=white, scale=0.55, font=\fontsize{10}{0}\selectfont]
\tikzstyle{nblue}=[circle, draw=blue!70, thin, fill=white, scale=0.55, font=\fontsize{10}{0}\selectfont]
\tikzstyle{nteal}=[circle,draw=teal!70, thin, fill=white, scale=0.55, font=\fontsize{10}{0}\selectfont]
\tikzstyle{nviolet}=[circle, draw=violet!70, thin, fill=white, scale=0.55, font=\fontsize{10}{0}\selectfont]
\tikzstyle{qgre}=[rectangle, draw, thin,fill=green!20, scale=0.8]
\tikzstyle{rpath}=[ultra thick, red, opacity=0.4]
\tikzstyle{legend_isps}=[rectangle, rounded corners, thin,fill=gray!20, text=blue, draw]
\usetikzlibrary{positioning}

\begin{document}

\title{Plug In and Learn: Federated Intelligence over a Smart Grid of Models\\
\thanks{This work has been supported by the Academy of Finland via funding granted 
	under decision numbers 331197 and 331197.}
}

\author{

\IEEEauthorblockN{S. Abdurakhmanova, Y. Sarcheshmehpour and A. Jung} %Yasmin Sarcheshmehpour\IEEEauthorrefmark{2}} 
  \IEEEauthorblockA{Aalto University, Finland
	\\}
%\IEEEauthorblockA{\IEEEauthorrefmark{2}Silo AI }
}

\maketitle

\begin{abstract}
We present a model-agnostic federated learning method that mirrors the operation of a 
smart power grid: diverse local models, like energy prosumers, train independently on their 
own data while exchanging lightweight signals to coordinate with statistically similar peers. 
This coordination is governed by a graph-based regularizer that encourages connected 
models to produce similar predictions on a shared, public unlabeled dataset. The resulting 
method is a flexible instance of regularized empirical risk minimization and supports a wide 
variety of local models—both parametric and non-parametric—provided they can be trained 
via regularized loss minimization. Such training is readily supported by standard ML libraries 
including \texttt{scikit-learn}, \texttt{Keras}, and \texttt{PyTorch}.
\end{abstract}


\section{Introduction}
\label{sec:intro}

Many application domains of machine learning (ML), such as numerical weather prediction, 
the Internet of Things, and healthcare generate decentralized data \cite{BigDataNetworksBook}. 
Decentralized data consists of local datasets that are related by an intrinsic network structure. 
Such a network structure might arise from relations between the generators of the \gls{localdataset}s 
or functional constraints of the computational infrastructure \cite{NewmannBook,MAL-078-3}. 
We can represent this networked data using an undirected weighted \gls{empgraph} \cite[Ch. 11]{SemiSupervisedBook}. 

A substantial body of work exists on ML and signal processing 
models and techniques for graph structured data \cite{SemiSupervisedBook,MAL-078-1,MAL-078-2,MAL-078-3}. 
Most existing work focus on parametric models for local datasets connected by an intrinsic network structure.
The scalar graph signal-in-noise model is perhaps the most basic instance of such a parametric model \cite{JungTVMin2019,EmergingGSP}. 
For this model, sampling theorems and generalization bounds have been derived using different 
smoothness or \gls{clustasspt}s \cite{WhenIsNLASSO,SamplingTheoremGraphSignals,JungTVMin2019,JungNetExp2020}. 
The extension from scalar signal-in-noise models to vector-valued graph signals and networked 
exponential families have been explored in \cite{JungNetExp2020,VarmaTSIPN2020}. 

\Gls{fl} broadly refers to the collaborative training of ML models using decentralized data. FL methods have been championed for high-dimensional 
parametric models, such as deep nets \cite{pmlr-v54-mcmahan17a,FLBookLudwig,FLBookYang}. 
So far, FL research has primarily focused on distributed optimization methods that exchange various forms of model parameter updates, such as gradients \cite{Liu2020,DeanDistSGD2012,TsiBerAth86,NEURIPS2018_8fb21ee7}. 
However, there exists little work on FL for non-parametric models, such as decision trees.
The adaption of specific decision tree algorithms to FL setting is discussed in \cite[Ch. 2]{FLBookLudwig}. 

The closest to our work is a recent study on using knowledge distillation to couple the training 
of \gls{localmodel}s \cite{Afonin2022,zhang2021parameterized,Cho21}. Similar to this approach, we also use predictions of the local models on a predefined unlabelled set of \gls{datapoint}s to couple 
their training processes. However, unlike in \cite{Afonin2022,zhang2021parameterized,Cho21}, we exploit 
the network structure of decentralized data to construct a \gls{regularizer}. 

Our regularization approach is similar in spirit to \cite{zhang2021parameterized} which studies a 
centralized \gls{fl} architecture to train personalized parametric \gls{localmodel}s for 
multi-class classification. In contrast, our method is compatible with both non-parametric models and various choices for the \gls{lossfunc} (regression or classification). Moreover, our 
method lends to distributed implementations as message passing over the data network. 

Similar to our approach, \cite{Cho21} employs knowledge distillation combined with a clustering step on the central server.
In contrast, we leverage a predefined similarity network to "pool" similar local datasets.

{\bf Contribution.} To the best of our knowledge, we present the first fully model-agnostic FL 
method for networks of data and personalized models. Our method copes with arbitrary collections 
of local models for which efficient implementations are available. Such implementations are typically 
available in high-level programming frameworks such as the Python libraries \texttt{scikit-learn}, \texttt{Keras}, 
or \texttt{PyTorch} \cite{JMLR:v12:pedregosa11a,AutoKeras2023,NEURIPS2019_bdbca288}. 
The proposed method couples the training of well-connected local models (forming a cluster) 
via enforcing them to deliver similar predictions for a pre-specified public dataset(s). 

{\bf Outline.} Section \ref{sec:non_parametric_nfl} formulates the problem of \gls{fl} from 
decentralized data. 
Section  \ref{sec_fedrelax} presents a model-agnostic \gls{fl} method , which couples the training of local models through regularization. The regularization is implemented by enforcing a small variation of local models at 
well-connected nodes (clusters). 
%Section \ref{sec_fedrelax} presents a model-agnostic \gls{fl} method for training heterogeneous networks of (local) ML models in a distributed fashion.
In section \ref{sec:num_exps}, we apply our approach to a network of heterogeneous local datasets and models.

\section{Problem Formulation}
\label{sec:non_parametric_nfl}

Section \ref{sec_emp_graph} introduces the \gls{empgraph} as a useful representation 
of collections of \gls{localdataset}s along with their similarities. 
%The edges of the empirical graph might arise not only from statistical 
%similarities but also from the computational infrastructure. Consider the extreme case of local datasets having 
%identical statistical properties (they are all sampled from the same \gls{probdist}). Here, it would be natural to 
%use a fully connected graph as the \gls{empgraph}. However, from a computational perspective it might be 
%useful to use a sparser graph with a smaller number of edges. 
Section \ref{sec_net_models} augments the \gls{empgraph} by assigning a separate 
local \gls{hypospace} (or model) to each node. 
Section \ref{sec_gtv_measure} introduces the \gls{gtv} as a quantitative measure for the variation of heterogeneous networks of ML models. 

\subsection{The Empirical Graph } 
\label{sec_emp_graph} 

Consider some application domain generating networked data which is constituted by a decentralized 
collection of local datasets $\localdataset{\nodeidx}$, for $\nodeidx = \{1,\ldots,\nrnodes\}$. We use 
the concept of an \gls{empgraph} to represent such a collection of local datasets along with their network structure. 
%using an \gls{empgraph} $\graph \defeq \pair{\nodes}{\edges}$ with nodes (vertices) $\nodes = \{1,\ldots,\nrnodes\}$. 
The \gls{empgraph} of networked data is an undirected weighted graph $\graph=(\nodes,\edges)$ 
whose nodes $\nodes \defeq \{1,\ldots,\nrnodes\}$ carry the \gls{localdataset}s $\localdataset{\nodeidx}$, for $\nodeidx \in \nodes$. 
%We  represent collections of \gls{localdataset}s
%along with their similarities by an associated 
In particular, each node $\nodeidx \!\in\!\nodes$ of the \gls{empgraph} $\graph$ carries the \gls{localdataset}
\begin{equation} 
	\label{equ_def_local_dataset_plain}
	\localdataset{\nodeidx} \defeq \left\{ \big(\featurevec^{(\nodeidx,1)},\truelabel^{(\nodeidx,1)}\big), \ldots,\big(\featurevec^{(\nodeidx,\samplesize_{\nodeidx})},\truelabel^{(\nodeidx,\samplesize_{\nodeidx})}\big) \right\}.  %\big(\featurevec^{(\samplesize_{\nodeidx})},\label{(\samplesize_{\nodeidx})} \big) \big\}. 
\end{equation} 
Here, $\featurevec^{(\nodeidx,\sampleidx)} \in \featurespace$ and $\truelabel^{(\nodeidx,\sampleidx)}\in \labelspace$ 
denote, respectively, the feature vector and the true label of the $\sampleidx$-th \gls{datapoint} in the \gls{localdataset} $\localdataset{\nodeidx}$. 
In principle, our method allows for an arbitrary \gls{featurespace} $\featurespace$ and \gls{labelspace} $\labelspace$. 
However, unless stated otherwise, we consider the choices $\featurespace \defeq \mathbb{R}^{\nrfeatures}$ and $\labelspace \defeq \mathbb{R}$. 
We allow the size $\samplesize_{\nodeidx}$ of the \gls{localdataset} to vary between the nodes $\nodeidx \in \nodes$. 

An undirected edge $\{\nodeidx,\nodeidx'\}\!\in\!\edges$ in the \gls{empgraph} indicates that the \gls{localdataset}s 
$\localdataset{\nodeidx}$ and $\localdataset{\nodeidx'}$ have similar statistical properties. 
We quantify the level of similarity by a positive edge weight $\edgeweight_{\nodeidx,\nodeidx'}\!>\!0$.\footnote{The 
	notion of statistical similarity could be made precise using a \gls{probmodel} that interprets the \gls{datapoint}s 
	in each \gls{localdataset} $\localdataset{\nodeidx}$ as \gls{iid} draws from an underlying \gls{probdist} 
	$p^{(\nodeidx)}\big(\featurevec,\truelabel\big)$. The analysis of the statistical aspects of our method using 
	a \gls{probmodel} is beyond the scope of this paper.
}
The neighbourhood of a node $\nodeidx \in \nodes$ is $\neighbourhood{\nodeidx} \defeq \{ \nodeidx' \in \nodes: \{\nodeidx,\nodeidx'\} \in \edges\}$. 

Note that the undirected edges $\edge{\nodeidx}{\nodeidx'}$ of an empirical graph encode a symmetric 
notion of similarity between \gls{localdataset}s. If the \gls{localdataset} $\localdataset{\nodeidx}$ at node $\nodeidx$ 
is (statistically) similar to the \gls{localdataset} $\localdataset{\nodeidx'}$ at node $\nodeidx'$, then also the 
\gls{localdataset} $\localdataset{\nodeidx'}$ is (statistically) similar to the \gls{localdataset} $\localdataset{\nodeidx}$.  

The \gls{empgraph} of \gls{netdata} is a design choice which is guided by \gls{compasp} and \gls{statasp} of 
the resulting ML method. For example, using an \gls{empgraph} with a relatively small number of edges (``sparse graphs'') 
typically results in a smaller computational complexity. Indeed, the amount of computation required by the \gls{fl} 
methods developed in Section \ref{sec_fedrelax} is proportional to the number of edges in the \gls{empgraph}. 

On the other hand, the \gls{empgraph} should contain a sufficient number of edges between 
nodes that carry statistically similar \gls{localdataset}s (and corresponding learning tasks). 
Indeed, our FL method uses the edge density to adaptively pool \gls{localdataset}s into clusters 
of (approximately) homogeneous data. In contrast to graph clustering methods \cite{FlowSpecClustering2021,Luxburg2007,JungLocalGraphClustering}, 
our method steers this pooling based not solely on the edge connectivity but also on the 
geometry of local models at the nodes of the \gls{empgraph}. 

A main requirement of our method is a suitable choice for the edge set of the empirical graph. 
Domain-specific approaches to construct an empirical graph include spatio-temporal 
proximity of weather stations \cite{LocalizedLinReg2019} or quantifying the overlap of \gls{localdataset}s. 

The presence of an edge between two local datasets can also be tested by the effect of pooling them 
for the training of local models \cite{Ghosh2020}. Graph signal processing methods can be used to 
learn the empirical graph from curated data for which trained local models are provided \cite{Dong2019,Chepuri2017,MAL-078-3}. Being essentially 
a hyper-parameter of our \gls{fl} method, we could also choose the \gls{empgraph} using \gls{validation} techniques. 

%Consider an \gls{empgraph} whose nodes carry local loss functions $\locallossfunc{\nodeidx}{\localparams{\nodeidx}}$ 
%that form few clusters. The \gls{clusterempgraph} structure of local \gls{lossfunc}s is reflected by a high 
%density of edges between nodes in the same \gls{clusterempgraph} but few boundary edges between them. 
%It then seems reasonable to require a small variation of local parameter vectors $\localparams{\nodeidx}$ 
%across edges. 

\subsection{Networked Models}
\label{sec_net_models}

Consider \gls{netdata} with an \gls{empgraph} $\graph$ whose nodes $\nodeidx \in \nodes$ carry \gls{localdataset}s 
$\localdataset{\nodeidx}$. For each node $\nodeidx \in \nodes$, we wish to learn a useful \gls{hypothesis} 
$\learntlocalhypothesis{\nodeidx}$ from a local \gls{hypospace} $\localmodel{\nodeidx}$. The 
learnt \gls{hypothesis} should incur a small average \gls{loss} over a \gls{localdataset} $\localdataset{\nodeidx}$,
\begin{equation} 
\label{equ_def_local_lossfunc}
\hspace*{-3mm}\locallossfunc{\nodeidx}{\learntlocalhypothesis{\nodeidx}} \defeq \frac{1}{\localsamplesize{\nodeidx}} \sum_{\sampleidx=1}^{\localsamplesize{\nodeidx}} \loss{\big( \featurevec^{(\nodeidx,\sampleidx)}, \truelabel^{(\nodeidx,\sampleidx)} \big)}{\learntlocalhypothesis{\nodeidx}}.
\end{equation}

A collection of \gls{localmodel}s $\localmodel{\nodeidx}$, for each $\nodeidx \in \nodes$, 
constitutes a \gls{netmodel} $\netmodel{\graph}$ over the \gls{empgraph} $\graph$, 
\begin{equation} 
\netmodel{\graph}: \nodeidx \mapsto \localmodel{\nodeidx} \mbox{ for each node } \nodeidx \in \nodes. 
\end{equation}
In other words, a \gls{netmodel} is constituted by networked \gls{hypothesis} maps $\hypothesis \in \netmodel{\graph}$. 
Each such networked \gls{hypothesis} map assigns each node $\nodeidx \in \nodes$ a local \gls{hypothesis},
\begin{equation} 
\hypothesis: \nodeidx \mapsto \localhypothesis{\nodeidx} \in \localmodel{\nodeidx}. 
\end{equation}

It is important to note a \gls{netmodel} may combine different types of \gls{localmodel}s $\localmodel{\nodeidx}$. 
For example, $\localmodel{\nodeidx}$ might be a linear model, while $\localmodel{\nodeidx'}$ 
might be a \gls{decisiontree} for some other node $\nodeidx' \neq \nodeidx$. The only restriction we 
place on the choice of the local models is the availability of computational means (a \texttt{.fit()} function) to 
train them by (approximately) solving instances of \gls{rerm}.

\subsection{Generalized Total Variation}
\label{sec_gtv_measure}

In principle, we could train each local model $\localmodel{\nodeidx}$ separately on the 
corresponding \gls{localdataset} $\localdataset{\nodeidx}$ for each node $\nodeidx \in \nodes$. 
However, the \gls{localdataset}s might be too small to train a \gls{localmodel} which might be 
a (deep) \gls{ann} or a \gls{linmodel} with a large number of \gls{feature}s. As a remedy, we 
could try to pool \gls{localdataset}s with similar statistics to obtain a sufficiently large 
dataset to successfully train the \gls{localmodel}s $\localmodel{\nodeidx}$.   

The main theme of this paper is to use the network structure of the \gls{empgraph} $\graph$ to 
adaptively pool \gls{localdataset}s with similar statistical properties. We implement this pooling 
by requiring local models at well-connected nodes (clusters) to behave similar on a pre-specified 
test set of \gls{datapoint}s. To make this informal idea more precise, we next introduce a 
quantitative measure for the variation (or discrepancy) of local models across the (weighted) 
edges $\edgeidx \in \edges$ of the \gls{empgraph} $\graph$. 

Consider two nodes $\nodeidx, \nodeidx' \in \nodes$ in the \gls{empgraph} that are connected by 
an edge $\edge{\nodeidx}{\nodeidx'}$ with weight $\edgeweight_{\nodeidx,\nodeidx'}$. We define 
the variation between $\hypothesis^{(\nodeidx)}$ and $\hypothesis^{(\nodeidx')}$ by their discrepancy 
(between their \gls{prediction}s)
\begin{multline}
	\label{equ_def_variation_non_parametric}
	%\begin{split}
	\hspace*{-2mm}\variation{\nodeidx}{\nodeidx'}{\localhypothesis{\nodeidx}}{\localhypothesis{\nodeidx'}} \defeq \\ 
	\frac{1}{\samplesize^{(pub)}} \hspace{-2mm} \sum_{\substack{\sampleidx=1 \\ \featurevec^{(\sampleidx)} \in \publicds{\nodeidx'}}}^{\samplesize^{(pub)}} \hspace{-3mm}  \gtvloss{\featurevec^{(\sampleidx)}}{\hypothesis^{(\nodeidx)}\big(\featurevec^{(\sampleidx)}\big)}{\hypothesis^{(\nodeidx')}\big(\featurevec^{(\sampleidx)}\big)}
%	&\hspace*{10mm} \!+\!\gtvloss{\pair{\featurevec^{(\sampleidx)}}{\hypothesis^{(\nodeidx')}\big(\featurevec^{(\sampleidx)}\big)}}{\hypothesis^{(\nodeidx)}} \bigg]
	%\end{split}
\end{multline}  
on a (public) dataset shared by node $\nodeidx'$  
\begin{equation}
	\label{equ_test_set}
	\publicds{\nodeidx'} = \left\{ \featurevec^{(1)},\ldots, \featurevec^{(\samplesize^{(pub)})}\right\}.
\end{equation}
where $\samplesize^{(pub)} = |\publicds{\nodeidx'}|$, is a sample size of the public dataset.
The dataset \eqref{equ_test_set}, which consists of $\samplesize^{(pub)}$ \gls{feature} vectors, 
must be shared with each neighbouring node $\nodeidx \in \neighbourhood{\nodeidx'}$ in the \gls{empgraph}. 

We then define the \gls{gtv} of a networked \gls{hypothesis} $\hypothesis \in \netmodel{\graph}$ by summing the discrepancy \eqref{equ_def_variation_non_parametric} over all nodes $\nodes$, 
\begin{align}
	\label{equ_def_gtv_non_param}
	\gtv{\hypothesis} \defeq  \sum_{\nodeidx \in \nodes} \sum_{\nodeidx' \in \neighbourhood{i}} \hspace{-2mm} \edgeweight_{\nodeidx,\nodeidx'} \variation{\nodeidx}{\nodeidx'}{\localhypothesis{\nodeidx}}{\localhypothesis{\nodeidx'}}. 
\end{align}
Note that $\gtv{\hypothesis}$ is parametrized by the choice of the \gls{lossfunc} 
$$\gtvloss{\cdot}{\cdot}{\cdot}: \mathbb{R}^{\nrfeatures}\times \mathbb{R} \times \mathbb{R} \rightarrow \mathbb{R},$$
used to compute the discrepancy $\variation{\nodeidx}{\nodeidx'}{\localhypothesis{\nodeidx}}{\localhypothesis{\nodeidx'}}$ with \eqref{equ_def_variation_non_parametric}. 
This \gls{lossfunc} depends on $\localhypothesis{\nodeidx}, \localhypothesis{\nodeidx'}$ 
only via their predictions $\localhypothesis{\nodeidx} \big( \featurevec^{(\sampleidx)}\big), \localhypothesis{\nodeidx'} \big( \featurevec^{(\sampleidx)}\big)$.  
The \gls{lossfunc} $L^{(\rm d)}$ in \eqref{equ_def_variation_non_parametric} 
might be very different from local \gls{lossfunc} \eqref{equ_def_local_lossfunc} 
used to measure the \gls{prediction} error of a local \gls{hypothesis} $\localhypothesis{\nodeidx}$. 
However, it might be useful to use similar \gls{lossfunc}s in \eqref{equ_def_variation_non_parametric} 
and \eqref{equ_def_local_lossfunc} (see Section \ref{equ_def_mod_agn_fredreg}).

Using \gls{gtv} \eqref{equ_def_gtv_non_param} as a \gls{regularizer} is beneficial if the edges in 
the \gls{empgraph} align with similarities in the statistical properties of \gls{localdataset}s. 
This informal requirement can be made precise using various forms of \gls{clustasspt}s \cite{SemiSupervisedBook,nflarxiv2022}. 

\section{A Model Agnostic FL Method}
\label{sec_fedrelax}

Consider networked data modelled by an \gls{empgraph}. It is natural to learn a local \gls{hypothesis} $\localhypothesis{\nodeidx}$ for each node $\nodeidx \in \nodes$ by balancing the local \gls{lossfunc} with the \gls{gtv} term in Eq. \eqref{equ_def_gtv_non_param}.  
The precise formulation of this balancing is given by the \gls{gtvmin} optimization problem:
\begin{equation} 
	\label{equ_def_gtv_min}
	\min_{\hypothesis \in \netmodel{\graph}}  \underbrace{\sum_{\nodeidx \in \nodes} \bigg[ 
	\locallossfunc{\nodeidx}{\localhypothesis{\nodeidx}}\!+\! \regparam \hspace*{-2mm}\sum_{\nodeidx' \in \neighbourhood{\nodeidx}} \hspace*{-2mm} \edgeweight_{\nodeidx,\nodeidx'} \variation{\nodeidx}{\nodeidx'}{\localhypothesis{\nodeidx}}{\localhypothesis{\nodeidx'}} \bigg]}_{\defeq f(\localhypothesis{1},\ldots,\localhypothesis{\nrnodes})}.
\end{equation} 
Note that \gls{gtvmin} \eqref{equ_def_gtv_min} is an instance of the \gls{rerm} principle. Specifically, the (aggregate) local \gls{lossfunc} can be interpreted as the \gls{trainerr} of the networked $\hypothesis \in \netmodel{\graph}$, while the \gls{gtv} term serves as a \gls{regularizer}.% \cite[Ch. 4]{MLBasics}.

We use block-coordinate minimization (BCM) \cite{ParallelDistrBook,CvxAlgBertsekas} to solve \gls{gtvmin} \eqref{equ_def_gtv_min}. BCM decomposes the global objective into smaller, node-wise subproblems. Each local model $\localhypothesis{\nodeidx}$ is updated iteratively while keeping others fixed, leveraging the structure of the graph-based regularization term. This allows for efficient, decentralized updates, where each node optimizes its loss function while ensuring similarity with neighboring models. 

Specifically, given some local \gls{hypothesis} maps $\estlocalhypositer{\nodeidx}{\iteridx}$, we compute the updated (and hopefully improved) local \gls{hypothesis} maps $\estlocalhypositer{\nodeidx}{\iteridx+1}$ for all nodes $\nodeidx\in \nodes$ by minimizing $f(\hypothesis)$ along $\localhypothesis{\nodeidx}$ while keeping the other local \gls{hypothesis} maps fixed:
\begin{equation} 
	\label{equ_def_coord_min_update}
	\begin{split}
	\estlocalhypositer{\nodeidx}{\iteridx+1} & \! \in \argmin_{\localhypothesis{\nodeidx} \in \localmodel{\nodeidx}}f\bigg( \estlocalhypositer{1}{\iteridx},\ldots,\estlocalhypositer{\nodeidx-1}{\iteridx},\localhypothesis{\nodeidx},\estlocalhypositer{\nodeidx+1}{\iteridx},\ldots \bigg) \\ 
%	& \stackrel{\eqref{equ_GTVMin_MOCHA_coordmin}}{=}  \argmin_{\localparams{\nodeidx} \in \mathbb{R}^{\dimlocalmodel}} \localobj{\nodeidx}{ \estlocalparamsiter{1}{\iteridx},\ldots,\estlocalparamsiter{\nodeidx-1}{\iteridx},\localparams{\nodeidx},\estlocalparamsiter{\nodeidx+1}{\iteridx},\ldots}  \nonumber \\ 
	& \! \stackrel{\eqref{equ_def_gtv_min}}{=}  \argmin_{{\localhypothesis{\nodeidx} \in \localmodel{\nodeidx}}} \locallossfunc{\nodeidx}{\localhypothesis{\nodeidx}} \! + \! \regparam \hspace{-2.2mm} \sum_{\nodeidx' \in \neighbourhood{\nodeidx}} \hspace{-2.5mm} \edgeweight_{\nodeidx,\nodeidx'} \variation{\nodeidx}{\nodeidx'}{\localhypothesis{\nodeidx}}{\estlocalhypositer{\nodeidx'}{\iteridx}} \hspace{-.4mm}.
	\end{split}
\end{equation} 

We obtain Algorithm \ref{alg_fed_relax_nonparam} by iterating \eqref{equ_def_coord_min_update} 
simultaneously at all nodes $\nodeidx \in \nodes$ until a \gls{stopcrit} is met. Examples of a \gls{stopcrit} 
include reaching a pre-specified number of iterations or monitoring the decrease in local \gls{lossfunc}. 

\begin{algorithm}[htbp]
	\caption{FedRelax}
	\label{alg_fed_relax_nonparam}
	{\bf Input}: empirical graph $\graph$ with edge weights $\edgeweight_{\nodeidx,\nodeidx'}$,
	local \gls{lossfunc}s $\locallossfunc{\nodeidx}{\cdot}$ for $\nodeidx \in \nodes$, public shared dataset $\mathcal{P}= \{ \featurevec^{(1)},\ldots, \featurevec^{(\samplesize^{(pub)})} \}$, \gls{gtv} parameter $\regparam$, discrepancy measure $\delta\big(\cdot, \cdot \big)$ \\
	{\bf Initialize}: $\iteridx \leftarrow 0$, $\estlocalhypositer{\nodeidx}{0}$ for all nodes $\nodeidx \in \nodes$ 
	%	using \eqref{eq:14}; %$\mQ^{(i)}\!\defeq\!\big(\mX^{(i)}\big)^{T} \mX^{(i)}$; $\tilde{\vy}^{(i)}\!\defeq\!\big(\mX^{(i)}\big)^{T} \vy^{(i)}$  \\
	\begin{algorithmic}[1]
		\While{stopping criterion is not satisfied}
		\For{all nodes $\nodeidx \in \nodes$ in parallel}
		\State share the predictions $\{ \estlocalhypositer{\nodeidx}{\iteridx} \big(\featurevec \big) \}_{\featurevec \in \publicds{\nodeidx}}$
		with neighbours $\nodeidx' \in \neighbourhood{\nodeidx}$ 
		\State \label{equ_update_fed_relax} update the local \gls{hypothesis} $\estlocalhypositer{\nodeidx}{\iteridx}$ by
		\begin{equation} 
			\label{equ_update_step_fed_relax}
			\hspace*{-6mm}	\estlocalhypositer{\nodeidx}{\iteridx+1} \!\in\!  \argmin_{\localhypothesis{\nodeidx} \in \localmodel{\nodeidx}} \hspace*{-1mm} \locallossfunc{\nodeidx}{\localhypothesis{\nodeidx}} \!+\! \regparam\hspace*{-3mm} \sum_{\nodeidx' \in \neighbourhood{\nodeidx}}\hspace*{-3mm} \edgeweight_{\nodeidx,\nodeidx'}  \variation{\nodeidx}{\nodeidx'}{\localhypothesis{\nodeidx}}{\estlocalhypositer{\nodeidx'}{\iteridx}}. 
		\end{equation}
		\EndFor
		\State $\iteridx \leftarrow \iteridx\!+\!1$
		\EndWhile
	\end{algorithmic}
	{\bf Output}: 	$\estlocalhypositer{\nodeidx}{\iteridx} $ for all nodes $\nodeidx \in \nodes$
\end{algorithm}

The main computational work of Algorithm \ref{alg_fed_relax_nonparam} is done in step \eqref{equ_update_fed_relax}. 
This step is an instance of \gls{rerm} for the local model $\localmodel{\nodeidx}$ at each node 
$\nodeidx \in \nodes$. The \gls{regularization} term in this \gls{rerm} instance is the weighted 
sum of the discrepancies \eqref{equ_def_variation_non_parametric} between the predictions 
of the local \gls{hypothesis} map $\localhypothesis{\nodeidx}$ and the predictions of the 
current local \gls{hypothesis} maps $\localhypothesis{\nodeidx'}$ at neighbouring nodes $\nodeidx' \in \neighbourhood{\nodeidx}$. 

\subsection{Model Agnostic Federated Least-Squares Regression} 
\label{equ_def_mod_agn_fredreg} 

Note that Algorithm \ref{alg_fed_relax_nonparam} is parametrized by the choices for the \gls{lossfunc} 
used to measure the \gls{trainerr} \eqref{equ_def_local_lossfunc} %of a local \gls{hypothesis} $\learntlocalhypothesis{\nodeidx}$ 
and the \gls{lossfunc} used to measure the discrepancy \eqref{equ_def_variation_non_parametric} 
between the \gls{localmodel}s across the edge $\edge{\nodeidx}{\nodeidx'} \in \edges$.

A popular choice for the \gls{lossfunc} when predicting numeric labels is the \gls{sqerrloss}
\begin{equation} 
\label{equ_squared_loss}
\loss{\big((\featurevec,\truelabel),h\big)} \defeq \big(\truelabel - \underbrace{h(\featurevec)}_{=\predictedlabel} \big)^{2}. 
\end{equation} 

Note that the update \eqref{equ_update_step_fed_relax} is nothing but \gls{rerm} for 
learning a local \gls{hypothesis} $\localhypothesis{\nodeidx} \in \localmodel{\nodeidx}$ from the 
\gls{localdataset} $\localdataset{\nodeidx}$. When using the \gls{sqerrloss} in \eqref{equ_def_variation_non_parametric}, the regularization term in \eqref{equ_update_step_fed_relax} is the average \gls{sqerrloss} incurred on the (``pseudo-'') labeled test set (see \eqref{equ_test_set}) 
\begin{equation}
 \hspace*{-3mm}	\bigcup_{\substack{\nodeidx' \in \neighbourhood{\nodeidx} \\ \featurevec^{(r)} \in \publicds{\nodeidx'} }} \hspace*{-2mm} \left\{ \pair{\featurevec^{(1)}}{\estlocalhypositer{\nodeidx'}{\iteridx}\big(\featurevec^{(1)}\big)},\ldots, \pair{\featurevec^{(\samplesize^{(pub)})}}{\estlocalhypositer{\nodeidx'}{\iteridx}\big(\featurevec^{(\samplesize^{(pub)})}\big)} \right\}.
\end{equation} 
% Is it possible to get this to fit into one line?

\subsection{Parametric Model Agnostic Federated Learning} 
\label{sec_param_model_agnostic} 

We now apply Algorithm \ref{alg_fed_relax_nonparam} to train a parametric 
\gls{netmodel} $\netmodel{\graph}$ with each \gls{localmodel} $\localmodel{\nodeidx}$ 
parametrized by a local \gls{modelparams} $\localparams{\nodeidx} \in \mathbb{R}^{\featuredim}$. 
The common dimension $\featuredim$ of all \gls{localmodel}s is identical to the length of 
the \gls{feature} vectors in the shared public datasets $\publicds{\nodeidx}$ \eqref{equ_test_set}. 
For every node $\nodeidx \in \nodes$, any \gls{hypothesis} map $\localhypothesis{\nodeidx} = \hypothesis^{(\localparams{\nodeidx})}$ 
in $\localmodel{\nodeidx}$ is determined by a specific choice of local \gls{modelparams} $\localparams{\nodeidx}$. 

The usefulness of a specific choice for $\localparams{\nodeidx}$ is measured by the 
local \gls{lossfunc} $\locallossfunc{\nodeidx}{\localparams{\nodeidx}} = \locallossfunc{\nodeidx}{\hypothesis^{(\localparams{\nodeidx})}}$. 
We measure the discrepancy \eqref{equ_def_variation_non_parametric} between local \gls{hypothesis} 
maps $\hypothesis^{(\localparams{\nodeidx})}$, $\hypothesis^{(\localparams{\nodeidx'})}$ across 
$\edge{\nodeidx}{\nodeidx'} \in \edges$ using the \gls{sqerrloss}, 
\begin{equation}
	\label{equ_def_discr_parametric}
\gtvloss{\featurevec}{\localhypothesis{\nodeidx}\big(\featurevec\big)}{\localhypothesis{\nodeidx'}\big(\featurevec\big)} \defeq \big( \featurevec^{T}  \big( \localparams{\nodeidx}-\localparams{\nodeidx'} \big) \big)^{2}. 
\end{equation}

Inserting \eqref{equ_def_discr_parametric} into \eqref{equ_def_variation_non_parametric} yields
\begin{equation}
\label{equ_def_variation_parametric}
\begin{split}
\variation{\nodeidx}{\nodeidx'}{\localparams{\nodeidx}}{\localparams{\nodeidx'}} & = \variation{\nodeidx}{\nodeidx'}{\localhypothesis{\localparams{\nodeidx}}}{\localhypothesis{\localparams{\nodeidx'}}} \\ 
& \hspace*{-17mm} =
\frac{1}{\samplesize^{(pub)}} \hspace{-3mm} \sum_{\substack{\sampleidx=1 \\ \featurevec^{(\sampleidx)} \in \publicds{\nodeidx'}}}^{\samplesize^{(pub)}} \hspace{-3mm} \bigg( \big( \localparams{\nodeidx} \big)^{T} \featurevec^{(\sampleidx)}\!-\!\big( \localparams{\nodeidx'} \big)^{T} \featurevec^{(\sampleidx)}  \bigg)^{2} \hspace{-1mm}. 
%& \hspace*{-15mm} = \frac{1}{\samplesize'} \big( \localparams{\nodeidx}\!-\!\localparams{\nodeidx'} \big)^{T} \mX^{'}  \big( \mX^{'} \big)^{T} \big( \localparams{\nodeidx}\!-\!  \localparams{\nodeidx'} \big).
\end{split} 
\end{equation} 
Here, we used the \gls{feature} matrix $\mX' = \big( \featurevec^{(1)},\ldots,\featurevec^{(\samplesize^{(pub)})} \big) \in \mathbb{R}^{\featuredim \times \samplesize^{(pub)}}$ whose columns are the \gls{feature} vectors in the public dataset $\publicds{\nodeidx'}$ \eqref{equ_test_set}. 
It is important to note that \eqref{equ_def_variation_parametric} depends on the local \gls{modelparams} 
only by their predictions $\big( \localparams{\nodeidx} \big)^{T} \featurevec^{(\sampleidx)}$, $\big( \localparams{\nodeidx'} \big)^{T} \featurevec^{(\sampleidx)}$ for the \gls{datapoint}s in the public dataset \eqref{equ_test_set}. 

We obtain Algorithm \ref{alg_fed_relax_param} from Algorithm \ref{alg_fed_relax_nonparam} by 
inserting \eqref{equ_def_variation_parametric} into \eqref{equ_update_step_fed_relax}. We also obtain 
the corresponding special case of \gls{gtvmin}, that is solved by Algorithm \ref{alg_fed_relax_param}, 
by inserting \eqref{equ_def_variation_parametric} into \eqref{equ_def_gtv_min},  
\begin{align} 
	\label{equ_def_gtv_min_param}
	%\begin{split}
	& \min_{\{ \localparams{\nodeidx} \in \mathbb{R}^{\nrfeatures}\}_{\nodeidx \in \nodes}} \sum_{\nodeidx \in \nodes} \bigg[  \locallossfunc{\nodeidx}{\localparams{\nodeidx}}  \\ 
	& + \frac{\regparam}{\samplesize^{(pub)}}
	\hspace*{-2.5mm}\sum_{\nodeidx' \in \neighbourhood{\nodeidx}} \hspace*{-2.5mm} \edgeweight_{\nodeidx,\nodeidx'} \hspace{-4mm} 
	\sum_{\substack{\sampleidx=1 \\ \featurevec^{(\sampleidx)} \in \publicds{\nodeidx'}}}^{\samplesize^{(pub)}} \hspace{-3.5mm} 
	\bigg( \big( \localparams{\nodeidx} \big)^{T} \featurevec^{(\sampleidx)}\!-\!\big( \localparams{\nodeidx'} \big)^{T} \featurevec^{(\sampleidx)}  \bigg)^{2} \bigg]. \nonumber
	%\end{split}
\end{align} 

\begin{algorithm}[htbp]
	\caption{FedRelax for Parametric Models}
	\label{alg_fed_relax_param}
	{\bf Input}: \gls{empgraph} $\graph$ with edge weights $\edgeweight_{\nodeidx,\nodeidx'}$,
	local \gls{lossfunc}s $\locallossfunc{\nodeidx}{\cdot}$, public dataset $\mathcal{P} = \{ \featurevec^{(1)},\ldots, \featurevec^{(\samplesize^{(pub)})} \}$, \gls{gtv} parameter $\regparam$ \\
	{\bf Initialize}: $\iteridx\!\defeq\!0$, $\estlocalparamsiter{\nodeidx}{0}$ for all nodes $\nodeidx \in \nodes$ 
	\begin{algorithmic}[1]
		\While{stopping criterion is not satisfied}
		\For{all nodes $\nodeidx \in \nodes$ in parallel}
		\State share the predictions $\left\{  \big( 	\estlocalparamsiter{\nodeidx'}{\iteridx} \big)^{T} \featurevec \, \right\}_{\featurevec \in \publicds{i}}$
		with the neighbours $\nodeidx' \in \neighbourhood{\nodeidx}$ 
		\State \label{equ_update_fed_relax} update the local \gls{modelparams} $\estlocalparamsiter{\nodeidx}{\iteridx} $ by
		\begin{align} 
			\label{equ_update_step_fed_relax_param}
			&\estlocalparamsiter{\nodeidx}{\iteridx+1} \!\in\!  \argmin_{\localparams{\nodeidx} \in \mathbb{R}^{\featuredim}} \locallossfunc{\nodeidx}{\localparams{\nodeidx}} \\ 
			&+\! \frac{\regparam}{\samplesize^{(pub)}} \hspace*{-2mm} 
			\sum_{\nodeidx' \in \neighbourhood{\nodeidx}} \hspace{-2.5mm} \edgeweight_{\nodeidx,\nodeidx'} \hspace{-4mm} 
			\sum_{\substack{{\sampleidx=1} \\ \featurevec^{(\sampleidx)} \in \publicds{\nodeidx'}}}^{\samplesize^{(pub)}} \hspace{-4.3mm} 
			\bigg( \big( \localparams{\nodeidx} \big)^{T} \featurevec^{(\sampleidx)}\!-\!\big(	\estlocalparamsiter{\nodeidx'}{\iteridx} \big)^{T} \featurevec^{(\sampleidx)} \hspace{-.5mm} \bigg)^{2}. \nonumber
		\end{align}
		\EndFor
		\State $\iteridx  \leftarrow \iteridx\!+\!1$
		\EndWhile
	\end{algorithmic}
	{\bf Output}: 	$\estlocalhypositer{\nodeidx}{\iteridx} $ for all nodes $\nodeidx \in \nodes$
\end{algorithm}

\section{Numerical Experiments}
\label{sec:num_exps}

We now present illustrative numerical experiments to study the computational and \gls{statasp} of Algorithms \ref{alg_fed_relax_nonparam} and \ref{alg_fed_relax_param}. The code for these experiments is available at \href{https://github.com/shamPJ/het-FL/tree/main/src}{GitHub}. Section \ref{sec_exp_clustered_ds} presents an experiment that applies Algorithm \ref{alg_fed_relax_param} to a clustered dataset using parametric linear models implemented in PyTorch. Additionally, Algorithm \ref{alg_fed_relax_nonparam} is applied to the same dataset but with non-parametric decision tree regressors as local models. We also evaluate our approach on heterogeneous, non-i.i.d. datasets in Section \ref{sec_exp_noniid_ds}.

\subsection{Synthetic Dataset - Clustered setting}
\label{sec_exp_clustered_ds}

Experiments were performed on a synthetic dataset, whose empirical graph $\mathcal {G}$,  consisting of 150 nodes, is partitioned into three equal-sized clusters (unless specified otherwise), denoted as $\mathcal{C} = \{\mathcal{C}^{(1)}, \mathcal{C}^{(2)},  \mathcal{C}^{(3)} \}$, where $|\mathcal{C}^{(1)}|=|\mathcal{C}^{(2)}|=|\mathcal{C}^{(3)}|=50$. We denote the cluster assignment of node $i \in \mathcal{V}$ by ${c}^{(i)} \in \{1,2,3\}$. The edges in $\mathcal {G}$  are generated as realizations of independent binary random variables ${b}_{i,{i}^{'}} \in \{0,1\}$. These random variables are indexed by pairs $(i,{i}^{'})$ of nodes, where an edge $\{i,{i}^{'}\} \in \mathcal{E}$ exists if and only if ${b}_{i,{i}^{'}}=1$.
 
Two nodes in the same cluster are connected with probability $\text{Prob}\{{b}_{i,{i}^{'}}=1\} :={p}_{in}=0.8$ if nodes $i \text{ and } {i}^{'}$ belong to the same cluster. In contrast, they are connected with probability $\text{Prob}\{{b}_{i,{i}^{'}}=1\} :={p}_{out}=0.2$ if they belong to different clusters. Every edge in  $\mathcal {G}$ has the same weight, given by ${A}_{\nodeidx, \nodeidx'}=1$ for all $\{\nodeidx,\nodeidx'\} \in \mathcal{E}$.

Each node $i \in \mathcal {V}$ of the empirical graph $\mathcal {G}$ holds a local dataset $\mathcal {D}^{(i)}$ of the form $\mathcal {D}^{(i)} := \{ (\mathbf{x}^{(i,1)}, {y}^{(i,1)}), \dots, (\mathbf{x}^{(i,{m}_{i})}, {y}^{(i,{m}_{i})}) \}$. Thus, the dataset $\mathcal {D}^{(i)}$ consists of ${m}_{i}$ data points, each characterized by a feature vector $\mathbf{x}^{(i,r)} \in \mathbb{R}^{\featuredim}$ and scalar label ${y}^{(i,r)}$, for $r=1,\dots,{m}_{i}$. 

The feature vectors $\mathbf{x}^{(i,r)} \sim \mathcal{N}(\mathbf{0},\mathbf{I}_{\featuredim})$ are drawn i.i.d. from a standard multivariate normal distribution. The labels of the data points are generated by a noisy linear model
\begin{equation}
\label{equ_def_true_linear_model_SBM}
{y}^{(\nodeidx,r)} = (\overline{\weights}^{(c^{(\nodeidx)})})^T\mathbf{x}^{(\nodeidx,r)} + {\varepsilon}^{(\nodeidx,r)}
\end{equation}

The noise values ${\varepsilon}^{(\nodeidx,r)} \sim \mathcal{N}(0, \sigma^2)$ are i.i.d. realizations of a normal distribution for $i \in \mathcal{V}$ and $r=1,\dots,{m}_{i}$. The true underlying vectors $\overline{\weights}^{(c^{(i)})} \sim \mathcal{N}(\mathbf{0},\mathbf{I}_d)$ are drawn from a standard normal distribution and are the same for nodes within the same cluster, i.e., if ${c}^{(\nodeidx)}={c}^{(\nodeidx')}$, then $\overline{\weights}^{{c}^{(\nodeidx)}}=\overline{\weights}^{{c}^{(\nodeidx')}}$. In addition to the local training dataset of size $m_i$, we generate validation subset of size ${m}^{(val)}_{\nodeidx}=100$ data points from the same distribution.

In principle, the shared, unlabeled public dataset $\publicds{\nodeidx}$ can be chosen arbitrarily; 
i.e. it can be the same for all nodes, different across nodes, or sampled from different probability 
distributions. If privacy constraints allow, the shared public dataset may be a subset of the local 
datasets. In the experiments below, the public datasets $\publicds{\nodeidx}$ consist of data points 
with feature vectors $\mathbf{x}^{(i,r)} \sim \mathcal{N}(\mathbf{0},\mathbf{I}_{\featuredim})$ 
drawn i.i.d. from the standard multivariate normal distribution, separately for each node. The sample size  is $m^{(\mathcal{P})}=100$. 

\vspace{1.5mm}
\subsubsection{Pytorch linear model}

To learn the local parameters $\mathbf{w}^{(i)}$, we use Algorithm \ref{alg_fed_relax_param}, with the loss described in Eq.(\ref{equ_update_step_fed_relax_param}), implemented using the PyTorch Python library. Each local model consists of one linear layer, uses MSE as the local loss $\locallossfunc{\nodeidx}{\localparams{\nodeidx}} $ and employs the RMSprop optimizer with a learning rate of 0.01.

As input to the algorithm, we generate a synthetic dataset as described above. We use varying dimensionality $\dimlocalmodel  \in \{2,10,20,50,100 \}$ for the feature vector to test whether our algorithm improves performance for a high $\frac{\dimlocalmodel}{m}$ ratio (i.e., a small local dataset size combined with a large number of features). As the stopping criterion in Algorithm \ref{alg_fed_relax_param}, we use a fixed number of $K = 500$ iterations. 

In Figure \ref{fig:param_clustered}, we plot the MSE values averaged across all nodes and five runs of the experiment. In each run, new local datasets $\mathcal{D}^{(i)}$, shared public datasets $\publicds{\nodeidx}$, and edges between the nodes are generated. 

We measure the performance of the learned local parametric model by computing the squared difference between the learnt model's parameter vector $\estlocalparams{\nodeidx}$ and the true cluster weight vector $\overline{\weights}^{(c^{(i)})}$:
\begin{equation}
		\label{equ_def_MSE}
{\rm MSE}^{(\widehat{\weights})} \defeq  \frac{1}{\dimlocalmodel} \normgeneric{ \widehat{\weights}^{(i)} -  \overline{\weights}^{(c^{(i)})}} {2}^{2}.
\end{equation}

Figure \ref{fig:param_clustered_err_dm} shows ${\rm MSE}^{(\widehat{\weights})}$ as defined in \eqref{equ_def_MSE}  averaged across all nodes and five repetitions of the experiment with different values of the regularization parameter: $\alpha=0, 0.01, 0.05$. With the regularization parameter set to zero ($\alpha=0$), the problem reduces to training the model only locally.
The local sample size is fixed at $\localsamplesize{\nodeidx}=10$ and we vary dimensionality $\dimlocalmodel$ of the feature vector to achieve different $\frac{\dimlocalmodel}{m}$ ratios. When generating the local datasets,  the noise level is set to zero  ($\sigma^2=0)$. 
From Figure \ref{fig:param_clustered_noise}, we can see that in the noiseless case, FedRelax is most beneficial for the local datasets with $\frac{\dimlocalmodel}{m} \geq 1$.

\begin{figure}[htbp]
	\centering
	\begin{tikzpicture}[remember picture]
	\begin{groupplot}[
		group style={
			group size=3 by 3,  % 2 column, 2 rows
			horizontal sep=0.1cm,
			vertical sep=1.5cm,    % Space between the two plots
			y descriptions at=edge left,
			ylabels at=edge left,
			yticklabels at=edge left,
			x descriptions at=edge bottom,
			xlabels at=edge bottom,
			xticklabels at=edge bottom,
			group name=myplots
			},
		height=4cm,
		width=0.22\textwidth,
		ymin=0,
		ymax=1.2,
		ylabel={\normalsize $MSE^{(\mathbf{w})}$}, % Only appears on the first plot
		ylabel style={align=center, yshift=0cm}, 
		ticklabel style={font=\footnotesize},
		cycle list name=mycolors,
		legend style={draw=none, at={(-0.5,-0.1)}, anchor=north, nodes={scale=0.55}, legend columns=-1, font=\Large},
		legend image post style={scale=0.4},
		]
		
		% First row
		\nextgroupplot[title={\small $\alpha=0$}, axis lines=left, axis line style={-}]
		\pgfplotsforeachungrouped \i/\j in {1/0.2, 2/1, 4/5}{
			\addplot+ [mark=none, very thick, solid] table [
			x index=0, 
			y index=\i,
			col sep=comma
			]{Fedrelax_dm_ratio.csv};
		}
		\nextgroupplot[title={\small $\alpha=0.01$}, axis lines=left, axis line style={-}]
		\pgfplotsforeachungrouped \i/\j in {6/0.2, 7/1, 9/5}{
			\addplot+ [mark=none, very thick] table [
			x index=0, 
			y index=\i,
			col sep=comma
			] {Fedrelax_dm_ratio.csv};
		}
		\nextgroupplot[title={\small $\alpha=0.05$}, axis lines=left, axis line style={-}]
		\pgfplotsforeachungrouped \i/\j in {11/0.2, 12/1, 14/5}{
			\addplot+ [mark=none, very thick] table [
			x index=0, 
			y index=\i,
			col sep=comma
			] {Fedrelax_dm_ratio.csv};
			\edef\temp{\noexpand\addlegendentry{$\dimlocalmodel /\/ \localsamplesize{\nodeidx} = \j$}}
			\temp
		}
		
		% Second row
		\nextgroupplot[axis lines=left, axis line style={-}]
		\pgfplotsforeachungrouped \i/\j in {1/0, 3/1, 4/5}{
			\addplot+ [mark=none, very thick, solid] table [
			x index=0, 
			y index=\i,
			col sep=comma
			] {Fedrelax_noise.csv};
		}
		\nextgroupplot[axis lines=left, axis line style={-}]
		\pgfplotsforeachungrouped \i/\j in {5/0, 7/1, 8/5}{
			\addplot+ [mark=none, very thick] table [
			x index=0, 
			y index=\i,
			col sep=comma
			] {Fedrelax_noise.csv};
		}
		\nextgroupplot[axis lines=left, axis line style={-}]
		\pgfplotsforeachungrouped \i/\j in {9/0, 11/1, 12/5}{
				\addplot+ [mark=none, very thick] table [
				x index=0, 
				y index=\i,
				col sep=comma
				] {Fedrelax_noise.csv};
				\edef\temp{\noexpand\addlegendentry{$\sigma$ = \j}}
				\temp
			}
		
		% Third row
		\nextgroupplot[axis lines=left, axis line style={-}, ymax=1.1]
		\pgfplotsforeachungrouped \i in {1, 2, 3, 4, 5}{
			\addplot+ [mark=none, very thick, solid] table [
			x index=0, 
			y index=\i,
			col sep=comma
			] {Fedrelax_benchmark.csv};
		}
		\nextgroupplot[xlabel={\small $k$}, axis lines=left, axis line style={-}, ymax=1.1]
		\pgfplotsforeachungrouped \i in {6,7,8,9,10}{
			\addplot+ [mark=none, very thick] table [
			x index=0, 
			y index=\i,
			col sep=comma
			] {Fedrelax_benchmark.csv};
		}
		\nextgroupplot[legend entries={$FedRelax$, $Oracle$, $IFCA 2$, $IFCA 5$, $FedAvg$}, 
			legend style={draw=none, at={(-0.5,-0.4)}, anchor=north, legend columns=-1, font=\Large}, 
			axis lines=left, axis line style={-}, ymax=1.1]
		\pgfplotsforeachungrouped \i in {11,12,13,14,15}{
			\addplot+ [mark=none, very thick] table [
			x index=0, 
			y index=\i,
			col sep=comma
			] {Fedrelax_benchmark.csv};
		}
	\end{groupplot}
	\tikzset{SubCaption/.style={
			text width=2in, yshift=0mm, align=center, anchor=north, outer sep=0.5cm,
	}}
	\tikzset{SubCaptionBottom/.style={
		text width=3in, yshift=-2mm, align=center, anchor=north, outer sep=1cm,
	}}
	\node[SubCaption] at ($(myplots c1r1.south)!0.5!(myplots c3r1.south)$) {\subcaption{\mbox{FedRelax w.r.t. $d/m$ ratio.}\label{fig:param_clustered_err_dm}}};
	\node[SubCaption] at ($(myplots c1r2.south)!0.5!(myplots c3r2.south)$) {\subcaption{\mbox{FedRelax  w.r.t. $\sigma$, $d/m=1$.}\label{fig:param_clustered_noise}}};
	\node[SubCaptionBottom] at ($(myplots c1r3.south)!0.5!(myplots c3r3.south)$) {\subcaption{\mbox{FedRelax vs other FL algorithms, $d/m=5$.} \label{fig:param_clustered_benchmark}}};
\end{tikzpicture}
	\caption{FedRelax ${\rm MSE}^{(\widehat{\weights})}$ on a clustered dataset with linear regressors as models. The code to generate the figure is available at \href{https://github.com/shamPJ/het-FL/blob/main/src/FedRelax.ipynb}{Github}.}
	\label{fig:param_clustered}
\end{figure}

The denoising effect of Algorithm \ref{alg_fed_relax_param} can be seen in 
Figure \ref{fig:param_clustered_noise}: although FedRelax only slightly accelerates 
the convergence to the correct cluster weight vector $\overline{\mathbf{w}}^{(c^{(i)})}$ 
for $\frac{\dimlocalmodel}{m}=1$ in the noiseless scenario, it prevents divergence 
of gradient descent in the high-noise settings ($\sigma=5$).

We then benchmark FedRelax against the Oracle model, FedAvg, and IFCA algorithms. We generated noiseless ($\sigma=0$) synthetic clustered datasets forming 5 clusters with 30 nodes per cluster ($|\nodes|=150$). As before, the sample size $m_i$ of local datasets is set to $m_i=10$, and the dimensionality of the feature vectors is set to $\dimlocalmodel=50$ ($\frac{\dimlocalmodel}{m}=5$). The initialization of weight vectors and the learning rate for FedAvg and IFCA follow the same procedure as PyTorch linear regressors, where $\mathbf{w}^{(i)} \sim \mathcal{U}(-\sqrt{1/\dimlocalmodel},\sqrt{1/\dimlocalmodel})$ and the learning rate is set to 0.01.

Figure \ref{fig:param_clustered_benchmark} shows that although FedRelax does not reach the level of Oracle performance (i.e., models trained on all 50 local datasets belonging to the same cluster), it significantly reduces MSE compared to training the models locally ($\alpha=0$).

IFCA is a two-step algorithm (a clustering step and a gradient aggregation step) that is well 
suited for clustered datasets. It requires the number of clusters to be specified as input. We 
can see in Figure \ref{fig:param_clustered_benchmark} (IFCA correct) that when provided 
the correct number of clusters ($|\mathcal{C}|=5$), the algorithm performs on par with Oracle 
model. However, when this information is unavailable and an incorrect number of clusters is 
provided ($|\mathcal{C}|=2$), IFCA performance drops significantly (IFCA incorrect). FedRelax 
outperforms IFCA in that case, but does not achieve the level shown by Oracle model or IFCA 
with correct number of clusters. 

FedAvg is a centralized FL method where gradients computed locally on the nodes are sent to a 
central server. The server then averages these gradients and sends them back to the nodes, where 
they are used to update the local parameter vectors via a gradient descent step. FedAvg performs 
worse than FedRelax, IFCA, and the Oracle model. Nevertheless, even FedAvg achieves a lower 
MSE on estimated weight vectors than training the model locally (Figure \ref{fig:param_clustered_benchmark}, 
$\alpha=0$). In other words, in high $\frac{\dimlocalmodel}{m}$ scenarios, any regularization, even 
suboptimal, is better than none.

\vspace{1.5mm}
\subsubsection{Scikit-learn decision tree regressor} 

To test Algorithm \ref{alg_fed_relax_nonparam} in a non-parametric model scenario, we apply it to 
the same clustered dataset as previously (3 clusters, 50 datasets per cluster, $m_i=10$) with training 
implemented using the \texttt{DecisionTreeRegressor().fit()} function from the scikit-learn Python library. 
This function allows the specifying of individual weights for each data point in the training set.

We apply the update in Eq.(\ref{equ_update_step_fed_relax}) for a candidate $\nodeidx$ by training a 
decision tree (with a maximum depth of 5) on the augmented dataset 
\begin{equation}
\mathcal{D}^{(\nodeidx, aug)} = \bigcup_{\nodeidx' \in \neighbourhood{i}} \left\{ \pair{\featurevec}{\widehat{\hypothesis}^{(\nodeidx')}(\featurevec)} \right\}_{\featurevec \in \publicds{\nodeidx'} } \cup \localdataset{\nodeidx}. 
\end{equation} 
This is possible because both, local loss $\locallossfunc{\nodeidx}{\cdot}$ and the variation between two hypotheses $\variation{\nodeidx}{\nodeidx'}{\localhypothesis{\nodeidx}}{\localhypothesis{\nodeidx'}}$, are defined as \gls{sqerrloss} \eqref{equ_squared_loss}. This process is equivalent to training $\hypospace$ on the local dataset $\localdataset{\nodeidx}$, augmented with data points $(\featurevec, \widehat{\hypothesis}^{(\nodeidx')}(\featurevec))$,  where $\featurevec \in \publicds{\nodeidx'}$ for all $\nodeidx' \in \neighbourhood{i}$. The data points in the augmented dataset are weighted by $\alpha$. 

The decision tree implementation in scikit-learn does not support incremental learning. 
At each iteration, the model is re-fitted to the incoming data, losing all previous knowledge. 
To address this, we introduce a simple trick similar to self-distillation. Before each training 
step on the augmented dataset, we select an arbitrary unlabelled dataset - sampled from a 
Gaussian distribution, $\mathbf{z} \sim \mathcal{N}(\mathbf{0},I_d)$ - and obtain predictions 
from the current (trained) local model, $\widehat{\hypothesis}^{(\nodeidx,k)}$, at iteration $k$. 
The updated model, $\widehat{\hypothesis}^{(\nodeidx,k+1)}$, is obtained by fitting the 
current model, $\widehat{\hypothesis}^{(\nodeidx,k)}$, on the combined dataset:
\begin{equation}
	 \left\{ \pair{\mathbf{z}^{(r)}}{\widehat{\hypothesis}^{(\nodeidx,k)}(\mathbf{z}^{(r)})} \right\}_{r=1}^{100}  \cup  \mathcal{D}^{(\nodeidx, aug)}. 
\end{equation} 
Alternatively, one can use a decision tree algorithm that supports online incremental learning. 

As a stopping criterion in Algorithm \ref{alg_fed_relax_nonparam}, we use a fixed number of $K = 5$ iterations. The MSE on the local validation set, $MSE^{(val)}$, is averaged across all nodes over five experiment runs. 

We apply Algorithm \ref{alg_fed_relax_nonparam} to clustered datasets obtained with $\dimlocalmodel \in \{2,10,50, \}$ and noise level $\sigma=0$ and present the results in Figure \ref{fig:nonparam_clustered_dm}. The curves present the $MSE^{(val)}$  of the local models normalized by $MSE^{(val)}$  incurred by the corresponding Oracle model on the local validation sets. The Oracle model is trained on all local datasets belonging to the same cluster.
We can see that, with the FedRelax approach, the MSE incurred on the local validation set decreases for the cases where $\frac{\dimlocalmodel}{m} \geq 1$.

\begin{figure}[htbp]
	\centering
	\begin{tikzpicture}[remember picture]
	\begin{groupplot}[
		group style={
			group size=3 by 2,  % 2 column, 2 rows
			horizontal sep=0.1cm,
			vertical sep=1.8cm,    % Space between the two plots
			y descriptions at=edge left,
			ylabels at=edge left,
			yticklabels at=edge left,
			x descriptions at=edge bottom,
			xlabels at=edge bottom,
			xticklabels at=edge bottom,
			group name=myplots
			},
		height=4cm,
		width=0.21\textwidth,
		ymin=0.7,
		ymax=22,
		ylabel={\normalsize $MSE^{(val)}$}, % Only appears on the first plot
		ylabel style={align=center, yshift=0cm}, 
%		ymode=log,
		ticklabel style={font=\footnotesize},
		cycle list name=mycolors,
		legend style={draw=none, at={(-0.5,-0.1)}, anchor=north, nodes={scale=0.55}, legend columns=-1, font=\Large},
		legend image post style={scale=0.4},
		]
		
		% First row
		\nextgroupplot[title={\small $\alpha=0$}, axis lines=left, axis line style={-}, ymode=log]
		\pgfplotsforeachungrouped \i/\j in {1/0.2, 2/1, 4/5}{
			\addplot+ [mark=none, very thick, solid] table [
			x index=0, 
			y index=\i,
			col sep=comma
			]{Fedrelax_DT_dm.csv};
		}
		\addplot[dashed, domain=0:4] {1};
		\nextgroupplot[title={\small $\alpha=0.001$}, axis lines=left, axis line style={-}, ymode=log]
		\pgfplotsforeachungrouped \i/\j in {6/0.2, 7/1, 9/5}{
			\addplot+ [mark=none, very thick] table [
			x index=0, 
			y index=\i,
			col sep=comma
			] {Fedrelax_DT_dm.csv};
		}
		\addplot[dashed, domain=0:4] {1};
		\nextgroupplot[title={\small $\alpha=1$}, axis lines=left, axis line style={-}, ymode=log]
		\pgfplotsforeachungrouped \i/\j in {11/0.2, 12/1, 14/5}{
			\addplot+ [mark=none, very thick] table [
			x index=0, 
			y index=\i,
			col sep=comma
			] {Fedrelax_DT_dm.csv};
			\edef\temp{\noexpand\addlegendentry{$\dimlocalmodel /\/ \localsamplesize{\nodeidx} = \j$}}
			\temp
		}
		\addplot[dashed]  coordinates {(0,1) (4,1)};
		
		% Second row
		\nextgroupplot[axis lines=left, axis line style={-}, ymax=3, xtick={0,1,2,3,4}, xticklabels={0,1,2,3}]
		\pgfplotsforeachungrouped \i/\j in {1/0, 2/1, 3/5}{
			\addplot+ [mark=none, very thick, solid] table [
			x index=0, 
			y index=\i,
			col sep=comma
			] {Fedrelax_DT_noise.csv};
		}
		\addplot[dashed, domain=0:4] {1};
		\nextgroupplot[xlabel={\small $k$}, axis lines=left, axis line style={-}, ymax=3, xtick={0,1,2,3,4}, xticklabels={0,1,2,3}]
		\pgfplotsforeachungrouped \i/\j in {4/0, 5/1, 6/5}{
			\addplot+ [mark=none, very thick] table [
			x index=0, 
			y index=\i,
			col sep=comma
			] {Fedrelax_DT_noise.csv};
		}
		\addplot[dashed, domain=0:4] {1};
		\nextgroupplot[legend style={draw=none, at={(-0.5,-0.4)}, anchor=north, legend columns=-1}, axis lines=left, axis line style={-}, ymax=3]
		\pgfplotsforeachungrouped \i/\j in {7/0, 8/1, 9/5}{
				\addplot+ [mark=none, very thick] table [
				x index=0, 
				y index=\i,
				col sep=comma
				] {Fedrelax_DT_noise.csv};
				\edef\temp{\noexpand\addlegendentry{$\sigma$ = \j}}
				\temp
			}
			\addplot[dashed, domain=0:4] {1};
	\end{groupplot}
	\tikzset{SubCaption/.style={
			text width=2in, yshift=0mm, align=center, anchor=north, outer sep=0.5cm,
	}}
	\tikzset{SubCaptionBottom/.style={
		text width=3in, yshift=-2mm, align=center, anchor=north, outer sep=1cm,
	}}
	\node[SubCaption] at ($(myplots c1r1.south)!0.2!(myplots c3r1.south)$) {\subcaption{\mbox{${\rm MSE}^{(val)}$ normalized by Oracle w.r.t. $\frac{d}{m}$ ratio.} \label{fig:nonparam_clustered_dm}}};
	\node[SubCaptionBottom] at ($(myplots c1r2.south)!0.4!(myplots c3r2.south)$) {\subcaption{\mbox{${\rm MSE}^{(val)}$ normalized by Oracle  w.r.t. $\sigma$, $d/m=1$.}\label{fig:nonparam_clustered_noise}}};
\end{tikzpicture}
	\caption{${\rm MSE}^{(val)}$, normalized by Oracle, incurred by FedRelax on a clustered dataset and decision tree regressors as models. The y-axis in subplot (a) is on a logarithmic scale. The code to generate the figure is available at \href{https://github.com/shamPJ/het-FL/blob/main/src/FedRelax.ipynb}{Github}.}  
	\label{fig:nonparam_clustered}
\end{figure}

Introducing noise into the dataset further highlights the denoising effect of FedRelax, particularly for local datasets with high noise levels (Figure \ref{fig:nonparam_clustered_noise}).

\subsection{Synthetic Dataset - non-i.i.d. distribution}
\label{sec_exp_noniid_ds}

In the clustered dataset described above, local datasets can be divided into disjoint clusters, with datasets belonging to the same cluster sampled from the same probability distribution. In the following experiments, we create heterogeneous (non-i.i.d.) datasets by sampling each node's data from a mixture of $K$ Gaussian distributions, with node-specific mixing vector $\rho^{(\nodeidx)}$. The local datasets are created using Algorithm \ref{data_gen_noniid}. 

\begin{algorithm}[htbp]
  %\floatname{algorithm}{ Non-i.i.d. dataset generation}
   \caption{Non-i.i.d. data generation}
	\label{data_gen_noniid}
	{\bf Input}: number of nodes $n$ for which local datasets $ \localdataset{\nodeidx}$ are generated, sample size $m_i$ of local dataset $ \localdataset{\nodeidx}$, noise level $\sigma^2$, number of distributions $K$ \\
	{\bf Initialize}: parameter vectors $ \overline{\weights}^{(k)} \in \mathbb{R}^{\nrfeatures}$; $ \overline{\weights}^{(k)}   \sim  \mathcal{N}(\mathbf{0}, \mathbf{I}_{\nrfeatures})$; $k=1,...,K$
	\begin{algorithmic}[1]
		\For{a node $\nodeidx=1,\dots,n$}
			\State sample random vector  $ s^{(i)} \sim  \mathcal{N}(\mathbf{0},\mathbf{I}_{K})$
			\State get a mixing vector $\rho^{(\nodeidx)} := softmax(s^{(\nodeidx)})$, with components $\rho^{(\nodeidx,k)} \in [0,1], \quad \sum_{k=1}^{K} \rho^{(\nodeidx,k)} =1$
			\For{a distribution $k=1,\dots,K$}
				\State compute sample size $m_i^{(k)} = m_i \cdot \rho^{(\nodeidx,k)}$
					\For{each data point $r = 1, \dots, m_i^{(k)} $}
						\State sample feature vector  $\mathbf{x}^{(i,r)} \sim \mathcal{N}(\mathbf{0},\mathbf{I}_{\dimlocalmodel})$
						\State generate label ${y}^{(i,r)} := (\overline{\weights}^{(k)})^T\mathbf{x}^{(i,r)} + {\varepsilon}^{(\nodeidx,r)}$, where ${\varepsilon}^{(\nodeidx,r)} \sim \mathcal{N}(0, \sigma^2)$
					\EndFor
			\EndFor
			\State collect all generated data points into a dataset on the node $\nodeidx$:  $\localdataset{\nodeidx} = \left\{ \big(\featurevec^{(\nodeidx,1)},\truelabel^{(\nodeidx,1)}\big), \ldots,\big(\featurevec^{(\nodeidx,\samplesize_{\nodeidx})},\truelabel^{(\nodeidx,\samplesize_{\nodeidx})}\big) \right\} $
		\EndFor
	\end{algorithmic}
	{\bf Output}: Local datasets for $n$ nodes
	$\{  \localdataset{\nodeidx} \}_{\nodeidx=1}^{n}$
\end{algorithm}

The algorithm generates non-i.i.d. datasets by assigning each node a unique mixture of $K$ Gaussian distributions. The components of the vector $\rho^{(\nodeidx)}$ are softmax-based mixing ratios $\rho^{(\nodeidx,k)}$, which determine the proportion of data points drawn from each of the $K$ distributions.

The similarity $\edgeweight_{\nodeidx,\nodeidx'}  \in [0,1]$ between nodes $\nodeidx$ and $\nodeidx'$ is determined by the proportion of similar data points they possess:
\begin{equation}
	 	\label{eq_noniid_similarity}
	 	\edgeweight_{\nodeidx,\nodeidx'} =  \sum_{k=1}^{K} \min(\rho^{(\nodeidx,k)}, \rho^{(\nodeidx',k)}).
\end{equation}
We create such synthetic heterogeneous local datasets with the number of distributions 
set to $K=3$ and the total number of nodes set to $|\nodes|=150$. Each local dataset consists 
of a training subset of size $m_i=10$ and a validation subset of size $m^{(val)}_i=100$. 
We test datasets with varying dimensionality $\dimlocalmodel \in \{2,10,50 \}$ for the 
feature vector  $\mathbf{x}^{(i,r)}  \in  \mathbb{R}^{\nrfeatures}$. 

The shared public dataset $\publicds{i}$ is again sampled from a standard multivariate 
normal distribution, $\mathbf{x}^{(i,r)} \sim \mathcal{N}(\mathbf{0},\mathbf{I}_{\dimlocalmodel})$, 
for each node. The sample size is $m_{i}^{(\mathcal{P})}=100$ for all nodes. 

\vspace{1.5mm}
\subsubsection{PyTorch linear model}

We generate a graph with synthetic heterogeneous local datasets on the nodes, as described 
above. Local models are linear regressors implemented in PyTorch (with the RMSprop optimizer 
and a learning rate of 0.01). As a stopping criterion in Algorithm \ref{alg_fed_relax_param}, 
we use a fixed number of $K = 500$ iterations. 

We run Algorithm \ref{alg_fed_relax_param} with different regularization strength $\alpha \in \{0,0.01,0.05\}$ 
and compute the $MSE^{(val)}$ incurred on the local validation subset for each $\alpha$ value. 
These MSE values, normalized by the MSE values for $\alpha=0$, i.e., normalized by the MSE 
incurred on the validation dataset by the locally trained model, are plotted in Figure \ref{fig:param_noniid}.
The curves represent the MSE values averaged across all nodes and the five runs of the experiment. 
For each run, new local $\mathcal{D}^{(i)}$ and public $\publicds{i}$  datasets are generated. 

In Figure \ref{fig:param_noniid}, the left and middle subplots show FedRelax's performance
 on noiseless ($\sigma=0$) non-i.i.d. datasets. It is evident that our pooling strategy results 
 in lower MSE on local validation datasets compared to the baseline (locally trained model) 
 for all $\frac{\dimlocalmodel}{m}$ ratios and values $\alpha \in \{0.01, 0.05\}$.
We then added different levels of noise to the datasets ($\sigma \in \{0, 1, 5\}$), with the 
regularization term fixed at $\alpha=0.01$. The results are depicted in the right subplot of 
Figure \ref{fig:param_noniid}, where we can see that FedRelax improves the performance 
of the local model even with high levels of noise injected into the datasets. 

\begin{figure}[htbp]
	\centering
	\begin{tikzpicture}[remember picture]
	\begin{groupplot}[
		group style={
			group size=3 by 1, 
			horizontal sep=0.1cm,
			vertical sep=1.8cm,    % Space between the two plots
			y descriptions at=edge left,
			ylabels at=edge left,
			yticklabels at=edge left,
			x descriptions at=edge bottom,
			xlabels at=edge bottom,
			xticklabels at=edge bottom,
			group name=myplots
			},
		height=4cm,
		width=0.21\textwidth,
		ymin=0,
		ymax=1.2,
		ylabel={\normalsize $MSE^{(val)}$}, % Only appears on the first plot
		ylabel style={align=center, yshift=0cm}, 
		ticklabel style={font=\footnotesize},
		cycle list name=mycolors,
		legend image post style={scale=0.4},
		]
		
		% First row
		\nextgroupplot[title={\small $\alpha=0.01$}, axis lines=left, axis line style={-},
			 					legend style={draw=none, at={(0,-0.6)}, anchor=south west, font=\footnotesize, legend columns=-1,}]
		\pgfplotsforeachungrouped \i/\j in {1/0.2, 2/1, 4/5}{
			\addplot+ [mark=none, very thick, solid] table [
			x index=0, 
			y index=\i,
			col sep=comma
			]{FedRelax_noniid_dm.csv};
			\edef\temp{\noexpand\addlegendentry{$\dimlocalmodel /\/ \localsamplesize{\nodeidx} = \j$}}
			\temp
		}
		\addplot[dashed, domain=0:499] {1};
		\nextgroupplot[title={\small $\alpha=0.05$}, xlabel={\small $k$}, axis lines=left, axis line style={-}]
		\pgfplotsforeachungrouped \i/\j in {6/0.2, 7/1, 9/5}{
			\addplot+ [mark=none, very thick] table [
			x index=0, 
			y index=\i,
			col sep=comma
			] {FedRelax_noniid_dm.csv};
		}
		\addplot[dashed, domain=0:499] {1};
		\nextgroupplot[title={\small $\alpha=0.01$}, axis lines=left, axis line style={-},
									legend style={font=\scriptsize, draw=none, fill=none, at={(1,1)}, anchor=north east}, cycle list name=grad]
		\pgfplotsforeachungrouped \i/\j in {1/0, 2/1, 3/5}{
			\addplot+ [mark=none, very thick] table [
			x index=0, 
			y index=\i,
			col sep=comma
			] {FedRelax_noniid_noise.csv};
			\edef\temp{\noexpand\addlegendentry{$\sigma$ = \j}}
			\temp
		}
		\addplot[dashed, domain=0:499] {1};
		
	\end{groupplot}
\end{tikzpicture}
	\caption{FedRelax on non-i.i.d. datasets and linear regressors as models, ${\rm MSE}^{(val)}$ 
		is normalized by the ${\rm MSE}^{(val)}$ incurred by the model trained locally ($\alpha=0$). 
		In the first two subplots, the noise level is fixed at $\sigma=0$, while in the last subplot, 
		$d/\localsamplesize{\nodeidx}=1$. The code to generate the figure is available at \href{https://github.com/shamPJ/het-FL/blob/main/src/FedRelax-noniid.ipynb}{Github}.}
\label{fig:param_noniid}
\end{figure}

\vspace{1.5mm}
\subsubsection{Scikit-learn decision tree regressor} 

We test  Algorithm \ref{alg_fed_relax_nonparam} on non-i.i.d. distributed data and use the 
sklearn  \texttt{DecisionTreeRegressor()} as a local model (with the maximum tree depth set to 5) 
on the nodes. The model update is implemented in the same way as for experiments on the 
clustered dataset.

The resulting normalized $MSE^{(val)}$ values are depicted in the Figure \ref{fig:nonparam_noniid}. 
Compared to training the model only locally ($\alpha=0$), the MSE incurred on the validation set is 
lower for different $\frac{\dimlocalmodel}{m}$ ratios (see Figure  \ref{fig:nonparam_noniid}, left subplot). 
On the other hand, strong regularization may lead to a deterioration in performance, especially for a 
low $\frac{\dimlocalmodel}{m}$ ratio (see Figure  \ref{fig:nonparam_noniid}, mid-subplot). We 
observe that adding the GTV term  \eqref{equ_def_gtv_non_param} also improves performance 
on datatsets with non-zero noise levels (see Figure  \ref{fig:nonparam_noniid}, right subplot).

\begin{figure}[htbp]
	%\centering
	\begin{tikzpicture}[remember picture]
	\begin{groupplot}[
		group style={
			group size=3 by 1, 
			horizontal sep=0.1cm,
			vertical sep=1.8cm,    % Space between the two plots
			y descriptions at=edge left,
			ylabels at=edge left,
			yticklabels at=edge left,
			x descriptions at=edge bottom,
			xlabels at=edge bottom,
			xticklabels at=edge bottom,
			group name=myplots
			},
		height=4cm,
		width=0.21\textwidth,
		ymin=0.4,
		ymax=1.8,
		ylabel={\normalsize $MSE^{(val)}$}, % Only appears on the first plot
		ylabel style={align=center, yshift=0cm}, 
		ymode=log,
		ticklabel style={font=\footnotesize},
		cycle list name=mycolors,
		legend image post style={scale=0.4},
		]
		
		% First row
		\nextgroupplot[title={\small $\alpha=0.001$}, axis lines=left, axis line style={-},
			 					legend style={draw=none, at={(0,-0.6)}, anchor=south west, font=\footnotesize, legend columns=-1},
			 					xtick={0,1,2,3,4}, xticklabels={0,1,2,3}]
		\pgfplotsforeachungrouped \i/\j in {1/0.2, 2/1, 4/5}{
			\addplot+ [mark=none, very thick, solid] table [
			x index=0, 
			y index=\i,
			col sep=comma
			]{FedRelax_noniid_DT_dm.csv};
			\edef\temp{\noexpand\addlegendentry{$\dimlocalmodel /\/ \localsamplesize{\nodeidx} = \j$}}
			\temp
		}
		\addplot[dashed, domain=0:4, on layer=axis background] {1};
		\nextgroupplot[title={\small $\alpha=1$}, xlabel={\small $k$}, axis lines=left, axis line style={-},
									xtick={0,1,2,3,4}, xticklabels={0,1,2,3}]
		\pgfplotsforeachungrouped \i/\j in {6/0.2, 7/1, 9/5}{
			\addplot+ [mark=none, very thick] table [
			x index=0, 
			y index=\i,
			col sep=comma
			] {FedRelax_noniid_DT_dm.csv};
		}
		\addplot[dashed, domain=0:4, on layer=axis background] {1};
		\nextgroupplot[title={\small $\alpha=1$}, axis lines=left, axis line style={-},
									legend style={font=\scriptsize, row sep=0.1pt, column sep=2pt, draw=none, fill=none, at={(1,0.95)}, anchor=north east}, cycle list name=grad]
		\pgfplotsforeachungrouped \i/\j in {1/0, 2/1, 3/5}{
			\addplot+ [mark=none, very thick] table [
			x index=0, 
			y index=\i,
			col sep=comma
			] {FedRelax_noniid_DT_noise.csv};
			\edef\temp{\noexpand\addlegendentry{$\sigma$ = \j}}
			\temp
		}
		\addplot[dashed, domain=0:4] {1};
		
	\end{groupplot}
\end{tikzpicture}
	\caption{FedRelax on non-i.i.d. datasets with decision tree regressors as models, normalized 
		${\rm MSE}^{(val)}$. The y-axis is on a logarithmic scale. In the first two subplots, the noise 
		level is fixed at $\sigma=0$, while in the last subplot, $d/\localsamplesize{\nodeidx}=1$. 
		The code to generate the figure is available at \href{https://github.com/shamPJ/het-FL/blob/main/src/FedRelax-noniid.ipynb}{Github}.}
\label{fig:nonparam_noniid}
\end{figure}

\section{Analysis of Algorithm \ref{alg_fed_relax_param} for Local Linear Models} 

The GTVmin term, as described in \eqref{equ_def_gtv_non_param}, enforces similarity among 
hypotheses at connected nodes, thus recovering the clustered structure of local datasets \eqref{equ_def_true_linear_model_SBM}.
 We aim to theoretically and empirically analyze the constraint this term imposes on local hypothesis 
 variation. For parametric models, we quantify the variation at a node $\nodeidx$ as the deviation 
 of its learned parameter vector from the cluster average:
 \begin{equation}
 	\label{equ_def_error_component_clustered}
 	\widetilde{\vw}^{(\nodeidx)} \defeq \estlocalparams{\nodeidx} - (1/|\cluster|)\sum_{\nodeidx' \in \cluster} \estlocalparams{\nodeidx'} \mbox{, for } \nodeidx \in \cluster,
 \end{equation} 
 For simplicity, we consider clustered datasets following \eqref{equ_def_true_linear_model_SBM}, 
 with a single cluster $\cluster$ sharing a true parameter vector $\overline{\weights}^{(\cluster)}$. 
 Each local dataset is generated as:
 \begin{equation} 
 	\label{equ_def_probmodel_linreg_node_i}
 	\hspace*{-3mm}\labelvec^{(\nodeidx)}\!=\! \featuremtx^{(\nodeidx)}  \overline{\weights}^{(\cluster)}\!+\!{\bm \varepsilon}^{(\nodeidx)},  \mbox{ for all } \nodeidx \in \cluster.
 \end{equation} 
 Furthermore, we assume the existence of a cluster-specific error tolerance  $\varepsilon^{(\cluster)}$ such that:
	\begin{equation} 
 	\label{equ_def_cluster_wise_opt}
 	\sum_{\nodeidx \in \cluster} \locallossfunc{\nodeidx}{\overline{\weights}^{(\cluster)}} \leq
 	\varepsilon^{(\cluster)}.
	\end{equation} 

Under these conditions, the Euclidean norm of the variation $\widetilde{\vw}^{(\nodeidx)}$ is bounded by (see \cite{jung2024analysis})
 \begin{equation} 
 	\sum_{\nodeidx \in \cluster} \normgeneric{\widetilde{\vw}^{(\nodeidx)} }{2}^{2} \!\leq\!\frac{\varepsilon^{(\cluster)} }{\regparam \eigval{2}\big( \mL^{(\cluster)} \big)}. \label{equ_upper_bound_single_cluster}
 \end{equation} 
 Thus, in a single-cluster setting ($\cluster = \nodes$), the bound depends on the error tolerance 
 $\varepsilon^{(\cluster)}$, the regularization parameter $\regparam$, and the second-smallest 
 eigenvalue $\eigval{2}\big( \mL^{(\cluster)} \big)$ of the cluster Laplacian $\LapMat{\cluster}$, 
 which reflects the strength of connectivity within the cluster.
  
  When using linear regression as the local model and MSE as the local loss, we have:
  \begin{align} 
  	\sum_{\nodeidx \in \cluster} \locallossfunc{\nodeidx}{\overline{\weights}^{(\cluster)}} 
  	&=\! 	\sum_{\nodeidx \in \cluster} \frac{1}{\localsamplesize{\nodeidx}} \normgeneric{ \labelvec^{(\nodeidx)} - \featuremtx^{(\nodeidx)} \nonumber \overline{\weights}^{(\cluster)}}{2}^{2} \nonumber \\
  	&=\! \sum_{\nodeidx \in \cluster} \frac{1}{\localsamplesize{\nodeidx}} \normgeneric{{\bm \varepsilon}^{(\nodeidx)}}{2}^{2} . \nonumber
\end{align}
Thus, setting
 \begin{equation} 
 	 \varepsilon^{(\cluster)} = \sum_{\nodeidx \in \cluster} (1/\localsamplesize{\nodeidx}) \normgeneric{{\bm \varepsilon}^{(\nodeidx)}}{2}^{2}  . \nonumber
 \end{equation} 
satisfies assumption \eqref{equ_def_cluster_wise_opt}.
 
The regularization parameter $\regparam$ controls the trade-off between minimizing the average 
local loss and reducing parameter variation. Larger values of $\regparam$ enforce greater parameter 
alignment at the cost of increased local loss.
 
The nodes of our clustered network are connected with probability ${p}_{in}$ (see section \ref{sec_exp_clustered_ds}), 
and this within-cluster connectivity is characterized by the $\eigval{2}\big( \mL^{(\cluster)} \big)$, 
second-smallest eigenvalue of the cluster Laplacian $\LapMat{\cluster}$,  defined element-wise as:
 \begin{equation} 
 	\LapMatEntry{\cluster}{\nodeidx}{\nodeidx'} \defeq \begin{cases} - \edgeweight_{\nodeidx,\nodeidx'} & \mbox{ for } \nodeidx\neq \nodeidx', \edge{\nodeidx}{\nodeidx'} \in \edges, \nodeidx,\nodeidx' \in \cluster \\ 
 		\sum_{\nodeidx'' \in \cluster \setminus \nodeidx} \edgeweight_{\nodeidx,\nodeidx''} & \mbox{ for } \nodeidx = \nodeidx' \in \cluster \\ 
 		0 & \mbox{ else.} \end{cases}  \nonumber
 \end{equation} 
 
We first study how the norm of variation $\normgeneric{\widetilde{\vw}^{(\nodeidx)}}{2}^{2}$ and 
its bound \eqref{equ_upper_bound_single_cluster} vary with the noise level $\varepsilon^{(\nodeidx)}$. 
We run Algorithm \ref{alg_fed_relax_param} on datasets generated according to \ref{equ_def_probmodel_linreg_node_i} 
and belonging to a single cluster $\cluster$. The number of local datasets is set to $n=50$, the 
sample size $m_i=10$, the dimensionality of the feature vector to $\dimlocalmodel=10$ and the 
within-cluster connectivity to $p_{in}=0.8$. As shown in Figure \ref{fig:variation_noise}, 
Algorithm \ref{alg_fed_relax_param} reduces the variation (solid lines), and the observed 
values remain below the theoretical bound (dashed lines) after a few hundred iterations. 
Higher regularization strengths ($\regparam \in \{0.0001, 0.01, 0.1\}$) yield lower variation, 
while increased noise levels ($\sigma \in \{0.1, 1, 5\}$) lead to larger variation.

\begin{figure}[htbp]
	\begin{tikzpicture}[remember picture]
	\begin{groupplot}[
		group style={
			group size=3 by 2,  % 2 column, 2 rows
			horizontal sep=0.1cm,
			vertical sep=1.8cm,    % Space between the two plots
			y descriptions at=edge left,
			ylabels at=edge left,
			yticklabels at=edge left,
			x descriptions at=edge bottom,
			xlabels at=edge bottom,
			xticklabels at=edge bottom,
			group name=myplots
		},
		height=4cm,
		width=0.21\textwidth,
		ylabel={\small $Variation$}, % Only appears on the first plot
		ylabel style={align=center, yshift=0cm}, 
		ymode=log,
		ticklabel style={font=\footnotesize},
		cycle list name=mycolors3,
		legend style={draw=none, at={(-0.5,-0.1)}, anchor=north, nodes={scale=0.55}, legend columns=-1, font=\Large},
		legend image post style={scale=0.4},
		]
		
		% First row
		\nextgroupplot[title={\small $\alpha=0.0001$}, axis lines=left, axis line style={-}, ymin=0.5e-2, ymax=1e6]
		\pgfplotsforeachungrouped \i in {1,2,3}{
			\addplot+ [mark=none, very thick, solid] table [
			x index=0, 
			y index=\i,
			col sep=comma
			] {FedRelax_w_tilde_noise.csv};
		}
		\pgfplotsforeachungrouped \i in {1,2,3}{
			\addplot+ [mark=none, dashed] table [
			x index=0, 
			y index=\i,
			col sep=comma
			] {FedRelax_bound_noise.csv};
		}
		\nextgroupplot[title={\small $\alpha=0.01$}, axis lines=left, axis line style={-}, ymin=0.5e-2, ymax=1e6]
		\pgfplotsforeachungrouped \i in {4,5,6}{
			\addplot+ [mark=none, very thick] table [
			x index=0, 
			y index=\i,
			col sep=comma
			] {FedRelax_w_tilde_noise.csv};
		}
		\pgfplotsforeachungrouped \i in {4,5,6}{
			\addplot+ [mark=none, dashed] table [
			x index=0, 
			y index=\i,
			col sep=comma
			] {FedRelax_bound_noise.csv};
		}
		\nextgroupplot[title={\small $\alpha=0.1$}, axis lines=left, axis line style={-}, ymin=0.5e-2, ymax=1e6]
		\pgfplotsforeachungrouped \i/\j in {7/0.1, 8/1, 9/5}{
			\addplot+ [mark=none, very thick] table [
			x index=0, 
			y index=\i,
			col sep=comma
			] {FedRelax_w_tilde_noise.csv};
			\edef\temp{\noexpand\addlegendentry{$\sigma$ = \j}}
			\temp
		}
		\pgfplotsforeachungrouped \i in {7,8,9}{
			\addplot+ [mark=none, dashed] table [
			x index=0, 
			y index=\i,
			col sep=comma
			] {FedRelax_bound_noise.csv};
		}
		
		% Second row
		\nextgroupplot[axis lines=left, axis line style={-}, ymin=0.5e-2, ymax=2e3]
		\pgfplotsforeachungrouped \i in {1,2,3}{
			\addplot+ [mark=none, very thick, solid] table [
			x index=0, 
			y index=\i,
			col sep=comma
			] {FedRelax_w_tilde_pin.csv};
		}
		\pgfplotsforeachungrouped \i in {1,2,3}{
		\addplot+ [mark=none, dashed] table [
		x index=0, 
		y index=\i,
		col sep=comma
		] {FedRelax_bound_pin.csv};
		}
		\nextgroupplot[xlabel={\small $k$}, axis lines=left, axis line style={-}, ymin=0.5e-2, ymax=2e3]
		\pgfplotsforeachungrouped \i in {4,5,6}{
			\addplot+ [mark=none, very thick] table [
			x index=0, 
			y index=\i,
			col sep=comma
			] {FedRelax_w_tilde_pin.csv};
		}
		\pgfplotsforeachungrouped \i in {4,5,6}{
			\addplot+ [mark=none, dashed] table [
			x index=0, 
			y index=\i,
			col sep=comma
			] {FedRelax_bound_pin.csv};
		}
		\nextgroupplot[legend style={draw=none, at={(-0.5,-0.4)}, anchor=north, legend columns=-1}, axis lines=left, axis line style={-}, ymin=0.5e-2, ymax=2e3]
		\pgfplotsforeachungrouped \i/\j in {7/0.2, 8/0.6, 9/1}{
			\addplot+ [mark=none, very thick] table [
			x index=0, 
			y index=\i,
			col sep=comma
			] {FedRelax_w_tilde_pin.csv};
			\edef\temp{\noexpand\addlegendentry{$p_{in}$ = \j}}
			\temp
		}
		\pgfplotsforeachungrouped \i in {7,8,9}{
			\addplot+ [mark=none, dashed] table [
			x index=0, 
			y index=\i,
			col sep=comma
			] {FedRelax_bound_pin.csv};
		}
	\end{groupplot}
	\tikzset{SubCaption/.style={
			text width=2in, yshift=0mm, align=center, anchor=north, outer sep=0.5cm,
	}}
	\tikzset{SubCaptionBottom/.style={
			text width=3in, yshift=-2mm, align=center, anchor=north, outer sep=1cm,
	}}
	\node[SubCaption] at ($(myplots c1r1.south)!0.5!(myplots c3r1.south)$) {\subcaption{\mbox{$Variation$ w.r.t. $\sigma$ ratio, $p_{in}=0.8$.} \label{fig:variation_noise}}};
	\node[SubCaptionBottom] at ($(myplots c1r2.south)!0.4!(myplots c3r2.south)$) {\subcaption{\mbox{$Variation$ w.r.t. $p_{in}$, $\sigma=0.1$.}\label{fig:variation_pin}}};
\end{tikzpicture}
	\caption{Variation $\sum_{\nodeidx \in \cluster} \normgeneric{\widetilde{\vw}^{(\nodeidx)} }{2}^{2}$ 
		and its bound under FedRelax on single-cluster dataset. Results are averaged over 50 runs. 
		The y-axis is on a logarithmic scale. The code to generate the figure is available at \href{https://github.com/shamPJ/het-FL/blob/main/src/FedRelax_variation.ipynb}{Github}.}
	\label{fig:bound}
\end{figure}

Next, we investigate the effect of cluster connectivity by running Algorithm \ref{alg_fed_relax_param} 
in a low-noise setting ($\sigma = 0.1$), varying the within-cluster connectivity $p_{in} \in \{0.2, 0.6, 1\}$. 
Higher connectivity results in a denser graph, leading to larger $\eigval{2}\big( \mL^{(\cluster)} \big)$. 
As predicted, Figure \ref{fig:variation_pin} shows that greater connectivity leads to smaller variation 
$\sum_{\nodeidx \in \cluster} \normgeneric{\widetilde{\vw}^{(\nodeidx)} }{2}^{2}$.

\section{Conclusion} 
We have introduced a knowledge distillation-based, model-agnostic method to train 
heterogeneous networks of personalized \gls{localmodel}s. Each \gls{localmodel} in the 
network can be trained on a local \gls{dataset} which, however, might not provide sufficient 
statistical power for successful training. Therefore, we couple the training of \gls{localmodel}s 
for statistically similar \gls{localdataset}s indirectly, through predictions on shared public datasets. 
The similarity structure between \gls{localdataset}s and their corresponding \gls{localmodel}s 
is represented by an \gls{empgraph}. We use the undirected and weighted edges of the \gls{empgraph} 
to construct a regularization term that couples the \gls{localmodel}s. In particular, 
the \gls{regularization} forces the \gls{localmodel}s at well-connected nodes of the \gls{empgraph} 
to agree in their predictions on shared public datasets of unlabeled \gls{datapoint}s. FedRelax 
can be used in networks with model (linear models, neural networks, SVMs) and data  heterogeneity 
(clustered, non-i.i.d. local datasets). 

\section*{Acknowledgement} 
We are grateful to Olga Kuznetsova and Mikko Seesto for providing feedback on the manuscript. 

\bibliographystyle{IEEEtran}
\bibliography{Literature}

\end{document}